\documentclass[sigconf]{acmart}
\acmSubmissionID{1234}

\usepackage{booktabs} 

\def\model{Palette\xspace}
\usepackage{graphicx}
\usepackage{amsmath}
\usepackage{amsfonts}

\usepackage{bm}
\usepackage{float}
\usepackage{enumitem}
\usepackage{multirow}

\def\eqref#1{equation~\ref{#1}}

\def\1{\bm{1}}

\def\vx{{\bm{x}}}
\def\vy{{\bm{y}}}
\def\vz{{\bm{z}}}

\DeclareMathAlphabet{\mathsfit}{\encodingdefault}{\sfdefault}{m}{sl}
\SetMathAlphabet{\mathsfit}{bold}{\encodingdefault}{\sfdefault}{bx}{n}

\newcommand{\E}{\mathbb{E}}

\usepackage{amsthm,amsmath,amsfonts,bm,xspace}
\usepackage{bbm}
\usepackage{color}
\usepackage{enumitem}

\def\1{\bm{1}}

\def\vx{{\bm{x}}}
\def\veps{{\bm{\epsilon}}}

\def\vty{\widetilde{\bm{y}}}
\def\vy{{\bm{y}}}
\def\vz{{\bm{z}}}

\def\stdnormal{\mathcal{N}(\bm{0}, \bm{I})}

\DeclareMathAlphabet{\mathsfit}{\encodingdefault}{\sfdefault}{m}{sl}
\SetMathAlphabet{\mathsfit}{bold}{\encodingdefault}{\sfdefault}{bx}{n}

\def\eg{{\em e.g.,\xspace}}
\def\ie{{\em i.e.,\xspace}}

\usepackage{algorithm, algorithmic}

\copyrightyear{2022}
\acmYear{2022}
\setcopyright{acmcopyright}
\acmConference{ACM SIGGRAPH}{August 8-11, 2022}{Vancouver}
\acmDOI{10.1145/8888888.7777777}
\acmISBN{978-1-4503-1234-5/22/07}

\begin{document}
\title{Palette: Image-to-Image Diffusion Models}


\author{Chitwan Saharia, William Chan, Huiwen Chang, Chris A Lee, Jonathan Ho, Tim Salimans}
\author{David J Fleet, Mohammad Norouzi} 
\affiliation{%
 \institution{Google Research, Brain Team}
\country{Canada}
 }
\email{{sahariac,williamchan,davidfleet,mnorouzi}@google.com}


\renewcommand\shortauthors{Saharia, C. et al}

\begin{abstract}
This paper develops a unified framework for image-to-image translation based on conditional diffusion models and evaluates this framework on four challenging image-to-image translation tasks, namely colorization, inpainting, uncropping, and JPEG restoration. Our simple implementation of image-to-image diffusion models outperforms strong GAN and regression baselines on all tasks, without task-specific hyper-parameter tuning, architecture customization, or any auxiliary loss or sophisticated new techniques needed. We uncover the impact of an L2 vs. L1 loss in the denoising diffusion objective on sample diversity, and demonstrate the importance of self-attention in the neural architecture through empirical studies. Importantly, we advocate a unified evaluation protocol based on ImageNet, with human evaluation and sample quality scores (FID, Inception Score, Classification Accuracy of a pre-trained ResNet-50, and Perceptual Distance against original images). We expect this standardized evaluation protocol to play a role in advancing image-to-image translation research. Finally, we show that a generalist, multi-task diffusion model performs as well or better than task-specific specialist counterparts. Check out \href{https://diffusion-palette.github.io/}{https://diffusion-palette.github.io/} for an overview of the results and code.
\end{abstract}

%
%

%
%


\maketitle

\section{Introduction}
\label{sec:intro}

\begin{figure}[t]
\small
\begin{center}
{\small
\hspace*{-0.25cm}
\begin{tabular}{@{}c@{\hspace{.15cm}}c@{\hspace{.095cm}}c@{\hspace{.095cm}}c@{}}
     & Input & Output & Original \\
     \raisebox{.4cm}{\rotatebox{90}{Colorization}} &
     \frame{\includegraphics[width=.3\linewidth]{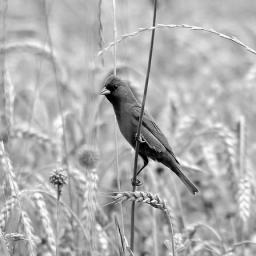}} &
     \frame{\includegraphics[width=.3\linewidth]{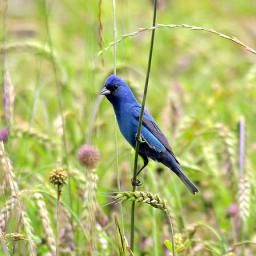}} &
     \frame{\includegraphics[width=.3\linewidth]{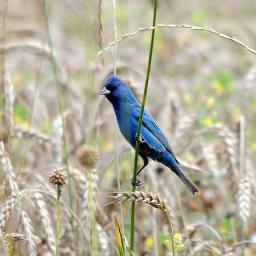}} \\
     \raisebox{.65cm}{\rotatebox{90}{Inpainting}} &
     \frame{\includegraphics[width=.3\linewidth]{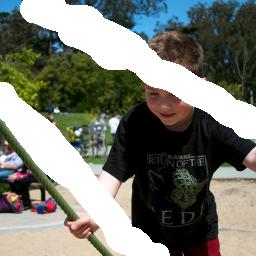}} &
     \frame{\includegraphics[width=.3\linewidth]{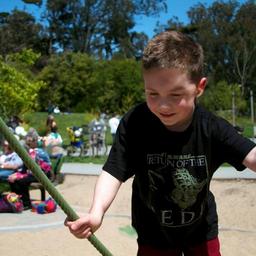}} &
     \frame{\includegraphics[width=.3\linewidth]{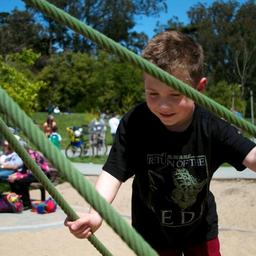}} \\
     \raisebox{.5cm}{\rotatebox{90}{Uncropping}} &
     \frame{\includegraphics[width=.3\linewidth]{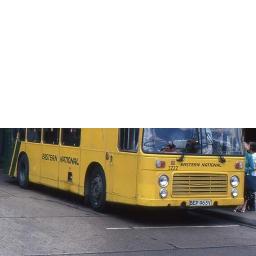}} &
     \frame{\includegraphics[width=.3\linewidth]{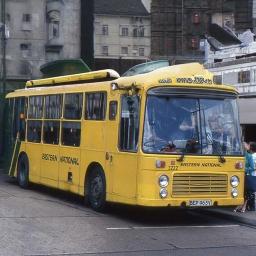}} &
     \frame{\includegraphics[width=.3\linewidth]{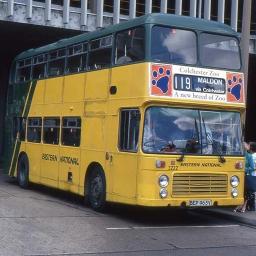}}\\     
     \raisebox{.2cm}{\rotatebox{90}{JPEG restoration}} &
     \frame{\includegraphics[width=.3\linewidth]{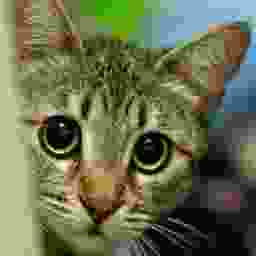}} &
     \frame{\includegraphics[width=.3\linewidth]{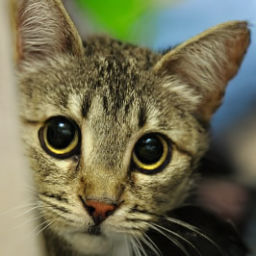}} &
     \frame{\includegraphics[width=.3\linewidth]{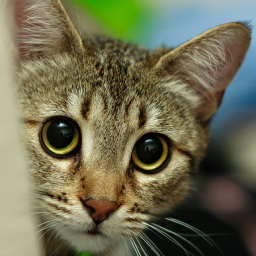}} \\       
\end{tabular}
}
\end{center}
\vspace{-0.3cm}
\caption{Image-to-image diffusion models are able to generate high-fidelity output across tasks without task-specific customization or auxiliary loss.
}
\vspace{-0.4cm}
\label{fig:1}
\end{figure}


Many problems in vision and image processing can be formulated as image-to-image translation. Examples include restoration tasks, like super-resolution, colorization, and inpainting, as well as pixel-level image understanding tasks, such as instance segmentation and depth estimation.  
Many such tasks, like those in Fig.\ \ref{fig:1}, are complex inverse problems, where multiple output images are consistent with a single input.
A natural approach to image-to-image translation is to learn the conditional distribution of output images given the input, using deep generative models that can capture multi-modal distributions in the high-dimensional space of images.

\begin{figure*}[t]
\small
\begin{center}
{\includegraphics[width=1.0\textwidth]{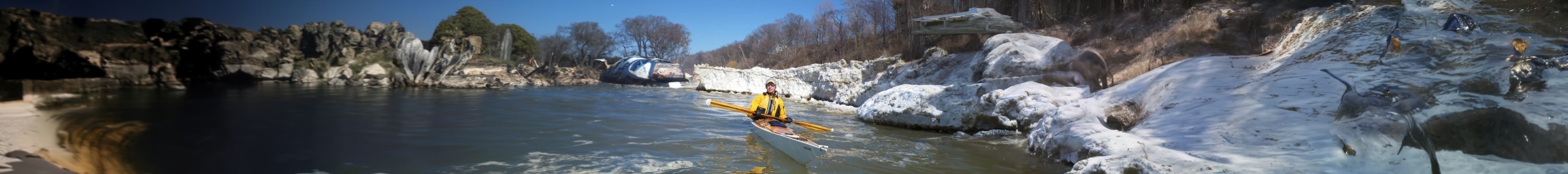}} \\ [.3em]
{\includegraphics[width=1.0\textwidth]{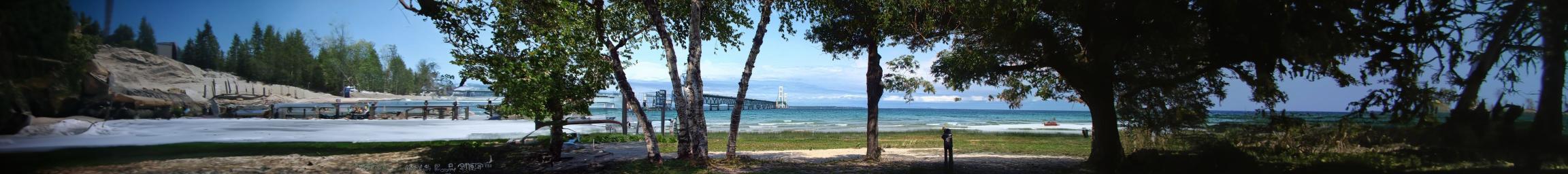}} \\ [.3em]
{\includegraphics[width=1.0\textwidth]{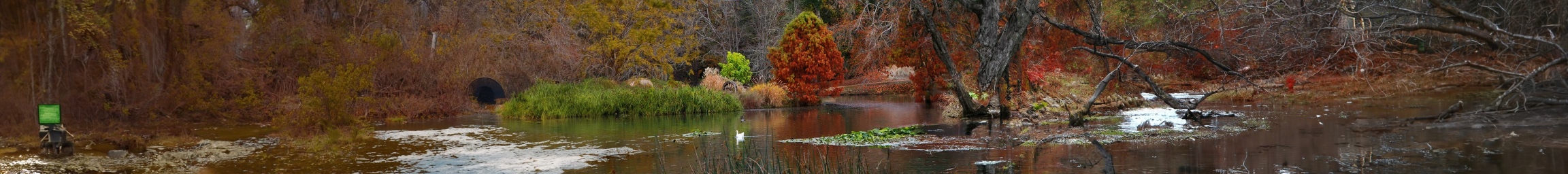}} 
\end{center}
\vspace{-.3cm}
\caption{Given the central 256$\times$256 pixels, we extrapolate to the left and right in steps of 128 pixels (2$\times$8 applications of 50\% \model  uncropping), to generate the final 256$\times$2304 panorama.
Figure D.3 in the Appendix shows more samples. \label{fig:panormauncrop}}
\vspace*{-.2cm}
\end{figure*}

Generative Adversarial Networks (GANs) \citep{goodfellow2014generative,radford2015unsupervised} have emerged as the model family of choice for many image-to-image tasks \citep{isola-cvpr-2017}; they are capable of generating high fidelity outputs, are broadly applicable, and support efficient sampling. 
Nevertheless, GANs can be challenging to train \cite{arjovsky-arxiv-2017,gulrajani2017improved}, and often drop modes in the output distribution~\cite{metz2016unrolled,ravuri2019classification}.
Autoregressive Models \citep{oord2016conditional,parmar2018image}, VAEs \citep{Kingma2013,vahdat2021nvae}, and Normalizing Flows \citep{dinh2016density,Kingma2018} have seen success in specific applications, but arguably, have not established the same level of quality and generality as GANs.  

Diffusion and score-based models \citep{sohl2015deep,song-arxiv-2020,ho2020denoising} have received a surge of recent interest~\cite{cai-eccv-2020,song-iclr-2021,hoogeboom2021argmax,vahdat2021score,kingma2021variational,austin2021structured}, resulting in several key advances in modeling continuous data.
On speech synthesis, diffusion models have achieved human evaluation scores on par with SoTA autoregressive models~\citep{chen-iclr-2021,chen-interspeech-2021,kong-arxiv-2020}.
On the class-conditional ImageNet generation challenge they have outperformed strong GAN baselines in terms of FID scores~\citep{dhariwal2021diffusion,ho-arxiv-2021}.
On image super-resolution, they have delivered impressive face enhancement results, outperforming GANs~\citep{saharia2021image}. Despite these results, it is not clear whether diffusion models rival GANs in offering a versatile and general framework for image manipulation.

This paper investigates the general applicability of {\em \model}, our implementation of image-to-image diffusion models, to a suite of  distinct and challenging tasks, namely colorization, inpainting, uncropping, and JPEG restoration
(see Figs.\ \ref{fig:1}, \ref{fig:panormauncrop}).
We show that 
\model, with no task-specific architecture customization, nor changes to hyper-parameters or the loss,  delivers high-fidelity outputs across all four tasks. It outperforms task-specific baselines and a strong regression baseline with an identical neural architecture.
Importantly, we show that a single {\em generalist} \model model, trained on colorization, inpainting and JPEG restoration, outperforms a task-specific
JPEG model and achieves competitive performance on the other tasks.

We study key components of \model, including the denoising loss function and the neural net architecture. We find that while $L_2$~\citep{ho2020denoising} and $L_1$~\citep{chen-iclr-2021} losses in the denoising objective yield similar sample-quality scores, 
$L_2$ leads to a higher degree of diversity in model samples, whereas $L_1$~\citep{chen-iclr-2021} produces more conservative outputs. 
We also find that removing self-attention layers from the U-Net architecture of \model, to build a fully convolutional model, hurts performance. 
Finally, we advocate a standardized evaluation protocol for inpainting, uncropping, and JPEG restoration based on ImageNet~\citep{deng2009imagenet}, and we report sample quality scores for several baselines. 
We hope this benchmark will help advance image-to-image translation research.
\section{Related work}

Our work is inspired by Pix2Pix \citep{isola-cvpr-2017}, which explored myriad image-to-image translation tasks with GANs. 
GAN-based techniques have also been proposed for image-to-image problems like unpaired translation \citep{zhu2017unpaired}, unsupervised cross-domain generation \citep{taigman2016unsupervised}, multi-domain translation \citep{choi2018stargan}, and few shot translation \citep{liu2019few}.
Nevertheless, existing GAN models are sometimes unsuccessful in holistically translating images with consistent structural and textural regularity.

Diffusion models  \citep{sohl2015deep} recently emerged with impressive results on image generation~\citep{ho2020denoising, ho-arxiv-2021, dhariwal2021diffusion}, audio synthesis~\citep{chen-iclr-2021, kong2020diffwave}, and image super-resolution~\citep{saharia2021image,Kadkhodaie2021}, as well as unpaired image-to-image translation~\citep{sasaki2021unit} and image editing~\citep{meng2021sdedit,sinha2021d2c}.
Our conditional diffusion models build on these recent advances, showing versatility on a suite of image-to-image translation tasks.

Most diffusion models for inpainting and other linear inverse problems have adapted unconditional models for use in conditional tasks \cite{sohl2015deep,song-iclr-2021,meng2021sdedit}.
This has the advantage that only one model need be trained.  However, unconditional tasks are often more difficult than conditional tasks.
We cast \model as a conditional model, opting for multitask training should one want a single model for multiple tasks.



Early \textbf{inpainting} approaches 
\cite{bertalmio2000image, barnes2009PAR, he2012statistics,hays2007scene}
work well on textured regions but often fall short in generating semantically consistent structure. 
GANs are widely used but often require auxiliary objectives on structures, context, edges, contours and hand-engineered features~\cite{iizuka2017globally, yu2018generative, yu2019free,nazeri2019edgeconnect, yi2020contextual, liu2020rethinking, kim2021zoomtoinpaint}, and they lack
diversity in their outputs  \citep{zheng2019pluralistic, zhao2021large}.

\textbf{Image uncropping} (a.k.a.\ outpainting) is considered more challenging than inpainting as it entails generating open-ended content with less context. 
Early methods 
relied on retrieval \citep{kopf2012quality, wang2014biggerpicture, shan2014uncrop}.
GAN-based methods are now predominant  \cite{teterwak2019boundless},
but are often domain-specific \cite{yang2019very, bowen2021oconet,wang2019srn, cheng2021out, lin2021infinitygan}. 
We show that conditional diffusion models trained on large  datasets reliably address both inpainting and uncropping across image domains.




{\bf Colorization} is a well-studied
task \cite{coltran,guadarrama2017pixcolor,royer-arxiv-2017,ardizzone-arxiv-2017}, requiring a degree of scene understanding, which makes it a natural choice for self-supervised learning~\citep{larsson2016learning}. 
Challenges include diverse colorization~\citep{deshpande2017learning}, respecting semantic categories~\citep{zhang2016colorful}, and producing high-fidelity color~\cite{guadarrama2017pixcolor}. 
While some prior work makes use of specialized auxiliary classification losses, we find that generic image-to-image diffusion models work well  without task-specific specialization.
\textbf{JPEG restoration} (aka. JPEG artifact removal) is the nonlinear inverse problem of removing compression artifacts. \cite{dong2015compression} applied deep CNN architectures for JPEG restoration, and \cite{galteri2017deep, galteri2019deep} successfully applied GANs  for artifact removal, but they have been restricted to quality factors above 10. 
We show the effectiveness of \model in removing compression artifacts for quality factors as low as 5. 

{\bf Multi-task training} is a relatively under-explored area in image-to-image translation.
\citep{guocheng2019trinity, yu2018crafting} train simultaneously on multiple  tasks, but they focus primarily on enhancement tasks like deblurring, denoising, and super-resolution, and they use smaller modular networks. Several works have also dealt with simultaneous training over multiple degradations on a single task \eg~multi-scale super-resolution \citep{kim2016deeply}, and
JPEG restoration on multiple quality factors \citep{galteri2019deep, liu2018multi}.
With \model we take a first step toward building multi-task image-to-image diffusion models for a wide variety of tasks.


\section{\model}

Diffusion models \citep{sohl2015deep,ho2020denoising} convert samples from a standard Gaussian distribution into samples from an empirical data distribution through an iterative denoising process.
Conditional diffusion models \citep{chen-iclr-2021,saharia2021image} make the denoising process conditional on an input 
signal. Image-to-image diffusion models are conditional diffusion models of the form $p(\vy \mid \vx)$, where both $\vx$ and $\vy$ are images, \eg~$\vx$ is a grayscale image and $\vy$ is a color image. These models have been applied to image super-resolution~\citep{saharia2021image, nichol2021improved}. We study 
the general applicability of image-to-image diffusion models on a broad set of tasks.

For a detailed treatment of diffusion models, please see Appendix A. Here, we briefly discuss the denoising loss function. Given a training output image $\vy$, we generate a noisy version $\vty$, and train a neural network $f_\theta$ to denoise $\vty$ given $\vx$ and a noise level indicator $\gamma$, for which the loss is
\begin{equation}
    \E_{(\vx, \vy)} \E_{\veps \sim \mathcal{N}(0, I)} \E_{\gamma}\, \bigg\lVert f_\theta(\vx,\, \underbrace{\sqrt{\gamma} \,\vy + \sqrt{1\!-\!\gamma}\, \veps}_{\vty}, \,\gamma) - \veps\, \bigg\rVert^{p}_p \, ,
\label{eq:main-loss}
\end{equation}
\cite{chen-iclr-2021} and \cite{saharia2021image} suggest using the $L_1$ norm, \ie~$p=1$, whereas 
the standard formulation is based on the usual $L_2$ norm \citep{ho2020denoising}. We perform careful ablations below, and analyze the impact of the choice of norm.
We find that $L_1$ yields significantly lower sample diversity compared to $L_2$. While $L_1$ may be useful, to reduce potential hallucinations in some applications,
here we adopt $L_2$ to capture the output distribution more faithfully.

\textbf{Architecture.}
\model uses a U-Net architecture \citep{ho2020denoising} with several  modifications inspired by recent work \citep{song-iclr-2021, saharia2021image, dhariwal2021diffusion}.
The network architecture is based on the 256$\times$256 class-conditional U-Net model of \cite{dhariwal2021diffusion}. The two main differences between our architecture and theirs are (i) absence of class-conditioning, and (ii) additional conditioning of the source image via concatenation, following \cite{saharia2021image}.

\begin{figure*}[t]
\setlength{\tabcolsep}{1pt}
\begin{center}
{\small 
\begin{tabular}{cccccc}
{\small Grayscale Input } & {\small PixColor\textsuperscript{\textdagger}}  & {\small ColTran\textsuperscript{\ddag} } & {\small Regression} &  {\small \model (Ours)} & {\small Original} 
\\
{\includegraphics[width=0.155\textwidth]{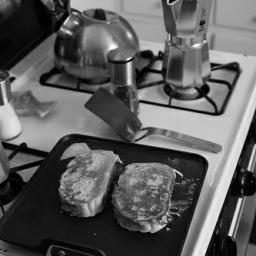}} &
{\includegraphics[width=0.155\textwidth]{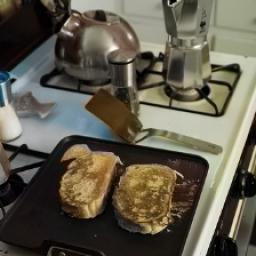}} &
{\includegraphics[width=0.155\textwidth]{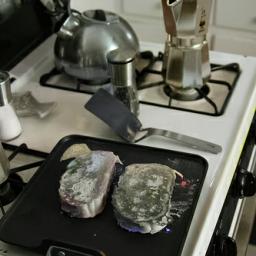}} &
{\includegraphics[width=0.155\textwidth]{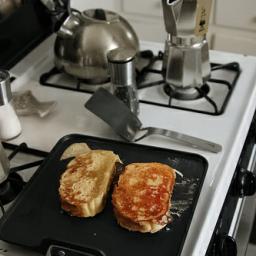}} &
{\includegraphics[width=0.155\textwidth]{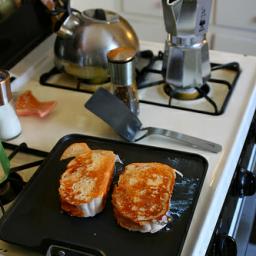}} &
{\includegraphics[width=0.155\textwidth]{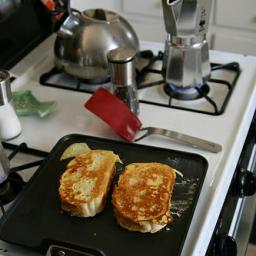}} \\

{\includegraphics[width=0.155\textwidth]{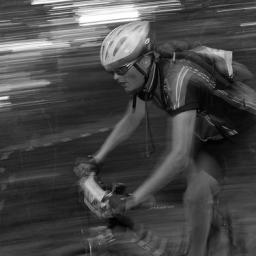}} &
{\includegraphics[width=0.155\textwidth]{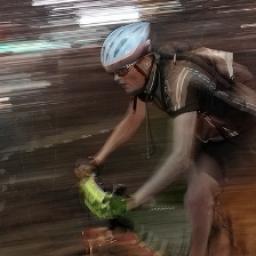}} &
{\includegraphics[width=0.155\textwidth]{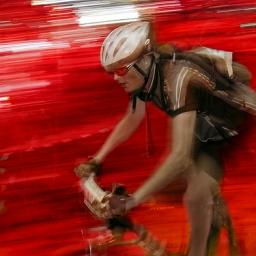}} &
{\includegraphics[width=0.155\textwidth]{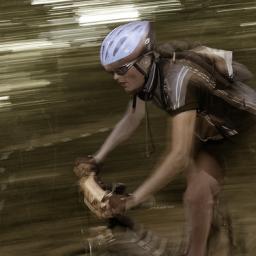}} &
{\includegraphics[width=0.155\textwidth]{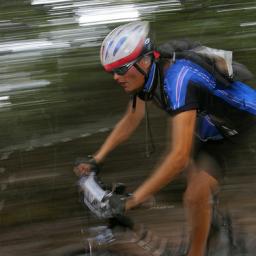}} &
{\includegraphics[width=0.155\textwidth]{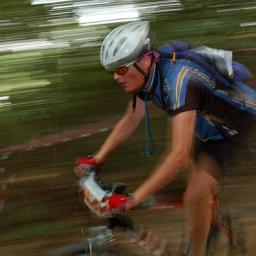}} \\

\end{tabular}
}
\end{center}
\vspace*{-0.3cm}
\caption{Colorization results on ImageNet validation images. Baselines: \textsuperscript{\textdagger}\citep{guadarrama2017pixcolor}, \textsuperscript{\ddag}\citep{coltran}, and our own strong regression baseline. Figure \ref{fig:colorization_comparison_appendix} shows more samples.
\label{fig:colorization_comparison}
}
\vspace*{-0.25cm}
\end{figure*}

\section{Evaluation protocol}

Evaluating image-to-image translation models is challenging.
Prior work on colorization \citep{zhang2016colorful, guadarrama2017pixcolor, coltran}  relied on FID scores and human evaluation for model comparison.
Tasks like inpainting \citep{yu2019free, yu2018generative} and uncropping \citep{teterwak2019boundless, wang2019wide} have often heavily relied on qualitative evaluation. For other tasks, like JPEG restoration \citep{dong2015compression,liu2018multi,galteri2019deep}, it has been common to use reference-based pixel-level similarity scores such as PSNR and SSIM. It is also notable that many tasks lack a standardized dataset for evaluation, \eg~different test sets with method-specific splits are used for evaluation.

We propose a unified evaluation protocol for inpainting, uncropping, and JPEG restoration on ImageNet \citep{deng2009imagenet}, due to its scale, diversity, and public availability. For inpainting and uncropping, existing work has relied on Places2 dataset \citep{zhou2017places} for evaluation. Hence, we also use a standard evaluation setup on Places2 for these tasks. Specifically, we advocate the use of ImageNet ctest10k split proposed by \cite{larsson2016learning} as a standard subset for benchmarking of all image-to-image translation tasks on ImageNet. We also introduce a similar category-balanced 10,950 image subset of Places2 validation set called \textit{places10k}. We further advocate the use of automated metrics that capture both image quality and diversity, in addition to controlled human evaluation. 
We avoid pixel-level metrics like PSNR and SSIM as they  are not reliable measures of sample quality for difficult tasks that require hallucination, like recent super-resolution work, where \citep{ledig2017photo, dahl2017pixel, menon2020pulse} observe that PSNR and SSIM
tend to prefer blurry regression outputs, unlike human perception.

We use four automated quantitative measures of sample quality for image-to-image translation: {\bf Inception Score (IS)}~\citep{salimans-iclr-2017}; {\bf Fréchet Inception Distance (FID)}; {\bf Classification Accuracy (CA)} (top-1) of a pre-trained ResNet-50 classifier; and  a simple measure of {\bf Perceptual Distance (PD)}, \ie~Euclidean distance in Inception-v1 feature space
(c.f., \cite{DosovitskiyBrox2016}). 
To facilitate benchmarking on our proposed subsets, we release our model outputs
together with other data such as the inpainting masks (see \url{https://bit.ly/eval-pix2pix}).
See Appendix C.5 for more details about 
our evaluation.
For some tasks, we also assess \textbf{sample diversity} through pairwise SSIM and LPIPS scores between multiple model outputs.
Sample diversity is challenging and has been a key limitation of many existing GAN-based methods \citep{zhu2017multimodal, yang2019diversity}.


The ultimate evaluation of image-to-image translation models is \textbf{human evaluation}; \ie~whether or not humans can discriminate model outputs from natural images.
To this end we use 2-alternative forced choice (2AFC) trials to evaluate the perceptual quality of model outputs against natural images from which we obtained test inputs ({\em c.f.,}~the Colorization Turing Test~\citep{zhang2016colorful}).
We summarize the results in terms of the \textbf{fool rate}, the percentage of human raters who select model outputs over natural images when they were asked ``Which image would you guess is from a camera?''.
(See Appendix C for details.)

\section{Experiments}

We apply \model to a suite of
challenging image-to-image  tasks:
\vspace{-.025cm}
\begin{enumerate}[topsep=0.1pt, partopsep=0pt, leftmargin=20pt, parsep=0pt, itemsep=0.1pt]
    \item \textbf{Colorization} transforms an input grayscale image to a plausible color image.
    \item \textbf{Inpainting} fills in user-specified masked regions of an image with realistic content. 
    \item \textbf{Uncropping} extends an input image along one or more directions to enlarge the image.
    \item \textbf{JPEG restoration} 
    corrects for JPEG compression artifacts, restoring plausible image detail.
\end{enumerate}
We do so without task-specific hyper-parameter tuning, architecture customization, or any auxiliary loss function.
Inputs and outputs for all tasks are represented as  256$\times$256 RGB images. Each task presents its own unique challenges.
Colorization entails a  representation of objects, segmentation and layout, with long-range image dependencies. Inpainting is  challenging with large masks, image diversity and cluttered scenes. Uncropping is widely considered even more challenging than inpainting as there is less surrounding  context to constrain semantically meaningful generation. While the other tasks are linear in nature, JPEG restoration is a non-linear inverse problem; it requires a good local model of natural image statistics to detect and correct compression artifacts.
While previous work has studied these problems extensively, it is rare that a model with no task-specific engineering achieves strong performance in all tasks, beating strong task-specific GAN and regression baselines. \model uses an $L_2$ loss for the denoising objective, unless otherwise specified. (Implementation details can be found in Appendix B.)

\subsection{Colorization}

While prior works \citep{zhang2016colorful, coltran} have adopted LAB or YCbCr color spaces to represent output images for colorization, we use the RGB color space to maintain generality across tasks. 
Preliminary experiments indicated that \model is equally effective in YCbCr and RGB spaces. We compare \model with  Pix2Pix~\citep{isola2017image}, PixColor~\citep{guadarrama2017pixcolor}, and ColTran~\citep{coltran}. 
Qualitative results are shown in Fig.\, \ref{fig:colorization_comparison}, with quantitative scores in Table \ref{tab:colorization_results}.
\model establishes a new SoTA, outperforming existing works by a large margin.
Further, the performance measures (FID, IS, and CA) indicate that 
\model outputs are close to being indistinguishable from the original images that were used to create the test greyscale inputs.
Surprisingly, our $L_2$ Regression baseline also outperforms prior task-specific techniques, highlighting the importance of modern architectures and large-scale training, even for a basic Regression model. On human evaluation, \model improves upon human raters' fool rate of ColTran by more than 10\%, approaching an ideal fool rate of 50\%.  

\setlength{\tabcolsep}{3pt}
\begin{table}[h]
    \centering
    {\small
    \begin{tabular}{lcccccc}
    \toprule
    \bfseries{Model} & \bfseries{FID-5K} $\downarrow$  & \bfseries{IS} $\uparrow$  & \bfseries{CA} $\uparrow$  & \bfseries{PD} $\downarrow$ & \bfseries{Fool rate} $\uparrow$  \\
    \midrule
    \textit{Prior Work} \\
    \quad pix2pix~\textsuperscript{\textdagger} & 24.41 & - & - & - & -\\
    \quad PixColor~\textsuperscript{\ddag}  & 24.32  & - & - & - & 29.90\%\\
    \quad Coltran~\textsuperscript{\textdagger\textdagger} & 19.37 & - & - & - & 36.55\%\\
    \textit{This paper} \\
    \quad Regression & 17.89 & 169.8 & 68.2\% & 60.0 & 39.45\% \\
    \quad  \model & \textbf{15.78} & \textbf{200.8} & \textbf{72.5\%} & \textbf{46.2} & \textbf{47.80\%} \\
    \midrule
    Original images & 14.68 & 229.6 & 75.6\% & 0.0 & - \\
    \bottomrule
    \end{tabular}
    }
    \vspace*{0.1cm}
    \caption{Colorization quantitative scores and fool rates on ImageNet val set indicate that \model outputs are bridging the gap to being indistinguishable from the original images from which the greyscale inputs were created. Baselines: \textsuperscript{\textdagger}\citep{isola2017image}, \textsuperscript{\ddag}\citep{guadarrama2017pixcolor} and \textsuperscript{\textdagger\textdagger}\citep{coltran}. Appendix C.1 provides more results.
    \label{tab:colorization_results}
    \vspace*{-0.6cm}
    }
\end{table}

\subsection{Inpainting}

We follow \cite{yu2019free} and train inpainting models on free-form generated masks, augmented with simple rectangular masks. To maintain generality of \model across tasks, in contrast to prior work, we do not pass a binary inpainting mask to the models. Instead, we fill the masked region with standard Gaussian noise, which is compatible with denoising diffusion models. The training loss only considers the masked out pixels, rather than the entire image, to speed up training. We compare \model with DeepFillv2 \citep{yu2019free}, HiFill \citep{yi2020contextual}, Photoshop's \textit{Content-aware Fill}, and Co-ModGAN \citep{zhao2021large}. While there are other important prior works on image inpainting, such as \citep{liu2018image, liu2020rethinking, zheng2019pluralistic}, we were not able to compare with all of them.

\begin{figure*}[t!]
\setlength{\tabcolsep}{1pt}
\begin{center}
{\small 
\begin{tabular}{cccccc}
{\small Masked Input } &{\small Photoshop 2021\textsuperscript{\ddag}} & {\small DeepFillv2\textsuperscript{\textdagger}} & {\small HiFill\textsuperscript{\textdagger\textdagger}} & {\small Co-ModGAN\textsuperscript{\ddag\ddag}} & {\small \model (Ours)} 
\\
{\includegraphics[width=0.15\textwidth]{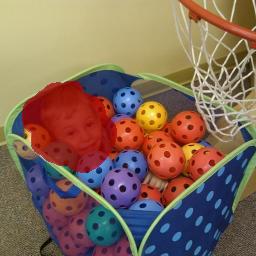}} &
{\includegraphics[width=0.15\textwidth]{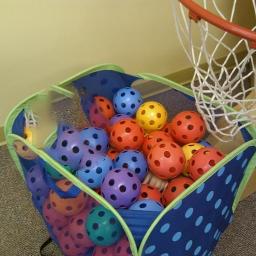}} &
{\includegraphics[width=0.15\textwidth]{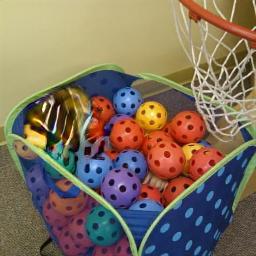}} &
{\includegraphics[width=0.15\textwidth]{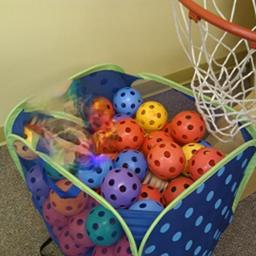}} &
{\includegraphics[width=0.15\textwidth]{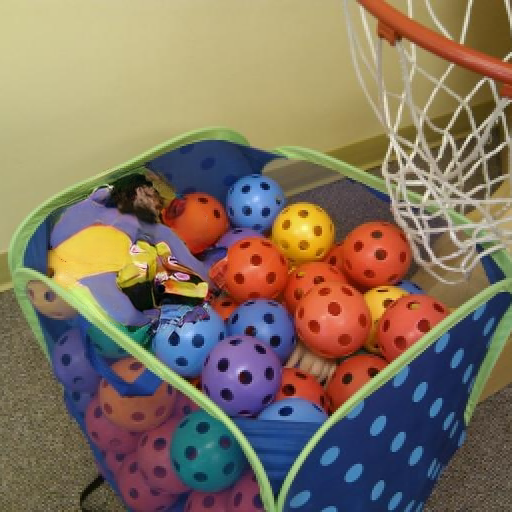}} &
{\includegraphics[width=0.15\textwidth]{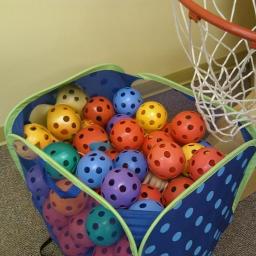}} \\

{\includegraphics[width=0.15\textwidth]{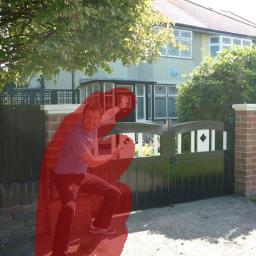}} &
{\includegraphics[width=0.15\textwidth]{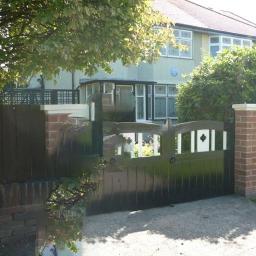}} &
{\includegraphics[width=0.15\textwidth]{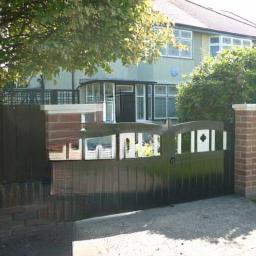}} &
{\includegraphics[width=0.15\textwidth]{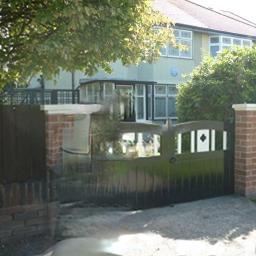}} &
{\includegraphics[width=0.15\textwidth]{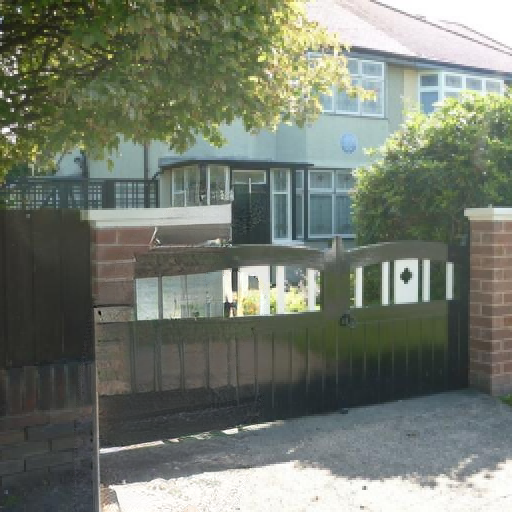}} &
{\includegraphics[width=0.15\textwidth]{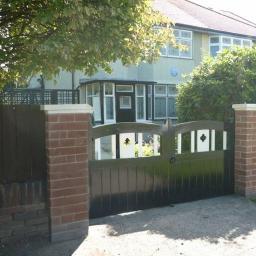}} \\

{\includegraphics[width=0.15\textwidth]{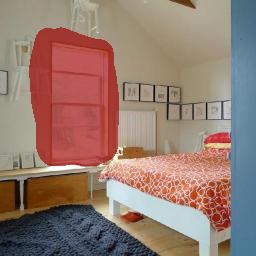}} &
{\includegraphics[width=0.15\textwidth]{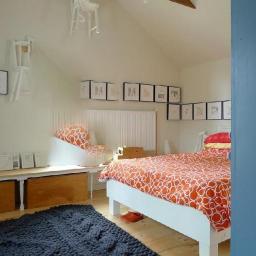}} &
{\includegraphics[width=0.15\textwidth]{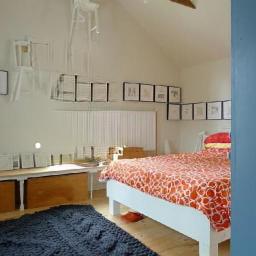}} &
{\includegraphics[width=0.15\textwidth]{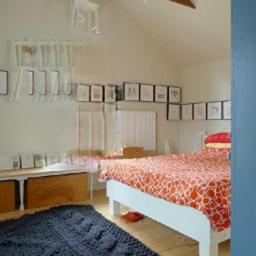}} &
{\includegraphics[width=0.15\textwidth]{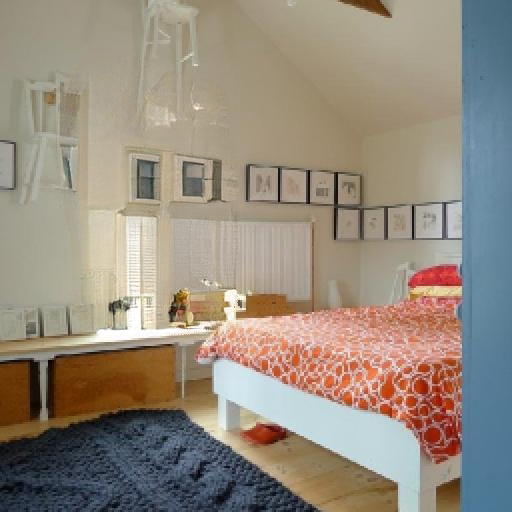}} &
{\includegraphics[width=0.15\textwidth]{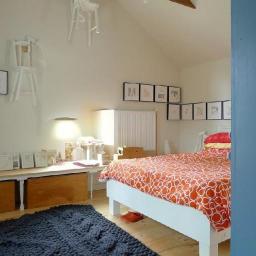}}

\end{tabular}
}
\end{center}
\vspace*{-0.3cm}
\caption{Comparison of inpainting methods on object removal. Baselines: \textsuperscript{\ddag}Photoshop's \textit{Content-aware Fill} built on PatchMatch \citep{barnes2009PAR}, \textsuperscript{\textdagger}\citep{yu2019free}, \textsuperscript{\textdagger\textdagger}\citep{yi2020contextual} and \textsuperscript{\ddag\ddag}\citep{zhao2021large}.  
See Figure C.5 in Appendix for more samples.
\label{fig:inpainting_comparison}
}
\end{figure*}

Qualitative and quantiative results are given in Fig.\ \ref{fig:inpainting_comparison} and Table \ref{tab:inpainting_results}. \model exhibits strong performance across inpainting datasets and mask configurations, outperforming DeepFillv2, HiFill and Co-ModGAN by a large margin. 
Importantly, like the colorization task above, the FID scores for \model outputs in the case of  20-30\% free-form masks, are extremely close to FID scores on the original images from which we created the masked test inputs. 
See Appendix C.2 for more results.

\begin{figure}[t]
\setlength{\tabcolsep}{1.5pt}
\begin{center}
\begin{tabular}{cccccc}
{\small Masked Input} &
{\small Boundless\textsuperscript{\textdagger}} &{\small InfinityGAN \textsuperscript{\textdagger\textdagger}} & {\small \model (Ours)} 
\\
\frame{\includegraphics[width=0.24\linewidth]{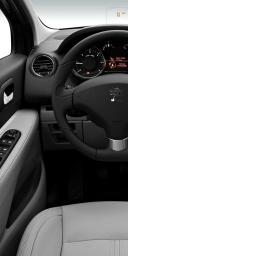}} &
\frame{\includegraphics[width=0.24\linewidth]{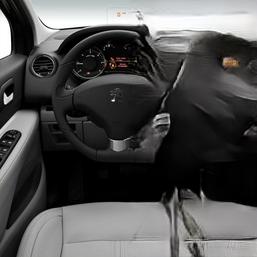}} &
\frame{\includegraphics[width=0.24\linewidth]{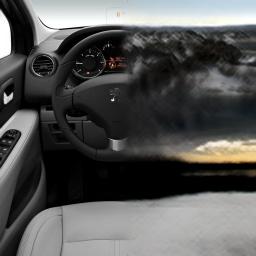}} &
\frame{\includegraphics[width=0.24\linewidth]{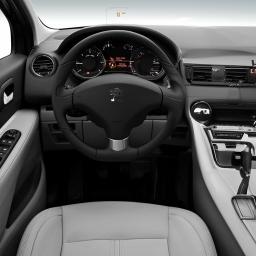}} 
\\

\frame{\includegraphics[width=0.24\linewidth]{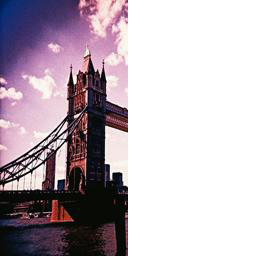}} &
\frame{\includegraphics[width=0.24\linewidth]{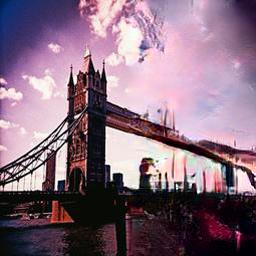}} &
\frame{\includegraphics[width=0.24\linewidth]{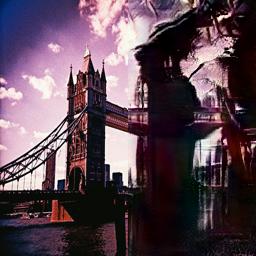}} &
\frame{\includegraphics[width=0.24\linewidth]{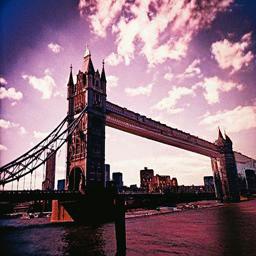}}
\\

\frame{\includegraphics[width=0.24\linewidth]{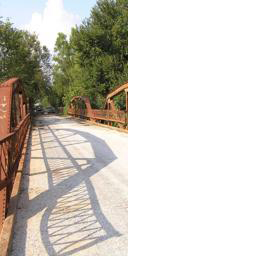}} &
\frame{\includegraphics[width=0.24\linewidth]{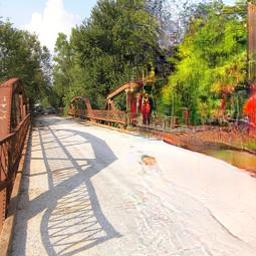}} &
\frame{\includegraphics[width=0.24\linewidth]{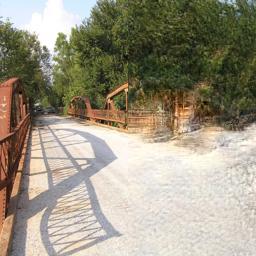}} &
\frame{\includegraphics[width=0.24\linewidth]{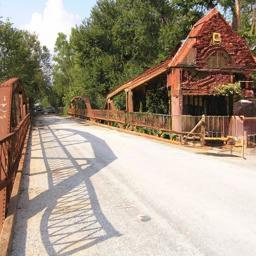}}
\\

\frame{\includegraphics[width=0.24\linewidth]{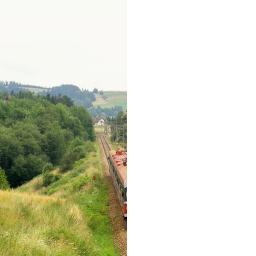}} &
\frame{\includegraphics[width=0.24\linewidth]{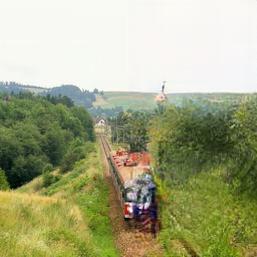}} &
\frame{\includegraphics[width=0.24\linewidth]{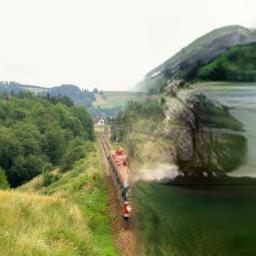}} &
\frame{\includegraphics[width=0.24\linewidth]{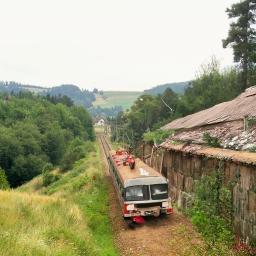}}
\\

\end{tabular}
\end{center}
\vspace*{-0.3cm}
\caption{Image uncropping results on Places2 validation images. Baselines:  Boundless\textsuperscript{\textdagger} \citep{teterwak2019boundless} and  InfinityGAN\textsuperscript{\textdagger\textdagger} \citep{lin2021infinitygan} trained on a scenery subset of Places2. 
Figure C.8 in the Appendix shows more samples.
\label{fig:extrapolation_comparison}
}
\vspace*{-0.4cm}
\end{figure}

\begin{figure}[t]
\setlength{\tabcolsep}{1.5pt}
\begin{center}
\begin{tabular}{cccc}
{\small Input (QF=5)} & {\small Regression}  & {\small \model (Ours)} & {\small Original} 
\\

{\includegraphics[width=0.24\linewidth]{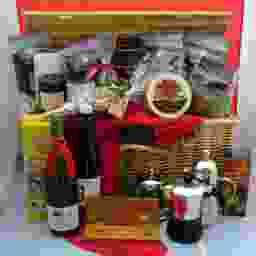}} &
{\includegraphics[width=0.24\linewidth]{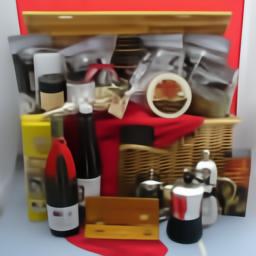}} &
{\includegraphics[width=0.24\linewidth]{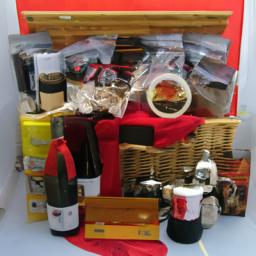}} &
{\includegraphics[width=0.24\linewidth]{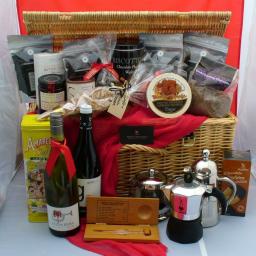}} \\

{\includegraphics[width=0.24\linewidth]{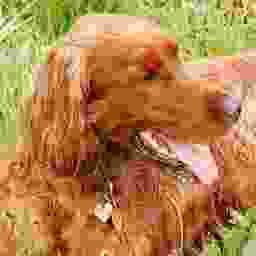}} &
{\includegraphics[width=0.24\linewidth]{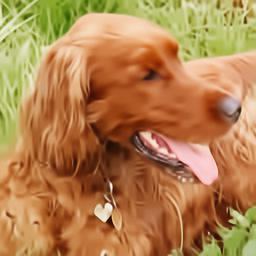}} &
{\includegraphics[width=0.24\linewidth]{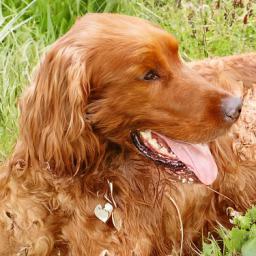}} &
{\includegraphics[width=0.24\linewidth]{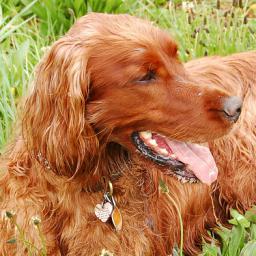}} \\

\end{tabular}
\end{center}
\vspace*{-0.3cm}
\caption{Example of JPEG restoration results.  Figure D.1 in the Appendix shows more samples.
\label{fig:jpeg_comparison}
}
\vspace*{-0.4cm}
\end{figure}

\setlength{\tabcolsep}{1pt}
\begin{table}[t]
    \centering
    {\small
    \begin{tabular}{lccccccc}
    \toprule
    \bfseries{Model} & \multicolumn{4}{c}{\bfseries{ImageNet}} & & \multicolumn{2}{c}{\bfseries{Places2}} \\
    \cmidrule(lr){2-5}
    \cmidrule(lr){7-8}
    & \bfseries{FID}  $\downarrow$ & \bfseries{IS}  $\uparrow$ & \bfseries{CA} $\uparrow$ & \bfseries{PD} $\downarrow$ & & \bfseries{FID} $\downarrow$ & \bfseries{PD} $\downarrow$ \\
    \midrule
    \textit{20-30\% free form} & & & & & & & \\
    \addlinespace[0.1cm]

    DeepFillv2 \citep{yu2019free} & 9.4 & 174.6 & 68.8\% & 64.7 & & 13.5 & 63.0 \\
    HiFill \citep{yi2020contextual} & 12.4 & 157.0 & 65.7\% & 86.2 & & 15.7 & 92.8  \\
    Co-ModGAN \citep{zhao2021large} & - & - & - & - & & 12.4 & 51.6  \\
    \model (Ours) & \textbf{5.2} & \textbf{205.5} & \textbf{72.3\%} & \textbf{27.6} & & \textbf{11.7} & \textbf{35.0}  \\
    \midrule
    \textit{128$\times$128 center} & & & & & & & \\
    \addlinespace[0.1cm]
    
    DeepFillv2 \citep{yu2019free} & 18.0 & 135.3 & 64.3\% & 117.2 & & 15.3 & 96.3 \\
    HiFill \citep{yi2020contextual} & 20.1 & 126.8 & 62.3\% & 129.7 & &  16.9 & 115.4  \\
    Co-ModGAN \citep{zhao2021large} & - & - & - & - & & 13.7 & 86.2 \\
    \model (Ours) & \textbf{6.6} & \textbf{173.9} &  \textbf{69.3\%} & \textbf{59.5} & & \textbf{11.9} & \textbf{57.3}  \\
    \midrule
    Original images & 5.1 & 231.6 & 74.6\% & 0.0 & &  11.4 & 0.0 \\
    \bottomrule
    \end{tabular}
    }
    \vspace*{0.1cm}
    \caption{Quantitative evaluation for free-form and center inpainting on ImageNet and Places2 validation images. 
    \label{tab:inpainting_results}
    }
    \vspace{-0.6cm}
\end{table}




\subsection{Uncropping} 



Recent works \citep{teterwak2019boundless,lin2021infinitygan} have shown impressive visual effects by extending (extrapolating) input images along the right border.
We train \model on uncropping in any one of the four directions, or around the entire image border on all four sides. In all cases, we keep the area of the masked region at 50\% of the image. Like inpainting, we fill the masked region with Gaussian noise, and keep the unmasked region fixed during inference. We compare \model with  Boundless~\citep{teterwak2019boundless} and InfinityGAN~\citep{lin2021infinitygan}.
While other uncropping methods exist (e.g., \citep{guo2020spiral, wang2019wide}), 
we only compare with  two representative methods. 
From the results in Fig.\ \ref{fig:extrapolation_comparison} and Table \ref{tab:extrapolation_results}, one can see that \model outperforms baselines on  ImageNet and Places2 by a large margin.
On human evaluation, \model has a 40\% fool rate, compared to 25\% and 15\% for Boundless and InfinityGAN (see Fig. C.2 in the Appendix for details).

We further assess the robustness of \model by generating panoramas through repeated application of left and right uncropping (see Fig.\ \ref{fig:panormauncrop}).
We observe that \model is surprisingly robust, generating realistic and coherent outputs even after 8 repeated applications of uncrop. 
We also generate zoom-out sequences by repeated uncropping around the entire border of the image with similarly appealing results (\url{https://diffusion-palette.github.io/}).

\begin{table}[h]
    \centering
    {\small
    \begin{tabular}{lccccccc}
    \toprule
    \bfseries{Model} & \multicolumn{4}{c}{\bfseries{ImageNet}} & & \multicolumn{2}{c}{\bfseries{Places2}}  \\
    \cmidrule(lr){2-5}
    \cmidrule(lr){7-8}
    & \bfseries{FID} $\downarrow$ & \bfseries{IS} $\uparrow$ & \bfseries{CA} $\uparrow$ & \bfseries{PD} $\downarrow$ &  & \bfseries{FID} $\downarrow$ & \bfseries{PD} $\downarrow$\\
    \midrule
    Boundless \citep{teterwak2019boundless} & 18.7 & 104.1 & 58.8\% & 127.9 & & 11.8 & 129.3  \\
    \model (Ours) & \textbf{5.8} & \textbf{138.1} & \textbf{63.4\%} & \textbf{85.9} & & \textbf{3.53} & \textbf{103.3}\\
    \midrule
    Original images & 2.7 & 250.1 & 76.0\% & 0.0 & &  2.1 & 0.0\\
    \bottomrule
    \end{tabular}
    }
    \vspace*{0.1cm}
    \caption{Quantitative scores and human raters' fool rates on uncropping. 
    Appendix C.3 provides more results.}
    \label{tab:extrapolation_results}
    \vspace*{-0.7cm}
\end{table}




\subsection{JPEG restoration}

Finally, we evaluate \model on the task of removing JPEG compression artifacts, a long standing image restoration problem \citep{dong2015compression, galteri2019deep, liu2018multi}. 
Like prior work \citep{ehrlich2020quantization, liu2018multi}, we train \model on inputs compressed with various quality factors (QF). While prior work has typically limited itself to a Quality Factor $\geq$ 10, we increase the difficulty of the task and train on Quality Factors as low as 5, producing severe compression artifacts. 
Table \ref{tab:compression_results} summarizes the 
ImageNet results, with \model exhibiting strong performance across all quality factors, outperforming
the regression baseline. 
As expected, the performance gap between \model and the  regression baseline widens with decreasing quality factor. Figure \ref{fig:jpeg_comparison} shows the qualitative comparison between \model and our Regression baseline at a quality factor of 5. It is easy to see that the regression model produces blurry outputs, while \model produces sharper images. 

\begin{table}[t]
    \centering
    {\small 
    \begin{tabular}{llcccc}
    \toprule
    \bfseries{QF} & \bfseries{Model} & \bfseries{FID-5K} $\downarrow$  & \bfseries{IS} $\uparrow$  & \bfseries{CA} $\uparrow$  & \bfseries{PD} $\downarrow$  \\
    \midrule
    \multirow{2}{*}{5} & Regression  & 29.0 & 73.9 & 52.8\% & 155.4 \\
    & \model (Ours) & \textbf{8.3} & \textbf{133.6}  & \textbf{64.2\%} & \textbf{95.5} \\
    \midrule
    \multirow{2}{*}{10} & Regression  & 18.0 & 117.2 & 63.5\% & 102.2 \\
    & \model (Ours)  & \textbf{5.4} & \textbf{180.5} & \textbf{70.7\%} & \textbf{58.3} \\
    \midrule
    \multirow{2}{*}{20} & Regression  & 11.5 & 158.7 & 69.7\% & 65.4 \\
    & \model (Ours) & \textbf{4.3} & \textbf{208.7} & \textbf{73.5\%} & \textbf{37.1} \\
    \midrule
    & Original images & 2.7 & 250.1 & 76.0\% & 0.0 \\
    \bottomrule
    \end{tabular}
    }
    \vspace{0.1cm}
    \caption{Quantitative evaluation for JPEG restoration for various Quality Factors (QF).}
    \label{tab:compression_results}
    \vspace{-0.5cm}
\end{table}

\subsection{Self-attention in diffusion model architectures}

Self-attention layers \citep{vaswani-nips-2017} have been an important component in recent U-Net architectures for diffusion models \citep{ho2020denoising, dhariwal2021diffusion}. 
While self-attention layers provide a direct form of global dependency, they prevent generalization to unseen image resolutions. Generalization to new resolutions at test time is convenient for many image-to-image  tasks, and therefore 
previous works have  relied primarily on fully convolutional architectures \citep{yu2019free, galteri2019deep}.

\begin{figure*}[t]
\setlength{\tabcolsep}{1pt}
\begin{center}
{\small 
\begin{tabular}{c@{\hspace{.15cm}}cccccc}
& {\small Input} & {\small Sample 1}  & {\small Sample 2} & {\small Sample 3} &  {\small Sample 4} 
\\
\raisebox{.8cm}{\rotatebox{90}{Inpainting}} &
\frame{\includegraphics[width=0.143\linewidth]{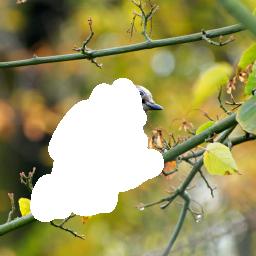}} \ &
\frame{\includegraphics[width=0.143\linewidth]{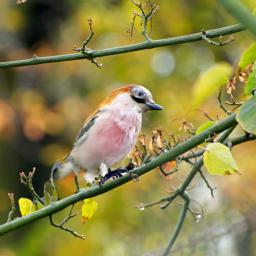}} \ &
\frame{\includegraphics[width=0.143\linewidth]{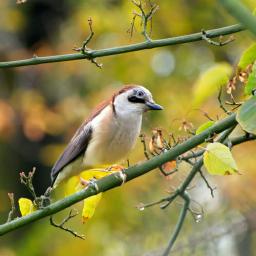}} \ &
\frame{\includegraphics[width=0.143\linewidth]{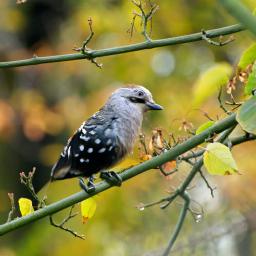}}\  &
\frame{\includegraphics[width=0.143\linewidth]{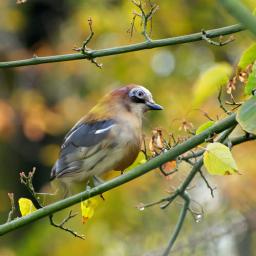}} \ &
\\

\raisebox{.6cm}{\rotatebox{90}{Colorization}} &
\frame{\includegraphics[width=0.143\linewidth]{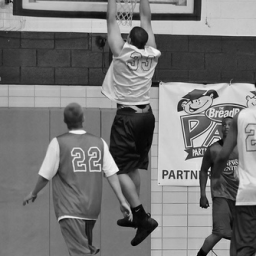}} &
\frame{\includegraphics[width=0.143\linewidth]{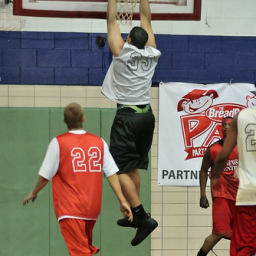}} &
\frame{\includegraphics[width=0.143\linewidth]{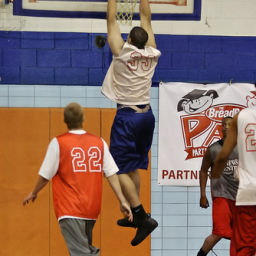}} &
\frame{\includegraphics[width=0.143\linewidth]{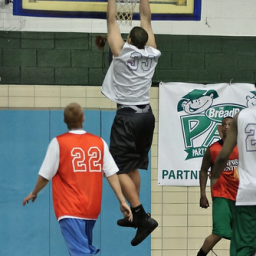}} &
\frame{\includegraphics[width=0.143\linewidth]{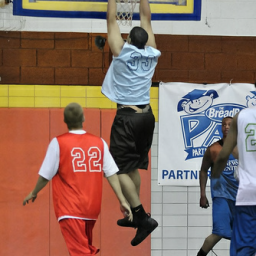}} &
\\


\raisebox{.7cm}{\rotatebox{90}{Uncropping}} &
\frame{\includegraphics[width=0.143\linewidth]{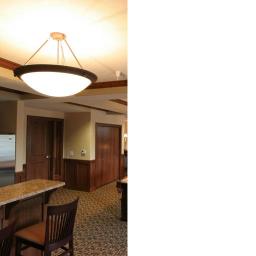}} &
\frame{\includegraphics[width=0.143\linewidth]{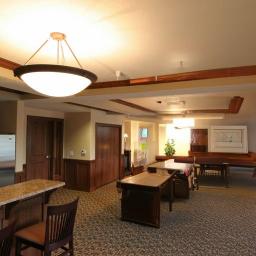}} &
\frame{\includegraphics[width=0.143\linewidth]{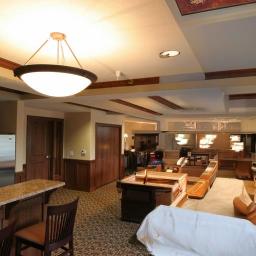}} &
\frame{\includegraphics[width=0.143\linewidth]{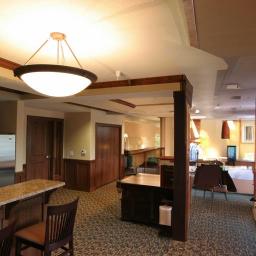}} &
\frame{\includegraphics[width=0.143\linewidth]{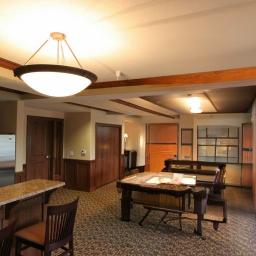}} &
\\

\end{tabular}
}
\end{center}
\vspace*{-0.3cm}
\caption{\model diversity for inpainting, colorization, and uncropping. Figures  C.4, C.6, C.9 and C.10 in the Appendix show more samples.
\label{fig:colorization_diversity}
}
\end{figure*}

We analyze the impact of these self-attention layers on sample quality for inpainting, one of the more difficult image-to-image translation tasks. In order to enable input resolution generalization for \model, we explore replacing global self-attention layers with different alternatives each of which represents a trade-off between large context dependency, and resolution robustness. In particular, we experiment with the following four configurations: 
\begin{enumerate}[topsep=0.21pt, partopsep=0pt, leftmargin=13pt, parsep=0pt, itemsep=0.1pt]
    \item \textbf{Global Self-Attention}: Baseline configuration with global self-attention layers at 32$\times$32, 16$\times$16 and 8$\times$8 resolutions. 
    \item \textbf{Local Self-Attention}: Local self-attention layers \citep{vaswani2021scaling} at 32$\times$32, 16$\times$16 and 8$\times$8 resolutions, at which feature maps are divided into 4 non-overlapping query blocks.
    \item \textbf{More ResNet Blocks w/o Self-Attention}: 2 $\times$ residual blocks at 32$\times$32, 16$\times$16 and 8$\times$8 resolutions allowing deeper convolutions to increase receptive field sizes.
    \item \textbf{Dilated Convolutions w/o Self-Attention}: Similar to 3. ResNet blocks at 32$\times$32, 16$\times$16 and 8$\times$8 resolutions with increasing dilation rates \citep{chen2017rethinking} allowing exponentially increasing receptive fields.
\end{enumerate}

We train models for 500K steps, with a batch size of 512. Table \ref{tab:architecture ablations} reports the performance of different configurations for inpainting. Global self-attention offers  better performance than fully-convolutional alternatives (even with $15\%$ more parameters), re-affirming the importance of self-attention layers for such tasks.  Surprisingly, local self-attention performs worse than fully-convolutional alternatives. 
Sampling speed is slower than GAN models.
There is a large overhead for loading models and the initial  jit compilation,
but for 1000 test images, \model requires 0.8 sec./image on a TPUv4.

\begin{table}[t]
    \centering
    {\small 
    \begin{tabular}{lccccc}
    \toprule
    \bfseries{Model} & \bfseries{\# Params} & \bfseries{FID} $\downarrow$  & \bfseries{IS} $\uparrow$  & \bfseries{PD} $\downarrow$  \\
    \midrule
    \textit{Fully Convolutional} & & & \\
    \quad Dilated Convolutions  & 624M & 8.0 & 157.5 & 70.6 \\
    \quad More ResNet Blocks & 603M & 8.1 & 157.1 & 71.9  \\
    \midrule
    \textit{Self-Attention} & & & \\
    \quad Local Self-Attention & 552M & 9.4 & 149.8 & 78.2 \\
    \quad Global Self-Attention  & 552M & \textbf{7.4}& \textbf{164.8} & \textbf{67.1} \\
    \bottomrule
    \end{tabular}
    }
    \vspace*{0.1cm}
    \caption{Architecture ablation for inpainting.}
    \label{tab:architecture ablations}
    \vspace*{-0.7cm}
\end{table}

\begin{table}[t]
    \centering
    {\small
    \begin{tabular}{lccccccc}
    \toprule
     & \multicolumn{3}{c}{\bfseries{Inpainting}} & \hspace*{0.4cm} & \multicolumn{3}{c}{\bfseries{Colorization}} \\
     \cmidrule(lr){2-4}
     \cmidrule(lr){6-8}
    \bfseries{Model} & \bfseries{FID} $\downarrow$ & \bfseries{PD} $\downarrow$ & \bfseries{LPIPS} $\uparrow$ & & \bfseries{FID} $\downarrow$ & \bfseries{PD} $\downarrow$ & \bfseries{LPIPS} $\uparrow$ \\
    \midrule
    Diffusion $L_1$ & 3.6 & \textbf{41.9} & 0.11 & &  3.4 & \textbf{45.8} & 0.09 \\
    Diffusion $L_2$ & 3.6 & 43.8 & \textbf{0.13} & &  3.4 & 48.0 & \textbf{0.15} \\
    \bottomrule
    \end{tabular}
    }
    \vspace*{0.1cm}
    \caption{Comparison of $L_p$ norm in denoising objective.}
    \label{tab:diversity_table}
    \vspace*{-0.7cm}    
\end{table}

\subsection{Sample diversity}

\begin{figure}[h]
\small
\vspace*{-0.1cm}
\begin{center}
    
\setlength{\tabcolsep}{2pt}
\begin{tabular}{c}
{\includegraphics[width=.19\textwidth]{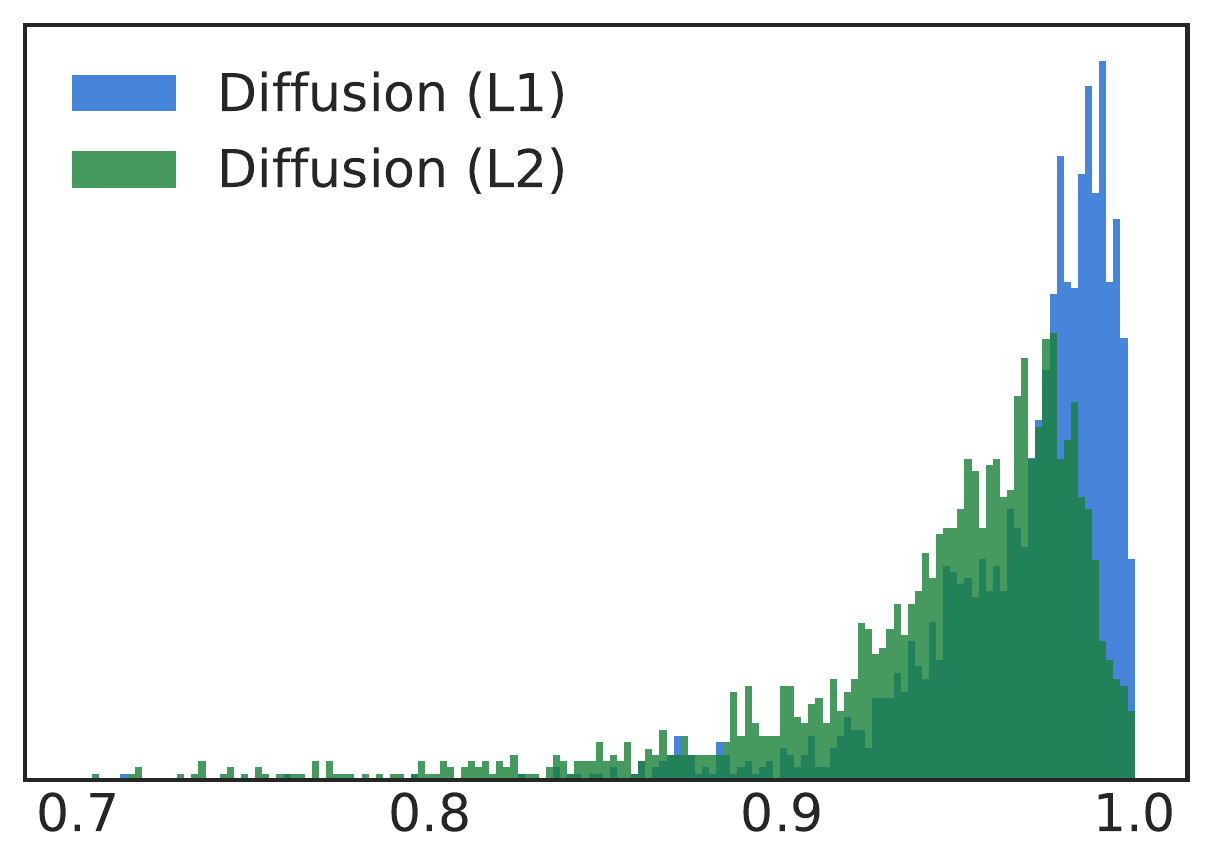}} \ 
{\includegraphics[width=.19\textwidth]{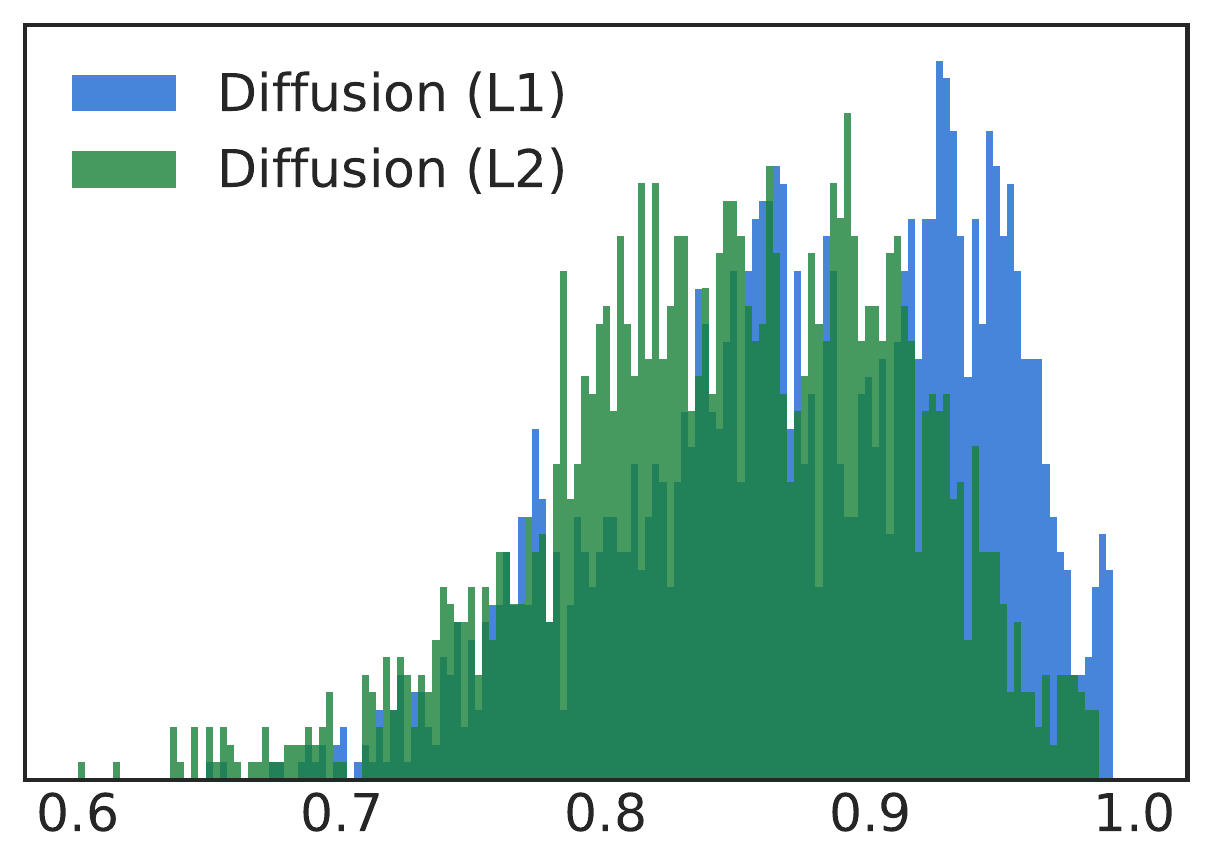}}
\end{tabular}
    
\end{center}
\vspace*{-0.45cm}
\caption{Pairwise multi-scale SSIM for colorization (left) and inpainting (right).
\label{fig:diversity_ssim}
}
\end{figure}

We next analyze sample diversity of \model on two tasks, colorization and inpainting. Specifically, we analyze the impact of the changing the diffusion loss function $L_{simple}$ \citep{ho2020denoising}, and compare $L_1$ vs. $L_2$ on sample diversity. While existing conditional diffusion models, SR3 \citep{saharia2021image} and WaveGrad \citep{chen-iclr-2021}, have found $L_{1}$ norm to perform better than the conventional $L_{2}$ loss, there has not been a detailed comparison of the two.
To quantitatively compare sample diversity, we use multi-scale SSIM \cite{guadarrama2017pixcolor} and the  LPIPS diversity score \cite{zhu2017multimodal}. Given multiple generated outputs for each input image, we compute pairwise multi-scale SSIM between the first output sample and the remaining samples. We do this for multiple input images, and then plot the histogram of SSIM values (see Fig.\ \ref{fig:diversity_ssim}). 
Following  \cite{zhu2017multimodal}, we also compute LPIPS scores between consecutive pairs of model outputs for a given input image, and then average  across all outputs and input images. 
Lower SSIM and higher LPIPS scores imply more sample diversity. 
The results in Table \ref{tab:diversity_table} thus clearly show that models trained with the $L_2$ loss have greater sample diversity than those trained with the $L_1$ loss.

Interestingly,  Table \ref{tab:diversity_table} also indicates that $L_1$ and $L_2$ models yield similar FID scores (i.e.,  comparable perceptual quality), but 
that $L_1$ has somewhat lower  Perceptual Distance scores than $L_2$.
One can speculate that $L_1$ models may drop more modes than $L_2$ models, thereby increasing the likelihood that  a single sample from an $L_1$ model is from the mode containing the corresponding original image, and hence a smaller Perceptual Distance.

Some existing GAN-based models explicitly encourage diversity;  \citep{zhu2017multimodal, yang2019diversity} propose methods for improving diversity of conditional GANs, and \citep{zhao2020uctgan, han2019finet} explore diverse sample generation for image inpainting. 
We leave comparison of sample diversity between \model and other such GAN based techniques to future work.

\subsection{Multi-task learning}

\begin{figure}[t]
\vspace*{-0.15cm}
\setlength{\tabcolsep}{1.5pt}
\begin{center}
\begin{tabular}{cccc}
{\small Input} & {\small Unconditional}  & {\small Multi-Task} & {\small Task-Specific} \\

{\includegraphics[width=0.24\linewidth]{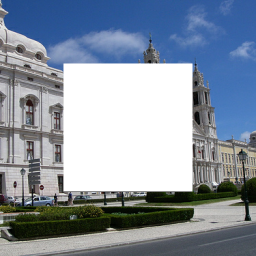}} &
{\includegraphics[width=0.24\linewidth]{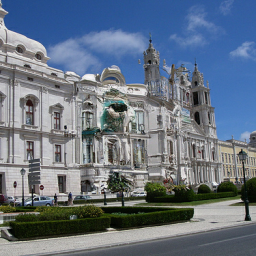}} &
{\includegraphics[width=0.24\linewidth]{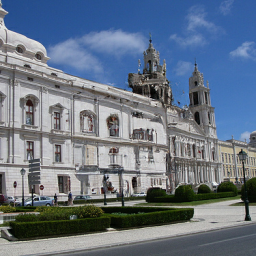}} &
{\includegraphics[width=0.24\linewidth]{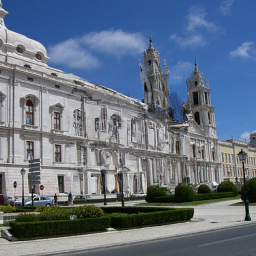}} \\

{\includegraphics[width=0.24\linewidth]{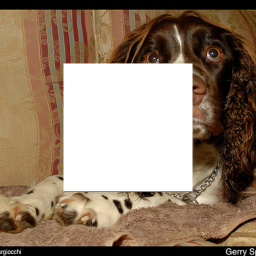}} &
{\includegraphics[width=0.24\linewidth]{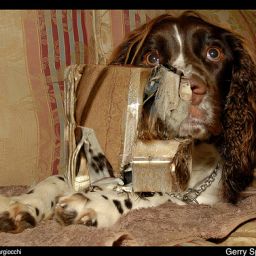}} &
{\includegraphics[width=0.24\linewidth]{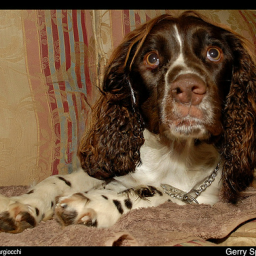}} &
{\includegraphics[width=0.24\linewidth]{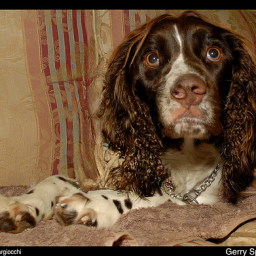}} \\

\end{tabular}
\end{center}
\vspace*{0.1cm}
\caption{Comparison of conditional and unconditional diffusion models for inpainting. 
Fig. C.7 in the Appendix
shows more results.
\label{fig:uncond_inpaint_main_paper}
}
\vspace*{-0.1cm}
\end{figure}

Multi-task training is a natural approach to learning a single model for multiple image-to-image tasks, i.e., blind image enhancement.
Another is to adapt an unconditional model to conditional tasks with imputation.
For example, \cite{song-iclr-2021} do this for inpainting; in each step of 
iterative refinement, they denoise the noisy image from the previous step, and then simply replace any pixels in the estimated image $\vy$ with pixels from the observed image regions, then adding noise and proceeding to the next denoising iteration.
Figure \ref{fig:uncond_inpaint_main_paper} compares this method  with a multi-task \model model trained on all four tasks, and 
a \model model trained solely on inpainting.  All models use the same architecture, training data and number of training steps.
The results in Fig.\ \ref{fig:uncond_inpaint_main_paper} are typical; the re-purposed unconditional model does not perform well, in part because it is hard to learn a good unconditional model 
on diverse datasets like ImageNet, and also because, during iterative refinement, noise
is added to all pixels, including the observed pixels.  By contrast, \model is condition directly on noiseless observations for all steps.

To explore the potential for multi-task models in greater depth, Table \ref{tab:multitask_results}
provides a quantitative comparison between a single generalist \model model trained 
simultaneously on JPEG restoration, inpainting, and colorization.
It indicates that multi-task generalist \model outperforms the task-specific JPEG restoration specialist model, but slightly lags behind task-specific \model models on inpainting and colorization. The 
multi-task and task-specific \model models had the same number of training steps; we expect multi-task  performance to improve with more training.

\begin{table}[t]
    \centering
    {\small
    \begin{tabular}{lcccc}
    \toprule
    \bfseries{Model} & \bfseries{FID} $\downarrow$ & \bfseries{IS} $\uparrow$ & \bfseries{CA} $\uparrow$ & \bfseries{PD} $\downarrow$ \\
    \midrule
    \textit{Inpainting (128$\times$128 center mask)} & & & & \\
    \quad \model \textit{(Task-specific)} & \textbf{6.6} & \textbf{173.9} &  \textbf{69.3\%} & \textbf{59.5} \\
    \quad \model \textit{(Multi-task)} & 6.8 & 165.7 & 68.9\% & 65.2 \\
    \midrule
    \textit{Colorization} & & & & \\
    \quad Regression \textit{(Task-specific)} & 5.5 & 176.9 & 68.0\% & 61.1 \\
    \quad \model \textit{(Task-specific)} & \textbf{3.4} & \textbf{212.9} & \textbf{72.0\%} & \textbf{48.0} \\
    \quad \model \textit{(Multi-task)} & 3.7 & 187.4 & 69.4\% & 57.1 \\
    \midrule
    \textit{JPEG Restoration (QF = 5)} & & & & \\
    \quad Regression \textit{(Task-specific)} & 29.0 & 73.9 & 52.8\% & 155.4 \\
    \quad \model \textit{(Task-specific)} & 8.3 & 133.6 & 64.2\% & 95.5 \\
    \quad \model \textit{(Multi-task)} & \textbf{7.0} & \textbf{137.8} & \textbf{64.7\%} & \textbf{92.4} \\
    \bottomrule
    \end{tabular}
    }
    \vspace*{0.1cm}
    \caption{Performance of multi-task Palette on various tasks.}
    \vspace*{-0.2cm}
    \label{tab:multitask_results}
\end{table}

\section{Conclusion}

We present \model, a simple, general framework for image-to-image translation.
\model achieves strong results on four challenging image-to-image translation tasks (colorization, inpainting, uncropping, and JPEG restoration), outperforming strong GAN and regression baselines. Unlike many GAN models, \model produces diverse and high fidelity outputs.
This is accomplished without task-specific customization nor optimization instability. We also present a multi-task \model model, that performs just as well or better over their task-specific counterparts. Further exploration and investigation of multi-task diffusion models is an exciting avenue for future work. 
This paper shows some of the potential of image-to-image diffusion models, but we look forward to seeing new applications.

\subsection*{Acknowledgements}
We thank John Shlens, Rif A. Saurous, Douglas Eck and the entire Google Brain team for helpful discussions and valuable feedback. We thank Lala Li for help preparing the codebase for public release, and  Erica Moreira for help with compute resources. We also thank Austin Tarango and Philip Parham for help with the approvals for releasing the paper, codebase and checkpoints.

\bibliographystyle{ACM-Reference-Format}
\bibliography{main}

\clearpage
\appendix

\twocolumn[
\begin{flushleft}
 {\LARGE \textbf{Supplementary Material for}}\\
 {\LARGE \textbf{Palette: Image-to-Image Diffusion Models}}\\
 \vspace{0.2cm}
 {\Large Saharia, C.\ et al.} \\ \ \\
\end{flushleft}
]
\counterwithin{figure}{section}
\counterwithin{table}{section}

\section{Diffusion Models}
\label{sec:math-diffusion}

Diffusion models comprise a forward diffusion process and a reverse denoising process that is used at generation time. The forward diffusion process is a Markovian process that iteratively adds Gaussian noise to a data point $\vy_0 \equiv \vy$ over $T$ iterations:
\begin{eqnarray}
    q(\vy_{t+1} | \vy_t) &=& \mathcal{N}(\vy_{t - 1} ; \sqrt{\alpha_t} \vy_{t - 1}, (1 - \alpha_t) I) \\
    q(\vy_{1:T} | \vy_0) &=& \prod_{t = 1}^T q(\vy_t | \vy_{t-1})
\end{eqnarray}
where $\alpha_t$ are hyper-parameters of the noise schedule. The forward process with $\alpha_t$ is constructed in a manner where at $t=T$, $\vy_T$ is virtually indistinguishable from Gaussian noise. Note, we can also marginalize the forward process at each step:
\begin{align}
    q(\vy_t | \vy_0) = \mathcal{N}(\vy_t ; \sqrt{\gamma_t} \vy_0, (1 - \gamma_t) I) ~,
\end{align}
where $\gamma_t = \prod_{t'}^t \alpha_t' \, $.

The Gaussian parameterization of the forward process also allows a closed form formulation of the posterior distribution of $\vy_{t-1}$ given $(\vy_0, \vy_t)$ as
\begin{equation}
    q(\vy_{t-1} \mid \vy_0, \vy_t) ~=~ \mathcal{N}(\vy_{t-1} \mid \bm{\mu}, \sigma^2 \bm{I})
    \label{eq:posteriror-ytm1}
\end{equation}
where $\bm{\mu} =  \frac{\sqrt{\gamma_{t-1}}\,(1-\alpha_t)}{1-\gamma_t}\, \vy_0 \!+\! \frac{\sqrt{\alpha_t}\,(1-\gamma_{t-1})}{1-\gamma_t}\vy_t   $ and
$ \sigma^2 = \frac{(1-\gamma_{t-1})(1-\alpha_t)}{1-\gamma_t} $.
This result proves to be very helpful during inference as shown below.


\textbf{Learning:} \model learns a reverse process which inverts the forward process. Given a noisy image $\vty$,
\begin{equation}
    \vty = \sqrt{\gamma}\, \vy_0 + \sqrt{1-\gamma} \,\veps~,
    ~~~~~~ \veps \sim \mathcal{N}(\bm{0},\bm{I})~,
\label{eq:noisy-y}
\end{equation}
the goal is to recover the target image $\vy_0$. We parameterize our neural network model $f_\theta(x, \vty, \gamma)$ to condition on the input $x$, a noisy image $\vty$, and the current noise level $\gamma$. 
Learning entails prediction of the noise vector $\veps$ 
by optimizing the objective 
\begin{equation}
    \E_{(\vx, \vy)} \E_{\veps, \gamma} \bigg\lVert f_\theta(\vx, \underbrace{\sqrt{\gamma} \,\vy_0 + \sqrt{1-\gamma}\, \veps}_{\vty}, \gamma) - \veps\, \bigg\rVert^{p}_p~.
\label{eq:loss}
\end{equation}
This objective,  also known as $L_{\mathrm{simple}}$ in \cite{ho2020denoising}, is  equivalent to maximizing a weighted variational lower-bound on the likelihood \citep{ho2020denoising}.

\textbf{Inference:} \model performs inference via the learned reverse process. Since 
the forward process is constructed so  the prior distribution p($y_T$) approximates a standard normal distribution $\mathcal{N}(\vy_T | \bm{0}, \bm{I})$, the sampling process can start at pure Gaussian noise, followed by $T$ steps of iterative refinement.  

Also recall that the neural network model $f_{\theta}$ is trained to estimate $\veps$, given any noisy image $\vty$,
and $\vy_t$. Thus, given $\vy_t$, we approximate $\vy_0$ by rearranging terms in \eqref{eq:noisy-y} as
\begin{equation}
    \hat{\vy}_0 = \frac{1}{\sqrt{\gamma_t}} \left( \vy_t - \sqrt{1 - \gamma_t}\, f_{\theta}(\vx, \vy_{t}, \gamma_t) \right)~.
\end{equation}

Following  \cite{ho2020denoising}, we substitute our estimate $\hat{\vy}_0$ into the posterior distribution of $q(\vy_{t-1} | \vy_0, \vy_t)$ in \eqref{eq:posteriror-ytm1} to parameterize the mean of $p_\theta(\vy_{t-1} | \vy_t, \vx)$ as
\begin{equation}
    \mu_{\theta}(\vx, {\vy}_{t}, \gamma_t) = \frac{1}{\sqrt{\alpha_t}} \left( \vy_t - \frac{1-\alpha_t}{ \sqrt{1 - \gamma_t}} f_{\theta}(\vx, \vy_{t}, \gamma_t) \right)~ .
\end{equation}
And we set the variance of $p_\theta(\vy_{t-1}|\vy_t, \vx)$ to $(1 - \alpha_t)$, a default given by the variance of the forward process \cite{ho2020denoising}.

\begin{figure}[t]
\small
\begin{minipage}[t]{0.49\textwidth}
\begin{algorithm}[H]
  \caption{Training a denoising model $f_\theta$} \label{alg:training}
  \small
  \begin{algorithmic}[1]
    \REPEAT
      \STATE $(\vx, \vy_0) \sim p(\vx, \vy)$
      \STATE $\gamma \sim p(\gamma)$
      \STATE $\bm{\epsilon}\sim\mathcal{N}(\mathbf{0},\mathbf{I})$
      \STATE Take a gradient descent step on \\
      $\qquad \nabla_\theta \left\lVert f_\theta(\vx, \sqrt{\gamma} \vy_0 + \sqrt{1-\gamma} \bm{\epsilon}, \gamma) - \veps \right\rVert_p^p$ 
    \UNTIL{converged}
  \end{algorithmic}
\end{algorithm}
\end{minipage}
\hfill
\begin{minipage}[t]{0.49\textwidth}
\begin{algorithm}[H]
  \caption{Inference in $T$ iterative refinement steps} \label{alg:sampling}
  \small
  \begin{algorithmic}[1]
    \vspace{.04in}
    \STATE $\vy_T \sim \mathcal{N}(\mathbf{0}, \mathbf{I})$
    \FOR{$t=T, \dotsc, 1$}
      \STATE $\vz \sim \mathcal{N}(\mathbf{0}, \mathbf{I})$ if $t > 1$, else $\vz = \mathbf{0}$
      \STATE $\vy_{t-1} = \frac{1}{\sqrt{\alpha_t}}\left( \vy_t - \frac{1-\alpha_t}{\sqrt{1-\gamma_t}} f_\theta(\vx, \vy_t, \gamma_t) \right) + \sqrt{1 - \alpha_t} \vz$
    \ENDFOR
    \STATE \textbf{return} $\vy_0$
  \end{algorithmic}
\end{algorithm}
\end{minipage}
\vspace*{-.2cm}
\end{figure}

With this parameterization, each iteration of the reverse process can be computed as
\begin{equation*}
\vy_{t-1} \leftarrow \frac{1}{\sqrt{\alpha_t}} \left( \vy_t - \frac{1-\alpha_t}{ \sqrt{1 - \gamma_t}} f_{\theta}(\vx, \vy_{t}, \gamma_t) \right) + \sqrt{1 - \alpha_t}\veps_t~,
\end{equation*}
where $\veps_t \sim \stdnormal$. This resembles one step of Langevin dynamics for which $f_{\theta}$ provides an estimate of the
gradient of the data log-density.

\section{Implementation Details}
\label{implementation_details}
\textbf{Training Details} : We train all models with a mini batch-size of 1024 for 1M training steps. We do not find over fitting to be an issue, and hence use the model checkpoint at 1M steps for reporting the final results. Consistent with previous works \citep{ho2020denoising, saharia2021image}, we use standard Adam optimizer with a fixed 1e-4 learning rate and 10k linear learning rate warmup schedule. We use 0.9999 EMA for all our experiments. We do not perform any task-specific hyper-parameter tuning, or architectural modifications. 

\textbf{Diffusion Hyper-parameters} : Following \citep{saharia2021image, chen-iclr-2021} we use $\alpha$ conditioning for training \model. This allows us to perform hyper-parameter tuning over noise schedules and refinement steps for \model during inference. During training, we use a linear noise schedule of ($1e^{-6}, 0.01$) with 2000 time-steps, and use 1000 refinement steps with a linear schedule of ($1e^{-4}, 0.09$) during inference.

\textbf{Task Specific Details:} We specify specific training details for each of the tasks below:
\begin{itemize}[topsep=0pt, partopsep=0pt, leftmargin=13pt, parsep=0pt, itemsep=0pt]
    \item \textbf{Colorization} : We use RGB parameterization for colorization. We use the grayscale image as the source image and train \model to predict the full RGB image. During training, following \citep{coltran}, we randomly select the largest square crop from the image and resize it to 256$\times$256.  
    \item \textbf{Inpainting} : We train \model on a combination of free-form and rectangular masks. For free-form masks, we use Algorithm 1 in \citep{yu2019free}. For rectangular masks, we uniformly sample between 1 and 5 masks. The total area covered by the rectangular masks is kept between 10\% to 40\% of the image. We randomly sample a free-form mask with 60\% probability, and rectangular masks with 40\% probability. Note that this is an arbitrary training choice. We do not provide any additional mask channel, and simply fill the masked region with random Gaussian noise. During training, we restrict the $L_{simple}$ loss function to the spatial region corresponding to masked regions, and use the model's prediction for only the masked region during inference. We train \model on two types of 256$\times$256 crops. Consistent with previous inpainting works \citep{yu2019free, yu2018generative, yi2020contextual}, we use random 256$\times$256 crops, and we combine these with the resized random largest square crops used in colorization literature \citep{coltran}.
    \item \textbf{Uncropping} : We train the model for image extension along all four directions, or just one direction. In both cases, we set the masked region to 50\% of the image. During training, we uniformly choose masking along one side, or masking along all 4 sides. When masking along one side, we further make a uniform random choice over the side. Rest of the training details are identical to inpainting. 
    \item \textbf{JPEG Restoration} : We train \model for JPEG restoration on quality factors in (5, 30). Since decompression for lower quality factors is a significantly more difficult task, we use an exponential distribution to sample the quality factor during training. Specifically, the sampling probability of a quality range $Q$ is set to $\propto e^{-\frac{Q}{10}}$.
\end{itemize}

\section{Additional Experimental Results}
\label{expt_appendix}

\subsection{Colorization}
\label{colorization_appendix}

Following prior work \citep{zhang2016colorful, guadarrama2017pixcolor, coltran}, we train and evaluate models on ImageNet \citep{deng2009imagenet}. In order to compare our models with existing works in Table \ref{tab:colorization_results}, we follow ColTran \citep{coltran} and use the first 5000 images from ImageNet validation set to report performance on standard metrics. We use the next 5000 images as the reference distribution for FID to mirror ColTran's implementation (as returned by TFDS \citep{TFDS} data loader). For benchmarking purposes, we also report the performance of \model on ImageNet ctest10k \citep{larsson2016learning} dataset in Table \ref{tab:colorization_benchmark}.

\begin{figure*}[t]
\small
    \begin{center}
    \setlength{\tabcolsep}{2pt}
    \begin{tabular}{c@{\hspace{.3cm}}c}
    {Fool Rate \% on Set-I (3 sec display)} & {Fool Rate \% on Set-II (3 sec display)} \\
    {\includegraphics[width=.45\textwidth]{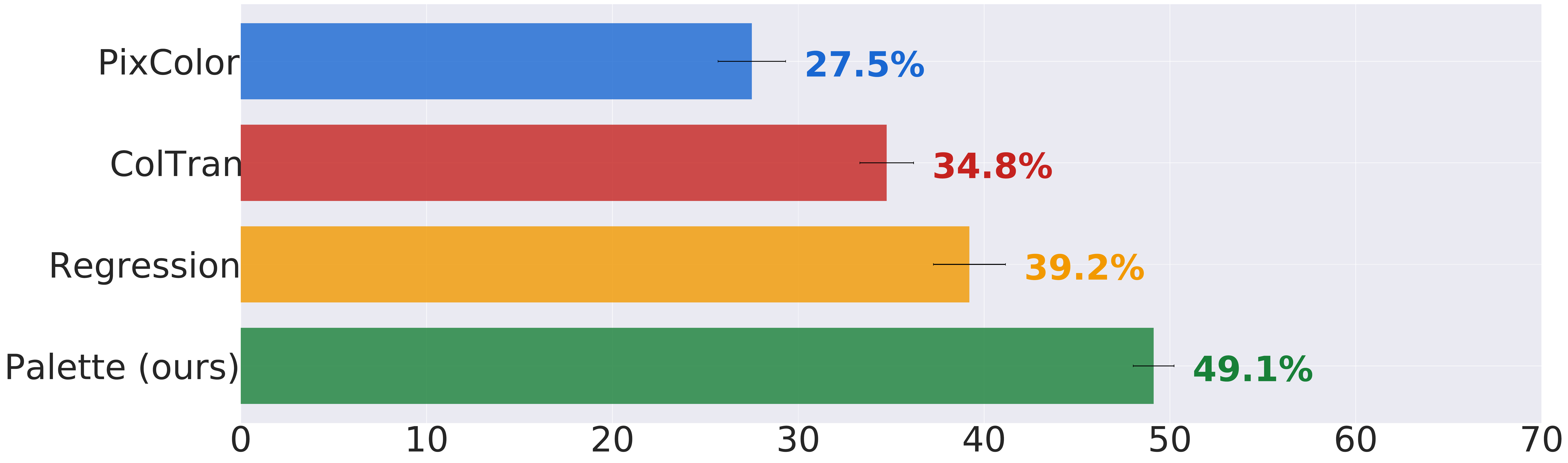}} &
    {\includegraphics[width=.45\textwidth]{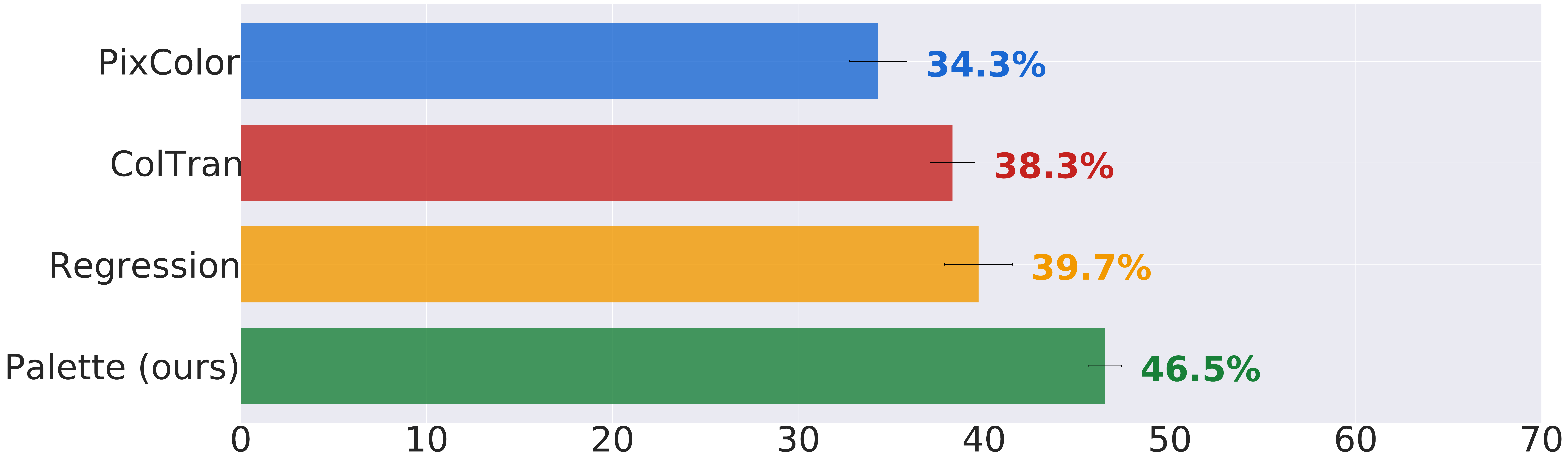}} \\
    
    {Fool Rate \% on Set-I (5 sec display)} & {Fool Rate \% on Set-II (5 sec display)} \\
    {\includegraphics[width=.45\textwidth]{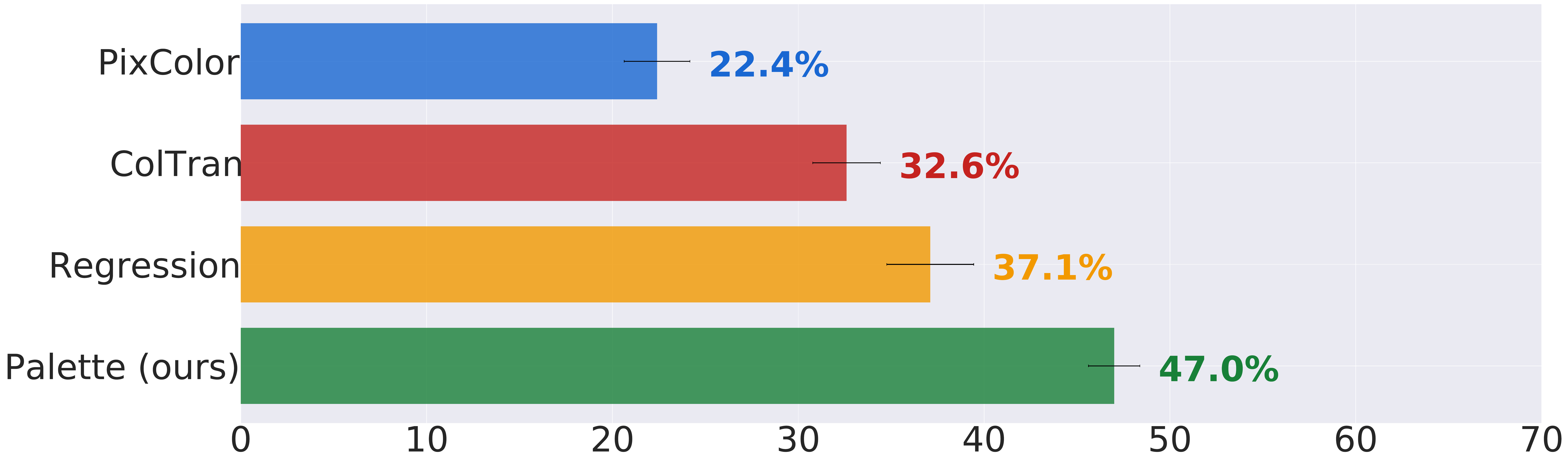}} &
    {\includegraphics[width=.45\textwidth]{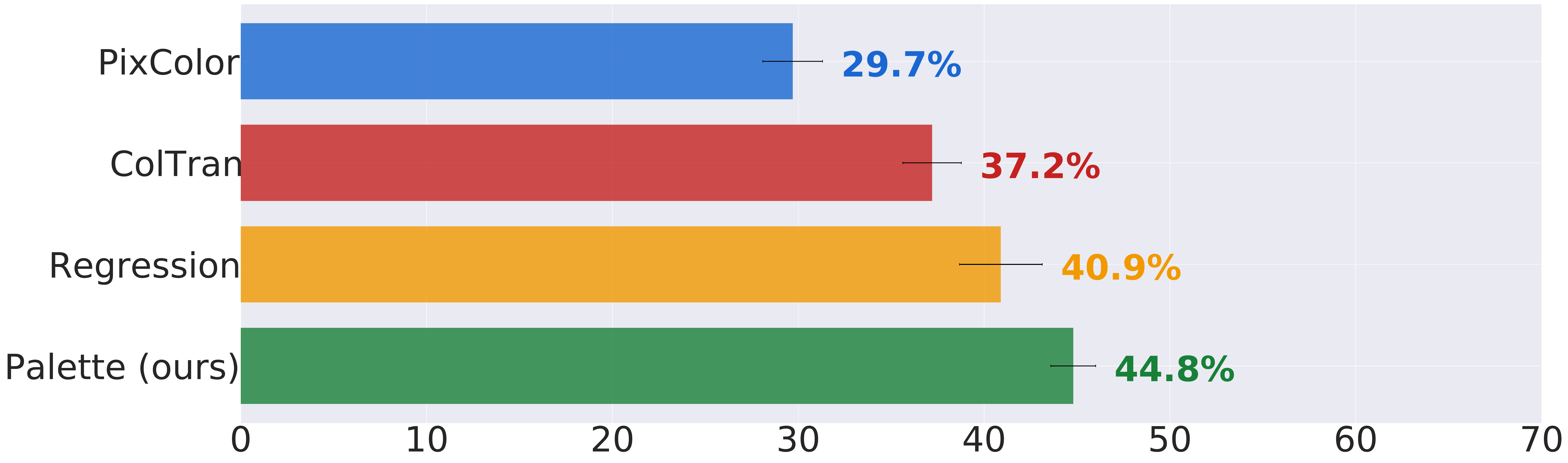}} \\
    \end{tabular}
    \end{center}
     \vspace*{-0.5cm}
    \caption{Human evaluation results on ImageNet colorization.
    \label{fig:fool_rates_colorization}
    }
    \vspace*{-0.1cm}
\end{figure*}
\begin{figure*}[t]
\small
    \begin{center}
    
    \setlength{\tabcolsep}{2pt}
    \begin{tabular}{c@{\hspace{.3cm}}c}
    {Fool Rate \% on Set-I (3 sec display)} & {Fool Rate \% on Set-II (3 sec display)} \\
    {\includegraphics[width=.45\textwidth]{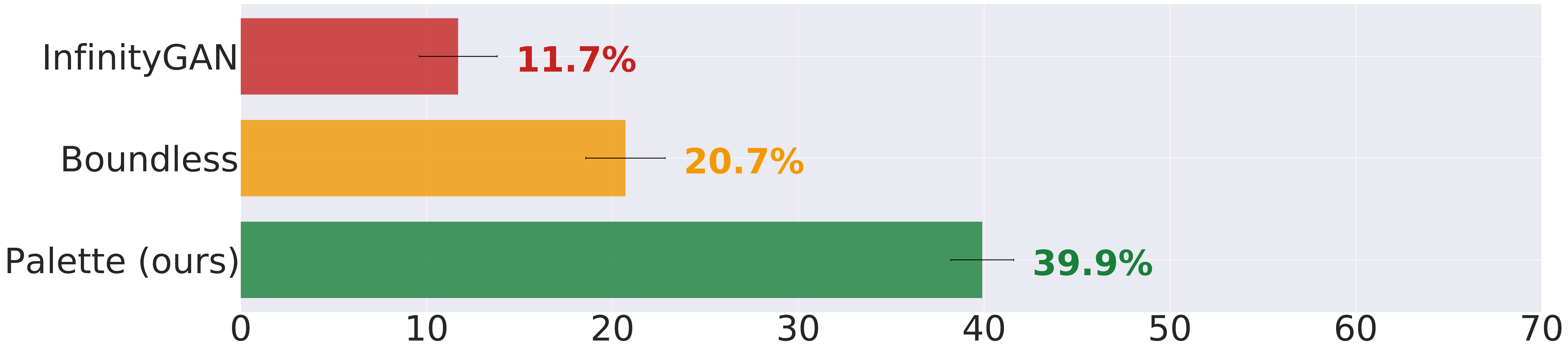}} &
    {\includegraphics[width=.45\textwidth]{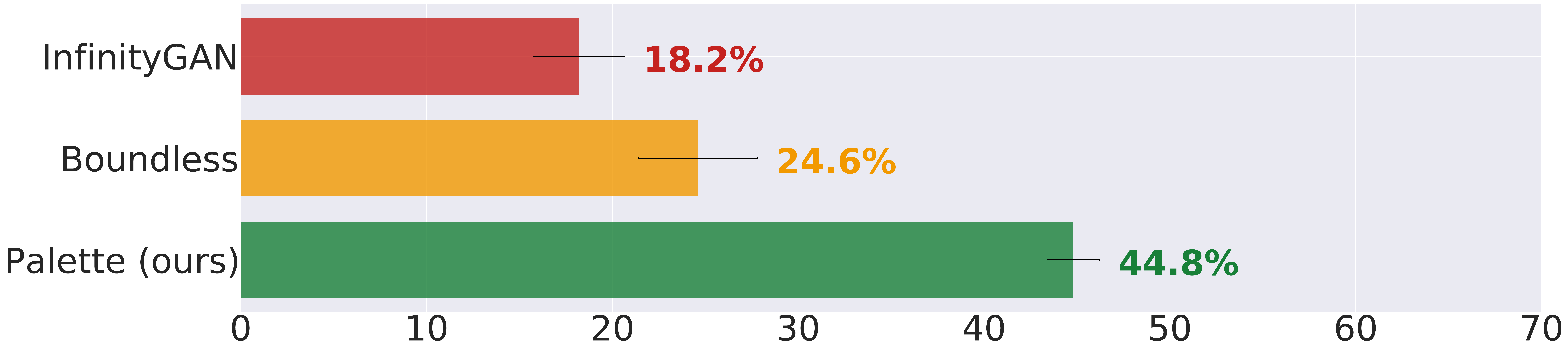}} \\
    
    {Fool Rate \% on Set-I (5 sec display)} & {Fool Rate \% on Set-II (5 sec display)} \\
    {\includegraphics[width=.45\textwidth]{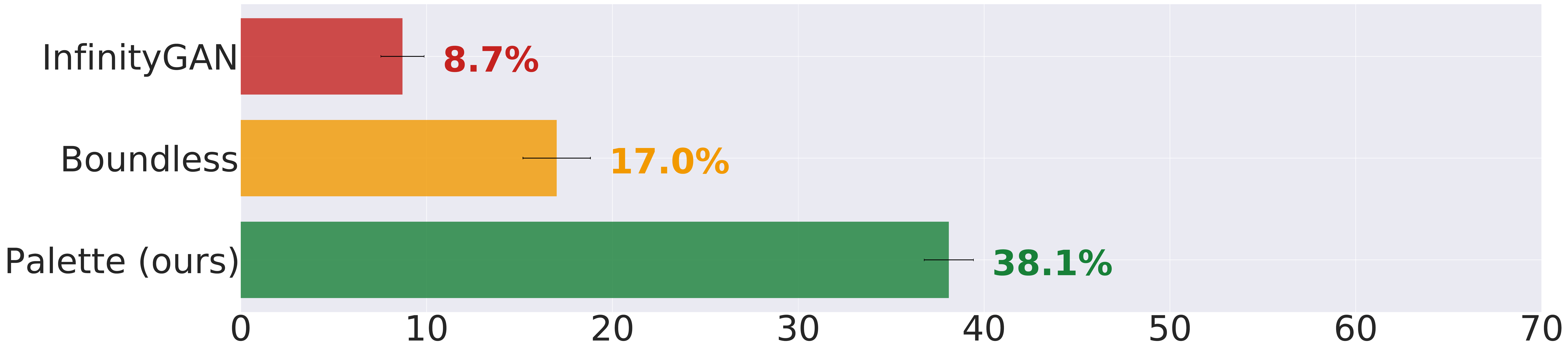}} &
    {\includegraphics[width=.45\textwidth]{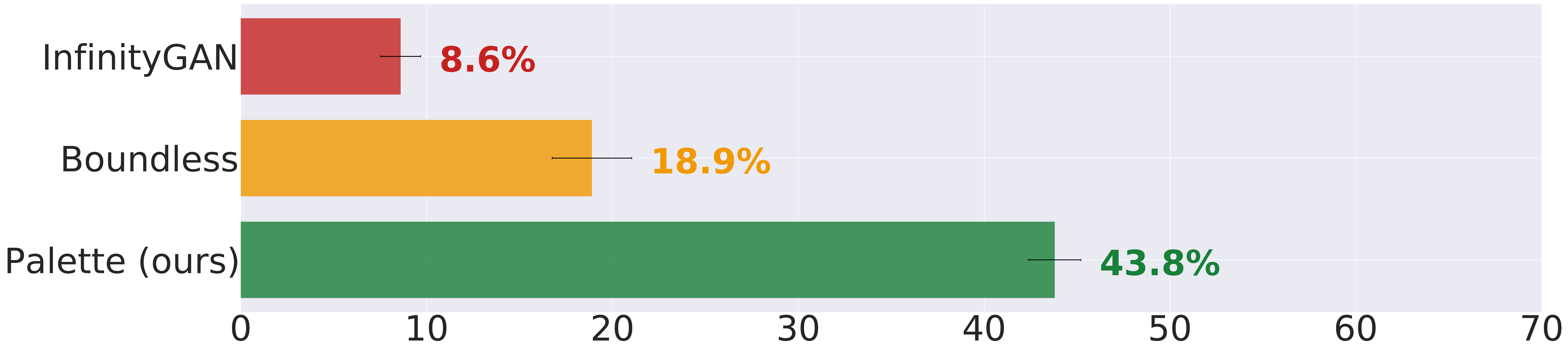}} \\
    
    \end{tabular}
    \end{center}
     \vspace*{-0.5cm}
    \caption{Human evaluation results on Places2 uncropping.}
    \label{fig:fool_rates_uncropping}
    \vspace*{-0.2cm}
    \end{figure*}

\begin{figure*}[h]
\setlength{\tabcolsep}{1pt}
\begin{center}
{\small 
\begin{tabular}{cccccc}
{\small Grayscale Input } & {\small PixColor\textsuperscript{\textdagger}}  & {\small ColTran\textsuperscript{\ddag} } & {\small Regression} &  {\small \model (Ours)} & {\small Original} 
\\
{\includegraphics[width=0.155\textwidth]{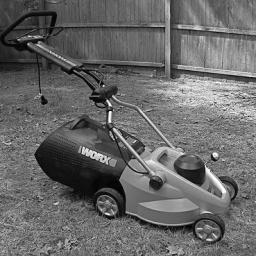}} &
{\includegraphics[width=0.155\textwidth]{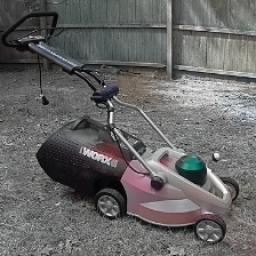}} &
{\includegraphics[width=0.155\textwidth]{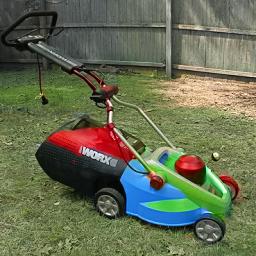}} &
{\includegraphics[width=0.155\textwidth]{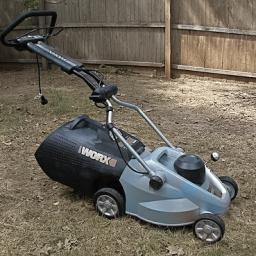}} &
{\includegraphics[width=0.155\textwidth]{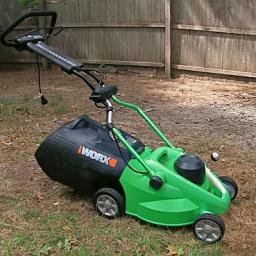}} &
{\includegraphics[width=0.155\textwidth]{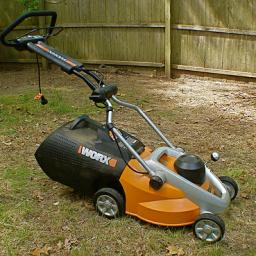}} \\

{\includegraphics[width=0.155\textwidth]{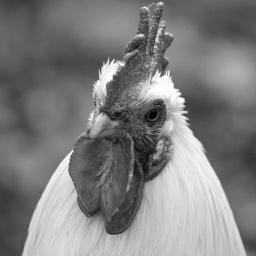}} &
{\includegraphics[width=0.155\textwidth]{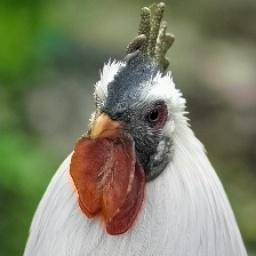}} &
{\includegraphics[width=0.155\textwidth]{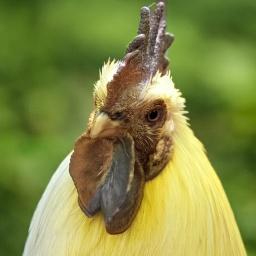}} &
{\includegraphics[width=0.155\textwidth]{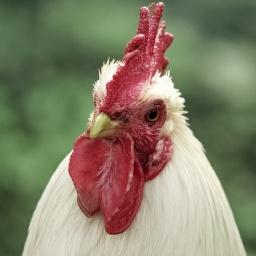}} &
{\includegraphics[width=0.155\textwidth]{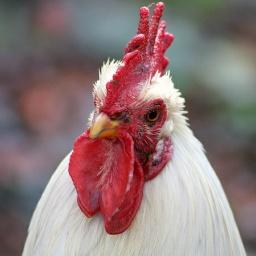}} &
{\includegraphics[width=0.155\textwidth]{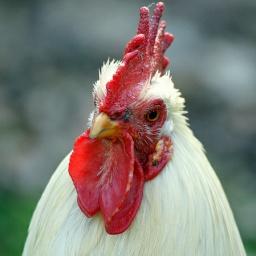}} \\

{\includegraphics[width=0.155\textwidth]{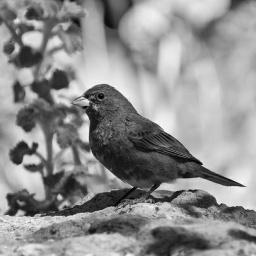}} &
{\includegraphics[width=0.155\textwidth]{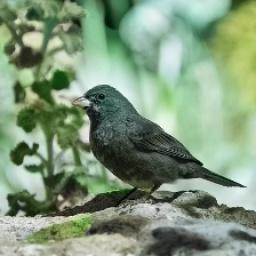}} &
{\includegraphics[width=0.155\textwidth]{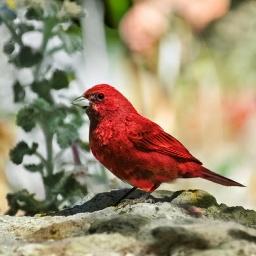}} &
{\includegraphics[width=0.155\textwidth]{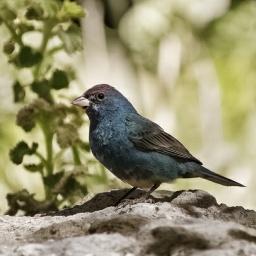}} &
{\includegraphics[width=0.155\textwidth]{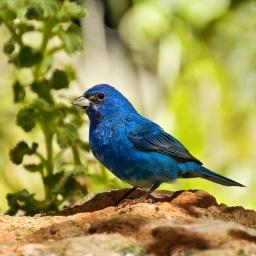}} &
{\includegraphics[width=0.155\textwidth]{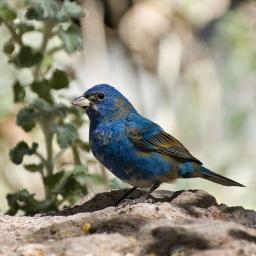}} \\

{\includegraphics[width=0.155\textwidth]{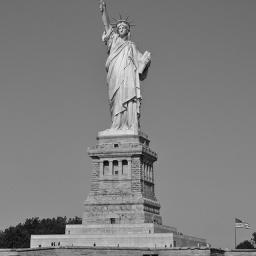}} &
{\includegraphics[width=0.155\textwidth]{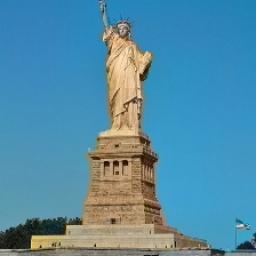}} &
{\includegraphics[width=0.155\textwidth]{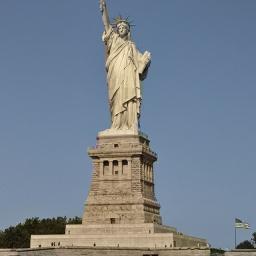}} &
{\includegraphics[width=0.155\textwidth]{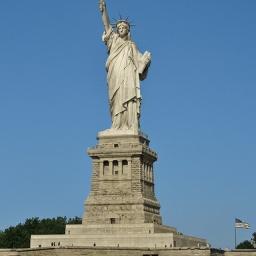}} &
{\includegraphics[width=0.155\textwidth]{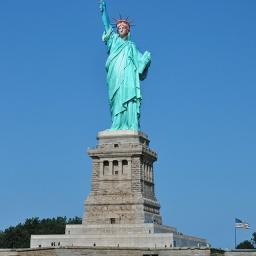}} &
{\includegraphics[width=0.155\textwidth]{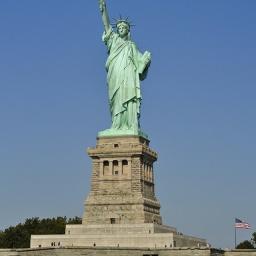}} \\

{\includegraphics[width=0.155\textwidth]{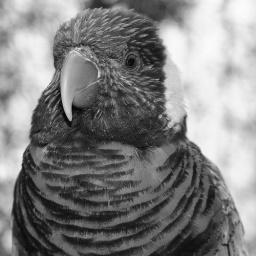}} &
{\includegraphics[width=0.155\textwidth]{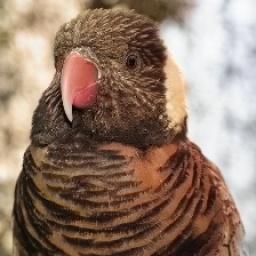}} &
{\includegraphics[width=0.155\textwidth]{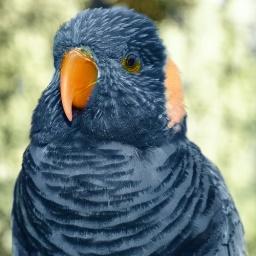}} &
{\includegraphics[width=0.155\textwidth]{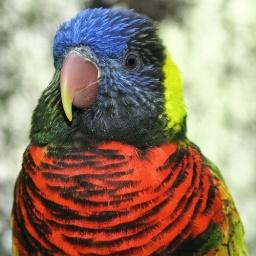}} &
{\includegraphics[width=0.155\textwidth]{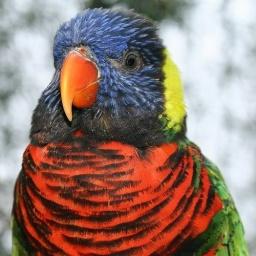}} &
{\includegraphics[width=0.155\textwidth]{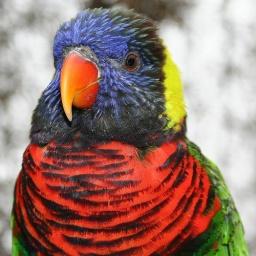}} \\

{\includegraphics[width=0.155\textwidth]{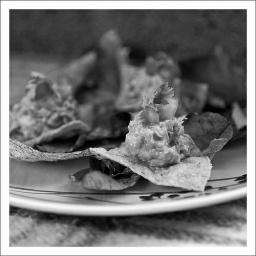}} &
{\includegraphics[width=0.155\textwidth]{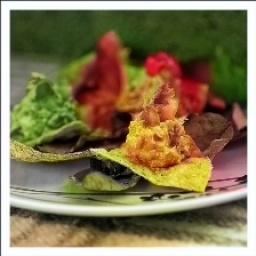}} &
{\includegraphics[width=0.155\textwidth]{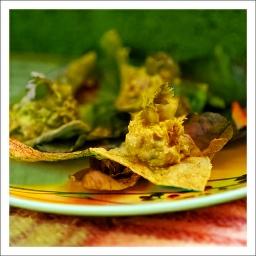}} &
{\includegraphics[width=0.155\textwidth]{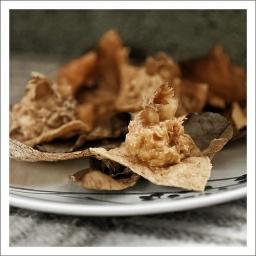}} &
{\includegraphics[width=0.155\textwidth]{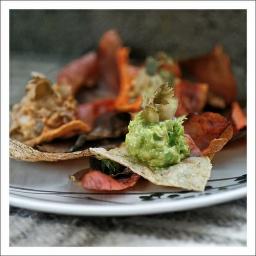}} &
{\includegraphics[width=0.155\textwidth]{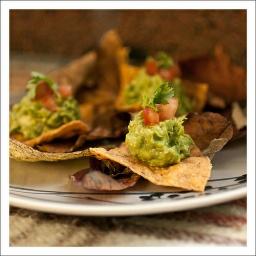}} \\

{\includegraphics[width=0.155\textwidth]{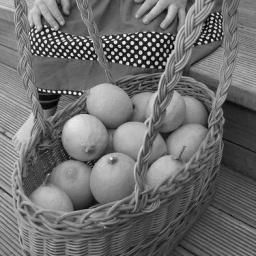}} &
{\includegraphics[width=0.155\textwidth]{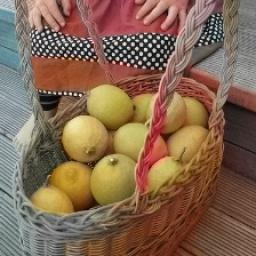}} &
{\includegraphics[width=0.155\textwidth]{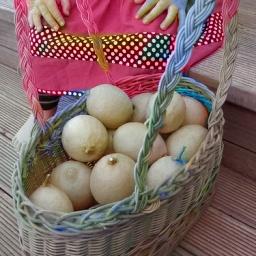}} &
{\includegraphics[width=0.155\textwidth]{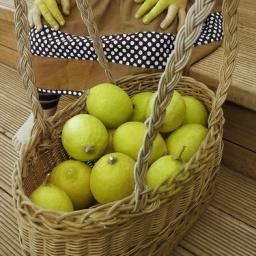}} &
{\includegraphics[width=0.155\textwidth]{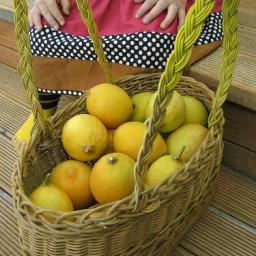}} &
{\includegraphics[width=0.155\textwidth]{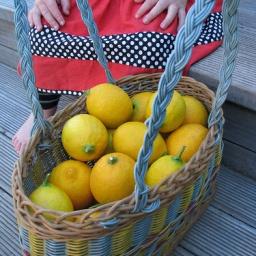}} \\


\end{tabular}
}
\end{center}
\vspace*{-0.6cm}
\caption{Comparison of different methods for colorization on ImageNet validation images. Baselines: \textsuperscript{\textdagger}\citep{guadarrama2017pixcolor}
and \textsuperscript{\ddag}\citep{coltran}.
\label{fig:colorization_comparison_appendix}
}
\end{figure*}

\begin{table}[t]
    \centering
    {\small
    \begin{tabular}{lcccc}
    \toprule
    \bfseries{Model} & \bfseries{FID-10K} $\downarrow$ & \bfseries{IS} $\uparrow$ & \bfseries{CA} $\uparrow$ & \bfseries{PD} $\downarrow$ \\
    \midrule
    \model ($L_2$) & 3.4 & 212.9 & 72.0\% & 48.0 \\
    \model ($L_1$) & 3.4 & 215.8 & 71.9\% & 45.8 \\
    \midrule
    Ground Truth & 2.7 & 250.1 & 76.0\% & 0.0 \\
    \bottomrule
    \end{tabular}
    }
    \vspace*{0.1cm}
    \caption{Benchmark numbers on ctest10k ImageNet subset for Image Colorization.}
    \label{tab:colorization_benchmark}
    \vspace*{-0.6cm}
\end{table}

\textbf{Human Evaluation:}
The ultimate evaluation of image-to-image translation models is human evaluation; \ie~whether or not humans can discriminate model outputs from reference images. To this end we use controlled human experiments. 
In a series of two alternative forced choice trials, we ask subjects which of two side-by-side images is the real photo and which has been generated by the model.
In particular, subjects are asked 
\textit{``Which image would you guess is from a camera?"}
Subjects viewed images for either 3 or 5 seconds before having to respond.  
For the experiments we compare outputs from four models against reference images, namely,  PixColor~\citep{guadarrama2017pixcolor},  Coltran~\citep{coltran}, our Regression baseline, 
and \model.
To summarize the result we compute the subject {\it fool rate}, i.e., the fraction of human raters who select the model outputs over the reference image. 
We use a total of 100 images for human evaluation, and divide these into two independent subsets - Set-I and Set-II, each of which is seen by 50 subjects.

\begin{figure*}[h]
\setlength{\tabcolsep}{2pt}
\begin{center}
{\small 
\begin{tabular}{cccccc}
{\small Input} & {\small Sample 1}  & {\small Sample 2} & {\small Sample 3} &  {\small Sample 4} & {Original} 
\\
{\includegraphics[width=0.157\textwidth]{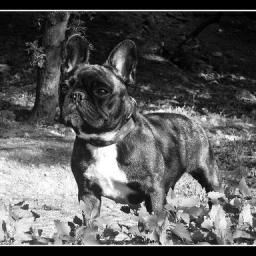}} &
{\includegraphics[width=0.157\textwidth]{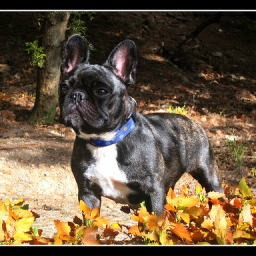}} &
{\includegraphics[width=0.157\textwidth]{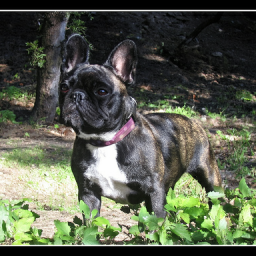}} &
{\includegraphics[width=0.157\textwidth]{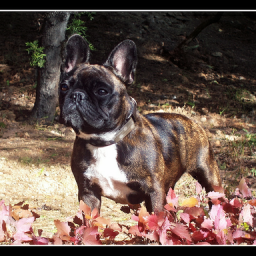}} &
{\includegraphics[width=0.157\textwidth]{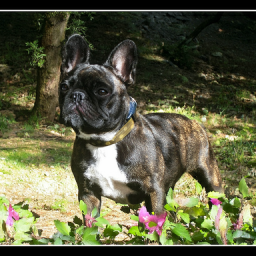}} &
{\includegraphics[width=0.157\textwidth]{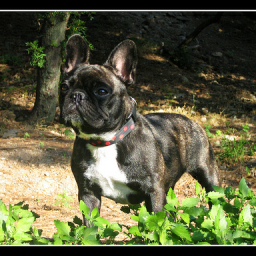}}\\

{\includegraphics[width=0.157\textwidth]{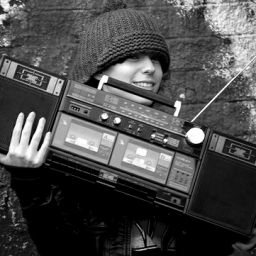}} &
{\includegraphics[width=0.157\textwidth]{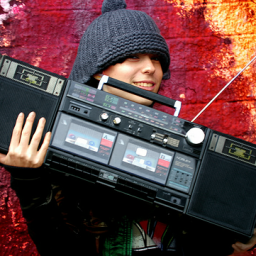}} &
{\includegraphics[width=0.157\textwidth]{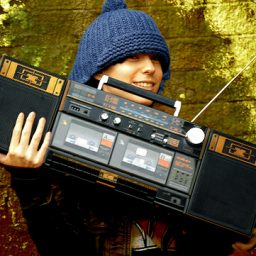}} &
{\includegraphics[width=0.157\textwidth]{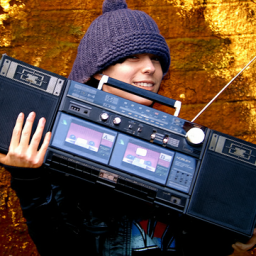}} &
{\includegraphics[width=0.157\textwidth]{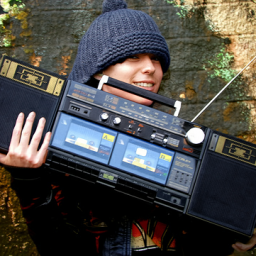}} &
{\includegraphics[width=0.157\textwidth]{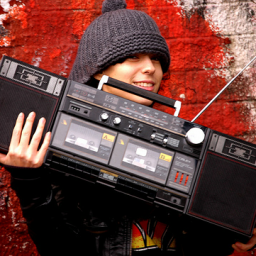}}\\

{\includegraphics[width=0.157\textwidth]{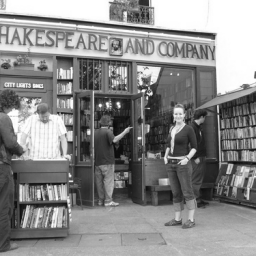}} &
{\includegraphics[width=0.157\textwidth]{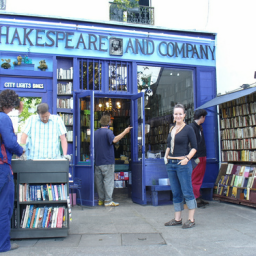}} &
{\includegraphics[width=0.157\textwidth]{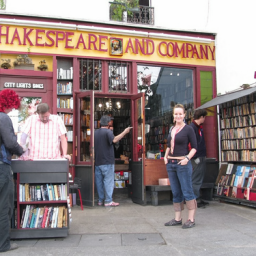}} &
{\includegraphics[width=0.157\textwidth]{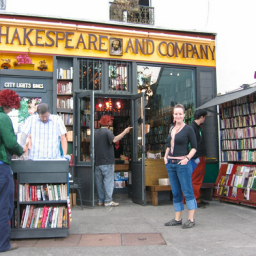}} &
{\includegraphics[width=0.157\textwidth]{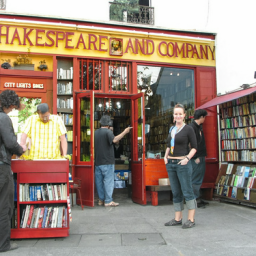}} &
{\includegraphics[width=0.157\textwidth]{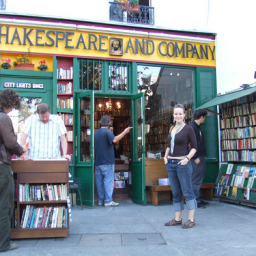}}\\

{\includegraphics[width=0.157\textwidth]{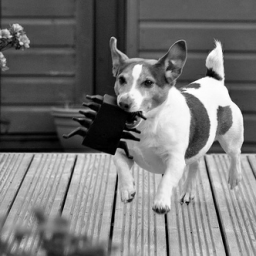}} &
{\includegraphics[width=0.157\textwidth]{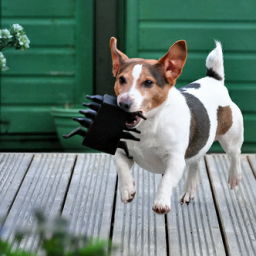}} &
{\includegraphics[width=0.157\textwidth]{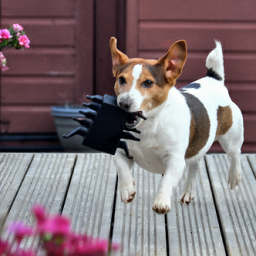}} &
{\includegraphics[width=0.157\textwidth]{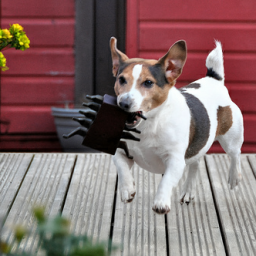}} &
{\includegraphics[width=0.157\textwidth]{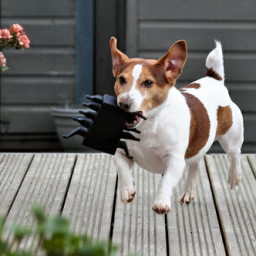}} &
{\includegraphics[width=0.157\textwidth]{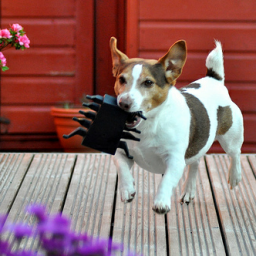}}\\

{\includegraphics[width=0.157\textwidth]{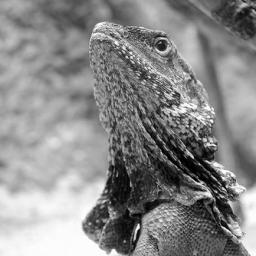}} &
{\includegraphics[width=0.157\textwidth]{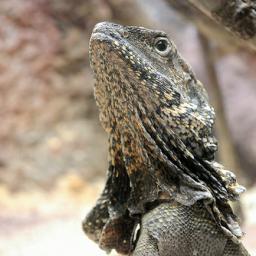}} &
{\includegraphics[width=0.157\textwidth]{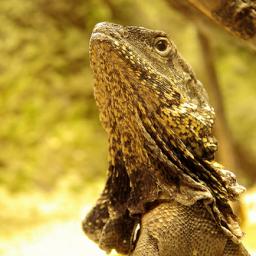}} &
{\includegraphics[width=0.157\textwidth]{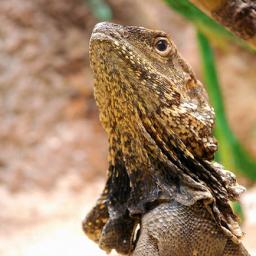}} &
{\includegraphics[width=0.157\textwidth]{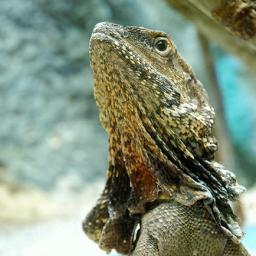}} &
{\includegraphics[width=0.157\textwidth]{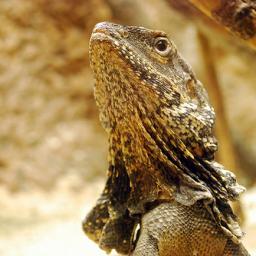}}\\

{\includegraphics[width=0.157\textwidth]{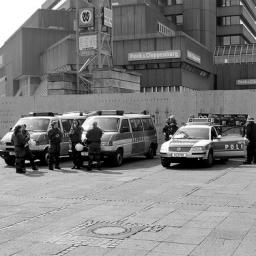}} &
{\includegraphics[width=0.157\textwidth]{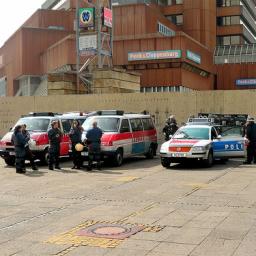}} &
{\includegraphics[width=0.157\textwidth]{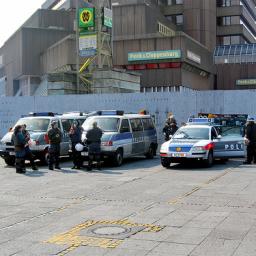}} &
{\includegraphics[width=0.157\textwidth]{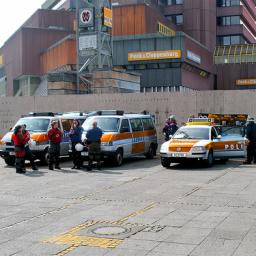}} &
{\includegraphics[width=0.157\textwidth]{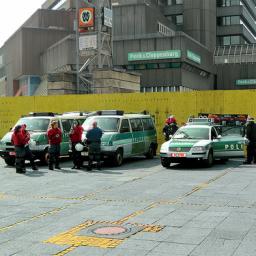}} &
{\includegraphics[width=0.157\textwidth]{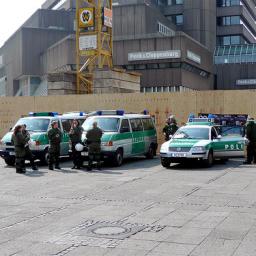}}\\


{\includegraphics[width=0.157\textwidth]{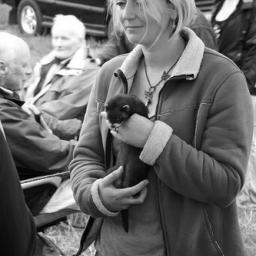}} &
{\includegraphics[width=0.157\textwidth]{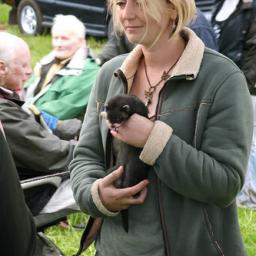}} &
{\includegraphics[width=0.157\textwidth]{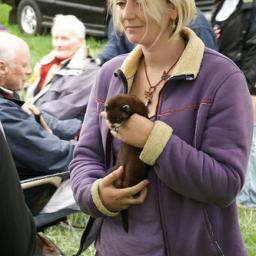}} &
{\includegraphics[width=0.157\textwidth]{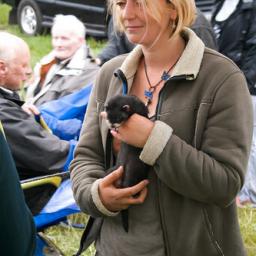}} &
{\includegraphics[width=0.157\textwidth]{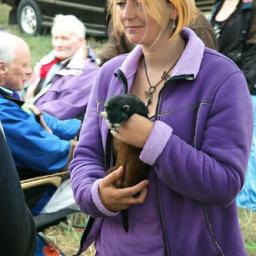}} &
{\includegraphics[width=0.157\textwidth]{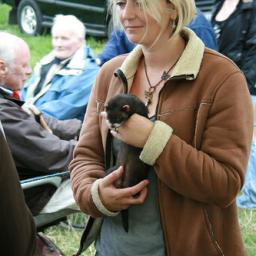}}\\


\end{tabular}
}
\end{center}
\vspace*{-0.3cm}
\caption{Diversity of \model outputs on ImageNet colorization validation images.
\label{fig:colorization_diversity_appendix}
}
\end{figure*}

As shown in Figure \ref{fig:fool_rates_colorization}, the fool rate for \model is close to 50\% and higher than baselines in all cases.  We note that when subjects are given less time to inspect the images the fool rates are somewhat higher, as expected.  We also note the strength of our regression baseline, which also performs better than PixColor and Coltran.
Finally, to provide insight into the human evaluation results we also show several more examples of \model output, with comparisons to benchmarks, in Figure \ref{fig:colorization_comparison_appendix}. 
One can see that in several cases, \model has learned colors that are more meaningful and consistent with the reference images and the semantic content of the images.
Figure \ref{fig:colorization_diversity_appendix} also shows the natural diversity of \model outputs for colorization model.

\subsection{Inpainting}
\label{inpainting_appendix}

\textbf{Comparison on 256$\times$256 images}: We report all inpainting results on 256$\times$256 center cropped images. Since the prior works we use for comparison are all trained on random 256$\times$256 crops, evaluation on 256$\times$256 center crops ensures fair comparison. Furthermore, we use a fixed set of image-mask pair for each configuration for all models during evaluation. Since HiFill \citep{yi2020contextual} and Co-ModGAN \citep{yi2020contextual} are primarily trained on 512$\times$512 images, we use 512$\times$512 center crops with exact same mask within the central 256$\times$256 region. This provides these two models with 4$\times$ bigger inpainting context compared to DeepFillv2 and \model. 

We train two \model models for Inpainting - i) \model (I) trained on ImageNet dataset, and ii) \model (I+P) trained on mixture of ImageNet and Places2 dataset. For \model (I+P), we use a random sampler policy to sample from ImageNet and Places2 dataset with a uniform probability. Table \ref{tab:inpainting_results_full} shows full comparison of \model with existing methods on all inpainting configurations. Based on the type of mask, and the area covered, we report results for the following categories - i) 10-20\% free-form region, ii) 20-30\% free-form region, iii) 30-40\% free-form region and iv) 128$\times$128 center rectangle region. \model consistently outperforms existing works by a significant margin on all configurations. Interestingly \model (I) performs slightly better than \model (I+P) on ImageNet indicating that augmentation with Places2 images during training doesn't boost to ImageNet performance. Furthermore, \model (I) is only slightly worse compared to \model (I+P) on Places2 even though it is not trained on Places2 images. We observe a significant drop in the performance of HiFill \citep{yi2020contextual} with larger masks. It is important to note that DeepFillv2 and HiFill are not trained on ImageNet, but we report their performance on ImageNet ctest10k primarily for benchmarking purposes.

\begin{table*}[t]
\small
    \centering
    {\small
    \begin{tabular}{llccccccc}
    \toprule
    \bfseries{Mask Type} & \bfseries{Model} & \multicolumn{4}{c}{\bfseries{ImageNet}} & & \multicolumn{2}{c}{\bfseries{Places2}} \\
    \midrule
    & & \bfseries{FID}  $\downarrow$ & \bfseries{IS}  $\uparrow$ & \bfseries{CA} $\uparrow$ & \bfseries{PD} $\downarrow$ & & \bfseries{FID} $\downarrow$ & \bfseries{PD} $\downarrow$ \\
    \midrule
    \textit{10-20\%} & DeepFillv2 \citep{yu2019free} & 6.7 & 198.2 & 71.6\% & 38.6 & &  12.2 & 38.1  \\
    \textit{Free-Form} & HiFill \citep{yi2020contextual} & 7.5 & 192.0 & 70.1\% & 46.9 & &  13.0 & 55.1  \\
    \textit{Mask} & \model (I) (Ours) & \textbf{5.1} & \textbf{221.0} & \textbf{73.8\%} & 15.6 & & 11.6 & 22.1 \\
    & \model (I+P) (Ours) & 5.2 & 219.2 & 73.7\% & \textbf{15.5} & & \textbf{11.6} & \textbf{20.3} \\
    \midrule
    \textit{20-30\%} & DeepFillv2 \citep{yu2019free} & 9.4 & 174.6 & 68.8\% & 64.7 & & 13.5 & 63.0 \\
    \textit{Free-Form} & HiFill \citep{yi2020contextual} & 12.4 & 157.0 & 65.7\% & 86.2 & & 15.7 & 92.8  \\
    \textit{Mask} & Co-ModGAN \citep{zhao2021large} & - & - & - & - & & 12.4 & 51.6  \\
    & \model (I) (Ours) & \textbf{5.2} & \textbf{208.6} & \textbf{72.6\%} & \textbf{27.4} & & 11.8 & 37.7 \\
    & \model (I+P) (Ours) & \textbf{5.2} & 205.5 & 72.3\% & 27.6 & & \textbf{11.7} & \textbf{35.0}  \\
    \midrule
    \textit{30-40\%} & DeepFillv2 \citep{yu2019free} & 14.2 & 144.7 & 64.9\% & 95.5 & &  15.8 & 90.1 \\
    \textit{Free-Form} & HiFill \citep{yi2020contextual} & 20.9 & 115.6 & 59.4\% & 131.0 & &  20.1 & 132.0  \\
    \textit{Mask} & \model (I)  & \textbf{5.5} & \textbf{195.2} & \textbf{71.4\%} & \textbf{39.9} & &  12.1 & 53.5 \\
    & \model (I+P)  & 5.6 & 192.8 &  71.3\% & 40.2 & &  \textbf{11.6} & \textbf{49.2}  \\
    \midrule
    \textit{128$\times$128} & DeepFillv2 \citep{yu2019free} & 18.0 & 135.3 & 64.3\% & 117.2 & & 15.3 & 96.3 \\
    \textit{Center} & HiFill \citep{yi2020contextual} & 20.1 & 126.8 & 62.3\% & 129.7 & &  16.9 & 115.4  \\
    \textit{Mask} & \model (I) & \textbf{6.4} & 173.3 & \textbf{69.7\%} & \textbf{58.8} & & 12.2 & 62.8 \\
    & Co-ModGAN \citep{zhao2021large} & - & - & - & - & & 13.7 & 86.2  \\
    & \model (I+P) & 6.6 & \textbf{173.9} &  69.3\% & 59.5 & & \textbf{11.9} & \textbf{57.3}  \\
    \midrule
    & Ground Truth & 5.1 & 231.6 & 74.6\% & 0.0 & &  11.4 & 0.0 \\
    \bottomrule
    \end{tabular}
    }
    \vspace*{0.1cm}
    \caption{Quantitative evaluation for inpainting on ImageNet and Places2 validation images.}
    \label{tab:inpainting_results_full}
\vspace{-.6cm}
\end{table*}

\begin{figure*}[h]
\setlength{\tabcolsep}{1pt}
\begin{center}
{\small 
\begin{tabular}{ccccccc}
{\small Masked Input } &{\small Photoshop 2021\textsuperscript{\ddag}} & {\small DeepFillv2\textsuperscript{\textdagger}} & {\small HiFill\textsuperscript{\textdagger\textdagger}} & {\small Co-ModGAN\textsuperscript{\ddag\ddag}} & {\small \model (Ours)} 
 & {\small Original}
\\
{\includegraphics[width=0.15\textwidth]{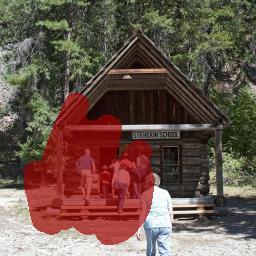}} &
{\includegraphics[width=0.15\textwidth]{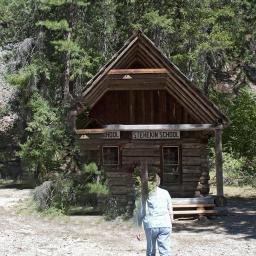}} &
{\includegraphics[width=0.15\textwidth]{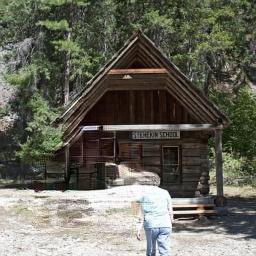}} &
{\includegraphics[width=0.15\textwidth]{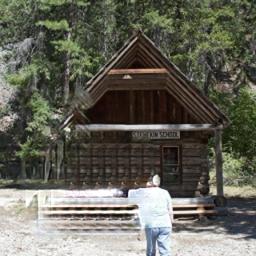}} &
{\includegraphics[width=0.15\textwidth]{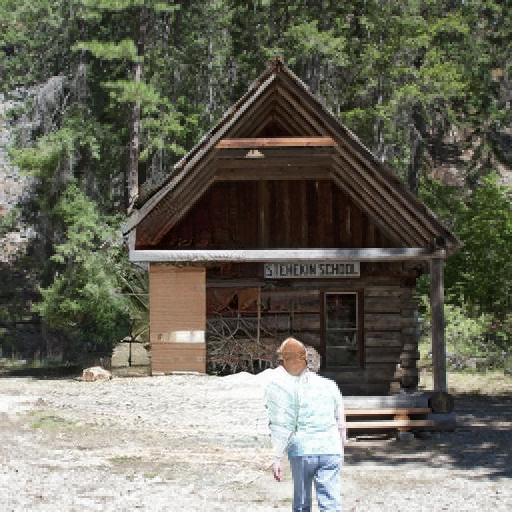}} &
{\includegraphics[width=0.15\textwidth]{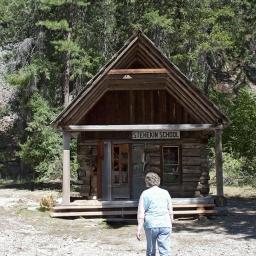}} &
{\includegraphics[width=0.15\textwidth]{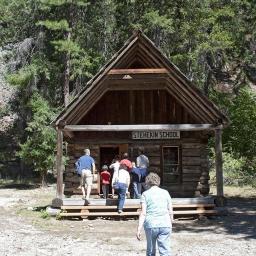}} \\

{\includegraphics[width=0.15\textwidth]{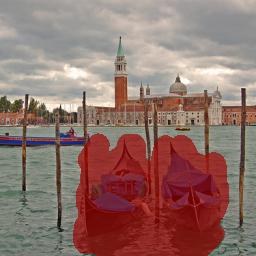}} &
{\includegraphics[width=0.15\textwidth]{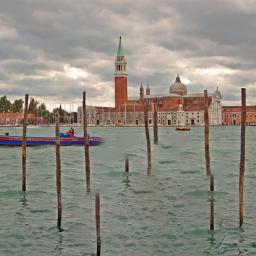}} &
{\includegraphics[width=0.15\textwidth]{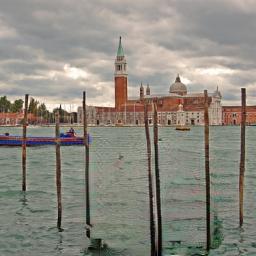}} &
{\includegraphics[width=0.15\textwidth]{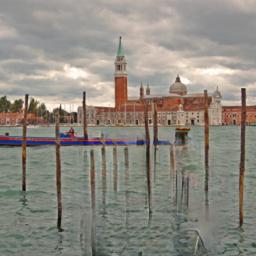}} &
{\includegraphics[width=0.15\textwidth]{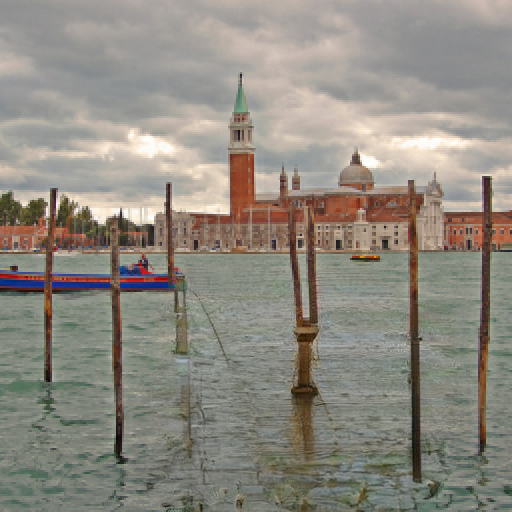}} &
{\includegraphics[width=0.15\textwidth]{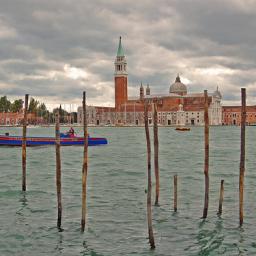}} &
{\includegraphics[width=0.15\textwidth]{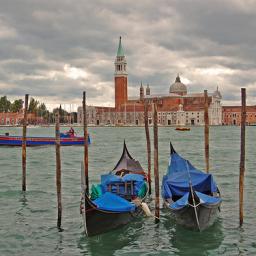}} \\

{\includegraphics[width=0.15\textwidth]{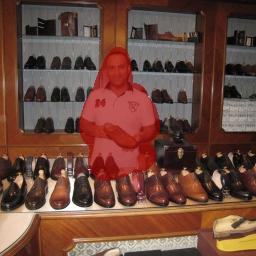}} &
{\includegraphics[width=0.15\textwidth]{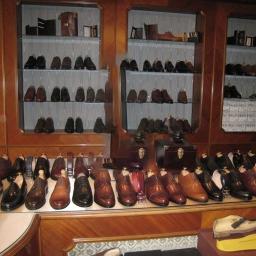}} &
{\includegraphics[width=0.15\textwidth]{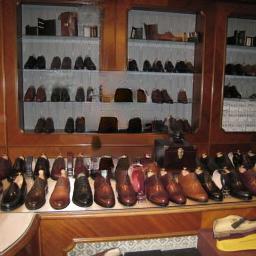}} &
{\includegraphics[width=0.15\textwidth]{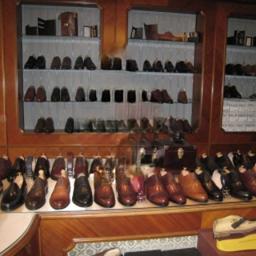}} &
{\includegraphics[width=0.15\textwidth]{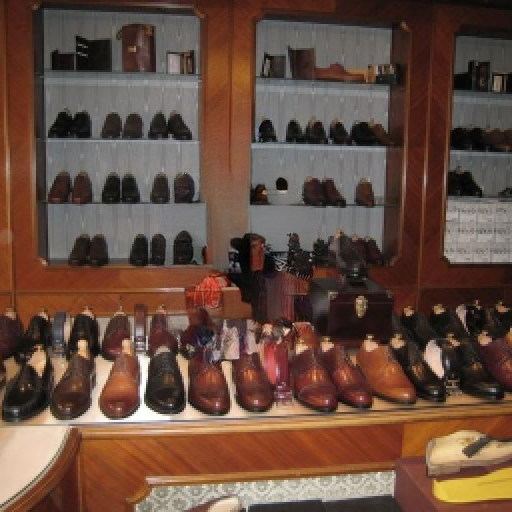}} &
{\includegraphics[width=0.15\textwidth]{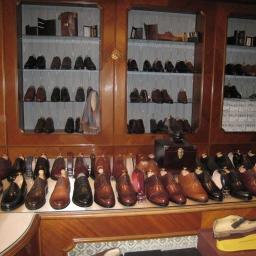}} &
{\includegraphics[width=0.15\textwidth]{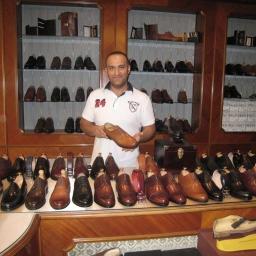}} \\


{\includegraphics[width=0.15\textwidth]{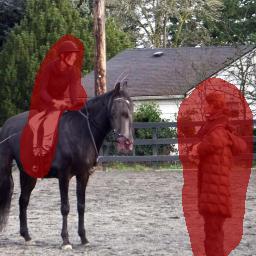}} &
{\includegraphics[width=0.15\textwidth]{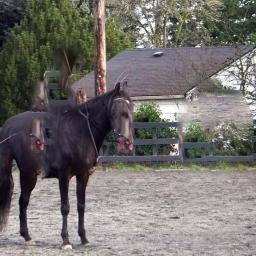}} &
{\includegraphics[width=0.15\textwidth]{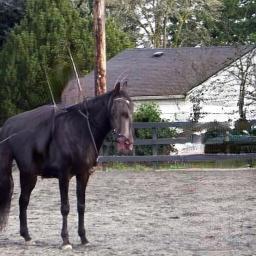}} &
{\includegraphics[width=0.15\textwidth]{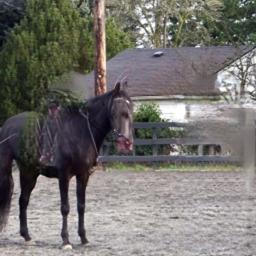}} &
{\includegraphics[width=0.15\textwidth]{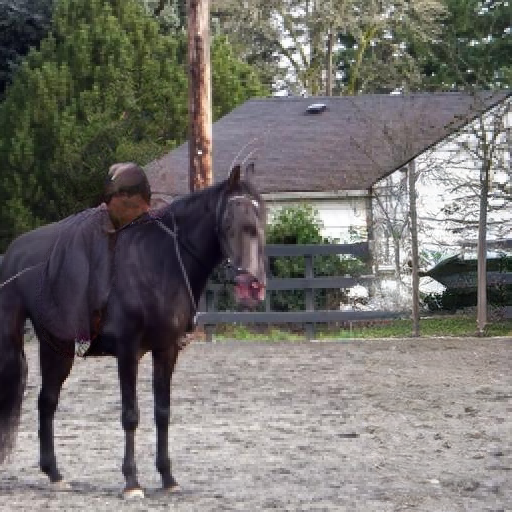}} &
{\includegraphics[width=0.15\textwidth]{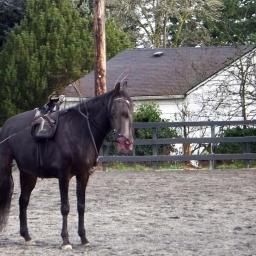}} &
{\includegraphics[width=0.15\textwidth]{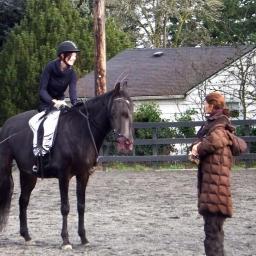}} \\

{\includegraphics[width=0.15\textwidth]{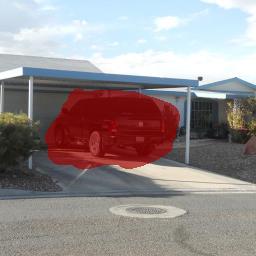}} &
{\includegraphics[width=0.15\textwidth]{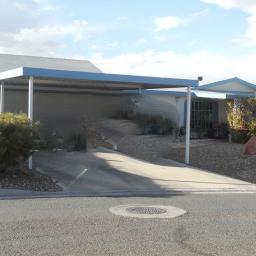}} &
{\includegraphics[width=0.15\textwidth]{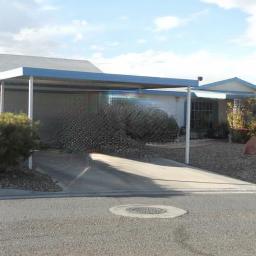}} &
{\includegraphics[width=0.15\textwidth]{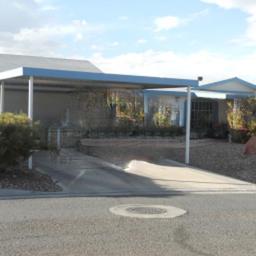}} &
{\includegraphics[width=0.15\textwidth]{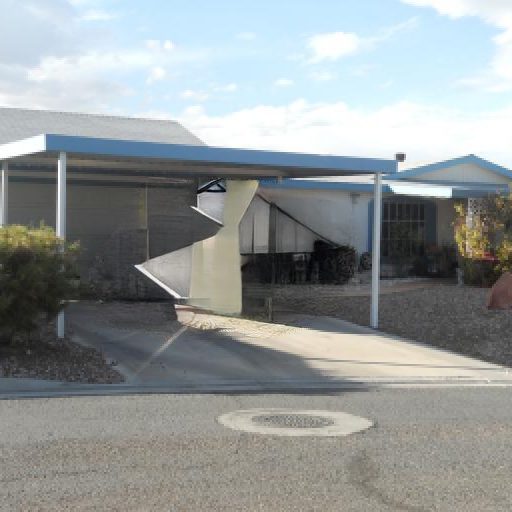}} &
{\includegraphics[width=0.15\textwidth]{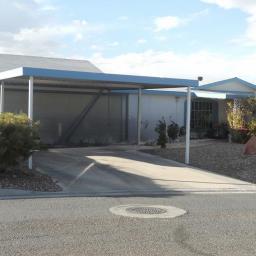}} &
{\includegraphics[width=0.15\textwidth]{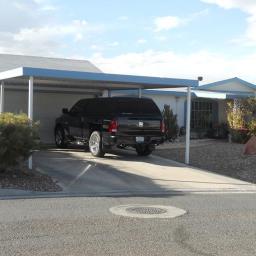}} \\

{\includegraphics[width=0.15\textwidth]{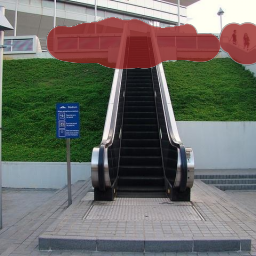}} &
{\includegraphics[width=0.15\textwidth]{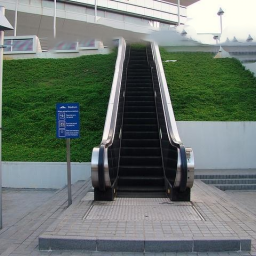}} &
{\includegraphics[width=0.15\textwidth]{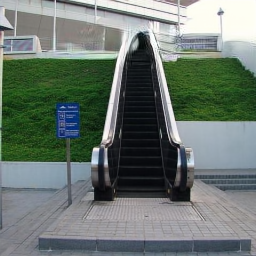}} &
{\includegraphics[width=0.15\textwidth]{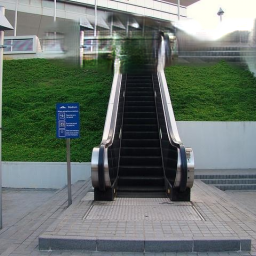}}&
{\includegraphics[width=0.15\textwidth]{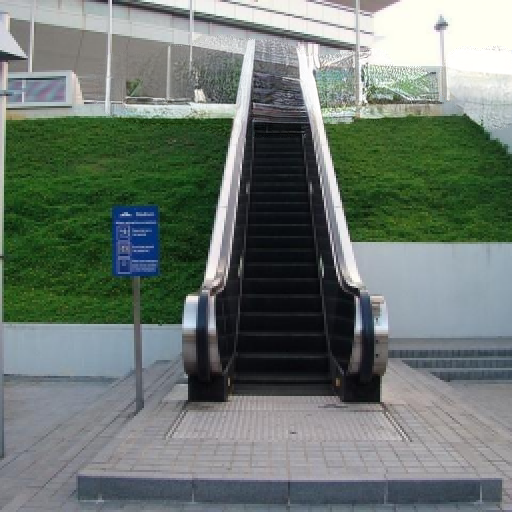}} &
{\includegraphics[width=0.15\textwidth]{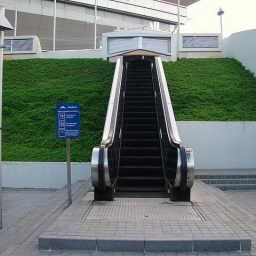}} &
{\includegraphics[width=0.15\textwidth]{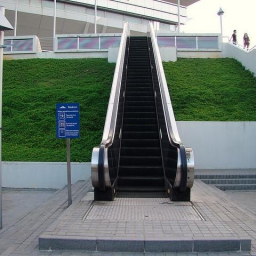}} \\

{\includegraphics[width=0.15\textwidth]{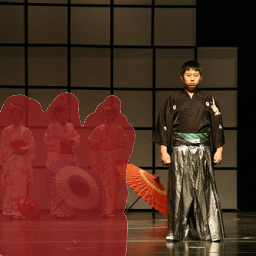}} & 
{\includegraphics[width=0.15\textwidth]{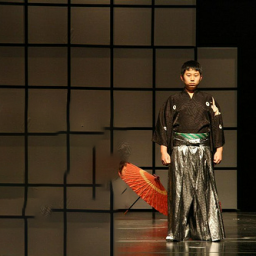}}&
{\includegraphics[width=0.15\textwidth]{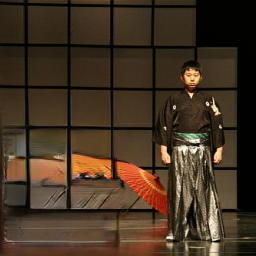}} & 
{\includegraphics[width=0.15\textwidth]{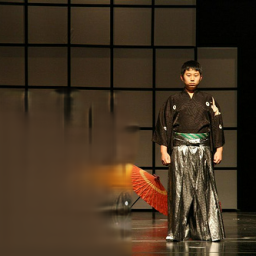}}&
{\includegraphics[width=0.15\textwidth]{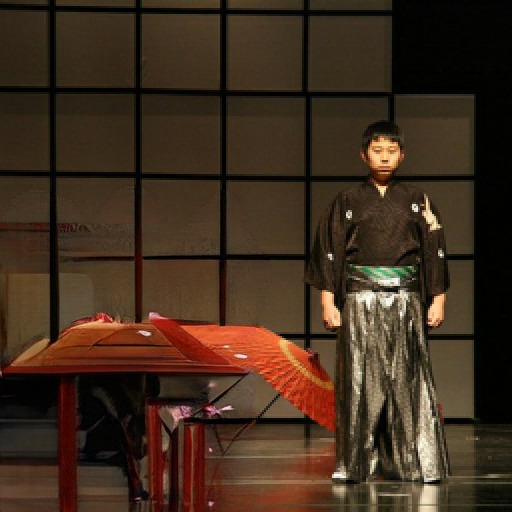}} &
{\includegraphics[width=0.15\textwidth]{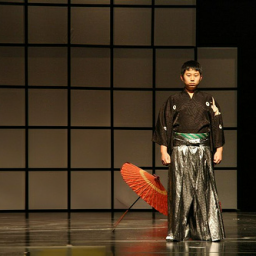}} &
{\includegraphics[width=0.15\textwidth]{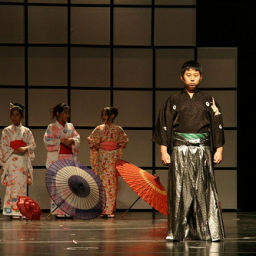}} \\


\end{tabular}
}
\end{center}
\vspace*{-0.5cm}
\caption{Comparison of inpainting methods on object removal. Baselines: \textsuperscript{\ddag}Photoshop's \textit{Content-aware Fill},  based on PatchMatch \citep{barnes2009PAR}, \textsuperscript{\textdagger}\citep{yu2019free}, \textsuperscript{\textdagger\textdagger}\citep{yi2020contextual} and \textsuperscript{\ddag\ddag}\citep{zhao2021large}.
\label{fig:inpainting_comparison_appendix}
\vspace{-.2cm}
}
\end{figure*}

\begin{figure*}[t]
\setlength{\tabcolsep}{1pt}
\begin{center}
{\small 
\begin{tabular}{cccccc}
{\small Masked Input} & {\small Sample 1}  & {\small Sample 2} & {\small Sample 3} &  {\small Sample 4} & {\small Original} 
\\
{\includegraphics[width=0.157\textwidth]{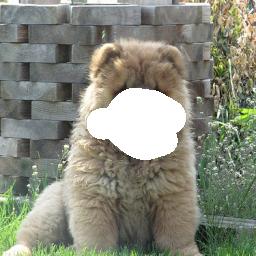}} &
{\includegraphics[width=0.157\textwidth]{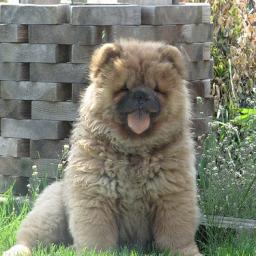}} &
{\includegraphics[width=0.157\textwidth]{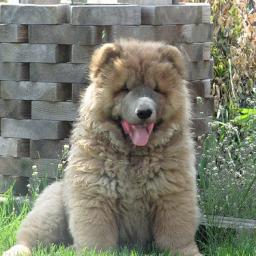}} &
{\includegraphics[width=0.157\textwidth]{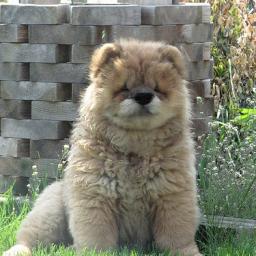}} &
{\includegraphics[width=0.157\textwidth]{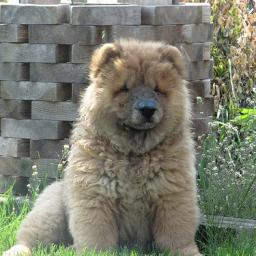}} &
{\includegraphics[width=0.157\textwidth]{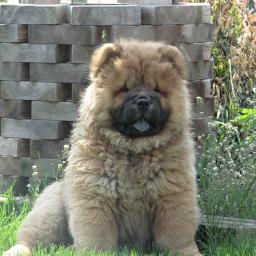}} \\

{\includegraphics[width=0.157\textwidth]{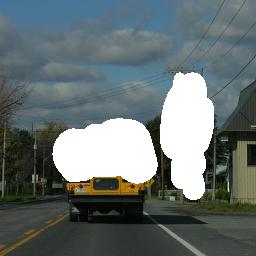}} &
{\includegraphics[width=0.157\textwidth]{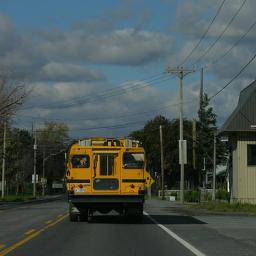}} &
{\includegraphics[width=0.157\textwidth]{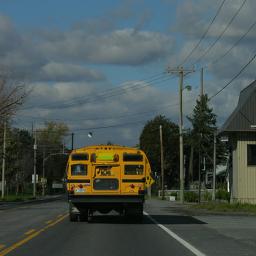}} &
{\includegraphics[width=0.157\textwidth]{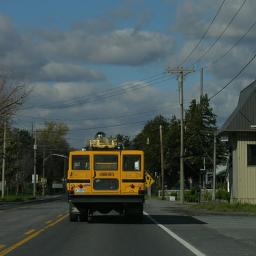}} &
{\includegraphics[width=0.157\textwidth]{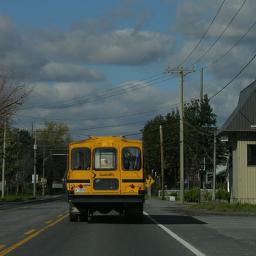}} &
{\includegraphics[width=0.157\textwidth]{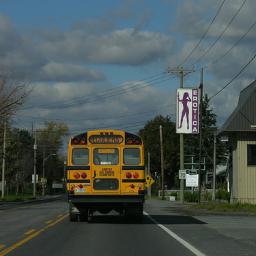}} \\

{\includegraphics[width=0.157\textwidth]{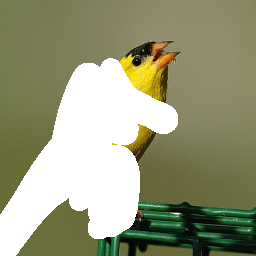}} &
{\includegraphics[width=0.157\textwidth]{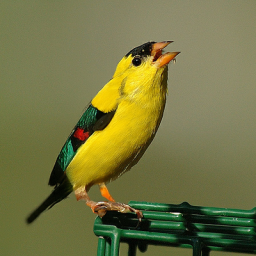}} &
{\includegraphics[width=0.157\textwidth]{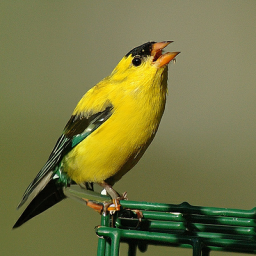}} &
{\includegraphics[width=0.157\textwidth]{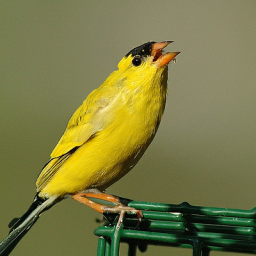}} &
{\includegraphics[width=0.157\textwidth]{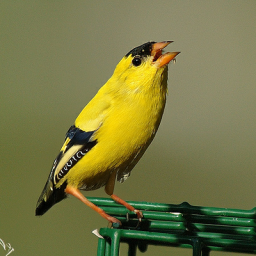}} &
{\includegraphics[width=0.157\textwidth]{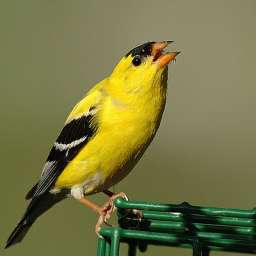}} \\

{\includegraphics[width=0.157\textwidth]{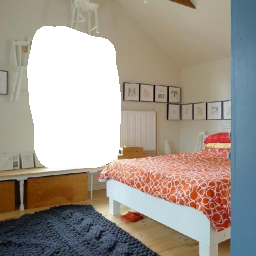}} &
{\includegraphics[width=0.157\textwidth]{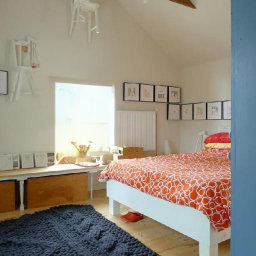}} &
{\includegraphics[width=0.157\textwidth]{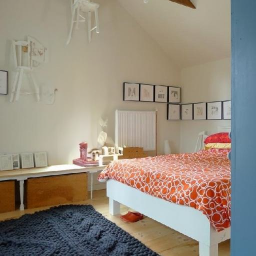}} &
{\includegraphics[width=0.157\textwidth]{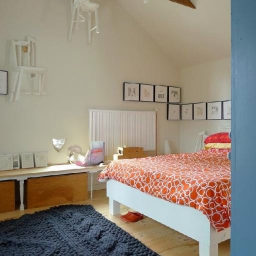}} &
{\includegraphics[width=0.157\textwidth]{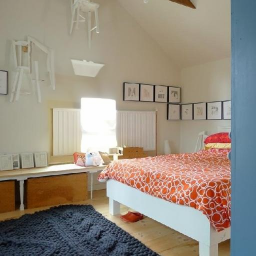}} &
{\includegraphics[width=0.157\textwidth]{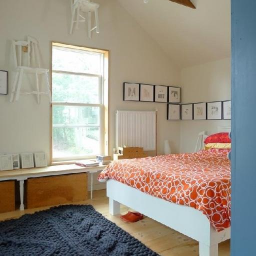}} \\

{\includegraphics[width=0.157\textwidth]{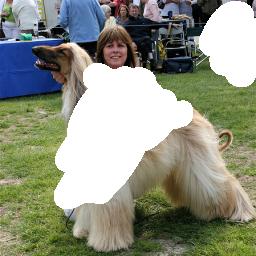}} &
{\includegraphics[width=0.157\textwidth]{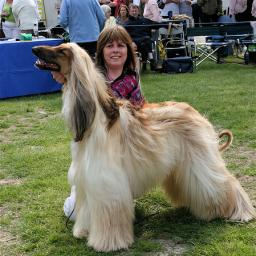}} &
{\includegraphics[width=0.157\textwidth]{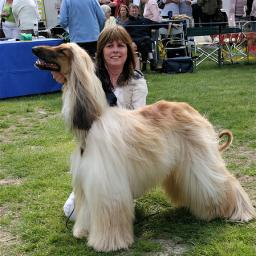}} &
{\includegraphics[width=0.157\textwidth]{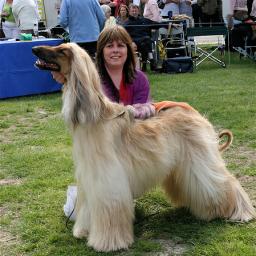}} &
{\includegraphics[width=0.157\textwidth]{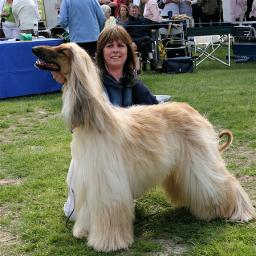}} &
{\includegraphics[width=0.157\textwidth]{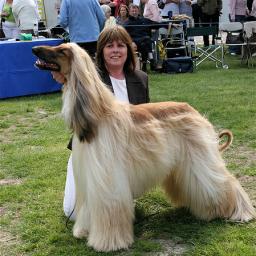}} \\

{\includegraphics[width=0.157\textwidth]{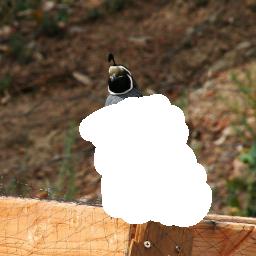}} &
{\includegraphics[width=0.157\textwidth]{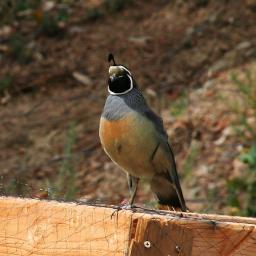}} &
{\includegraphics[width=0.157\textwidth]{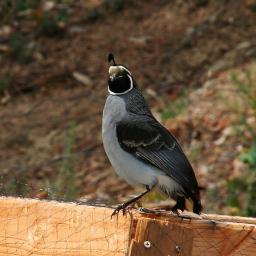}} &
{\includegraphics[width=0.157\textwidth]{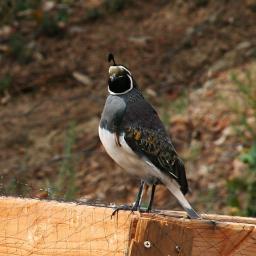}} &
{\includegraphics[width=0.157\textwidth]{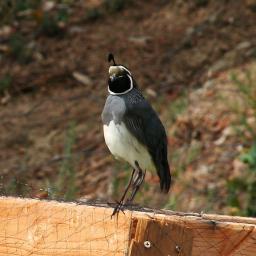}} &
{\includegraphics[width=0.157\textwidth]{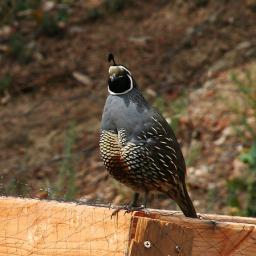}} \\

{\includegraphics[width=0.157\textwidth]{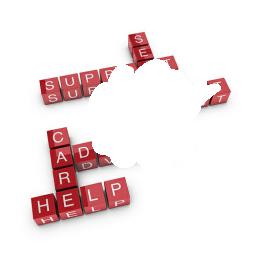}} &
{\includegraphics[width=0.157\textwidth]{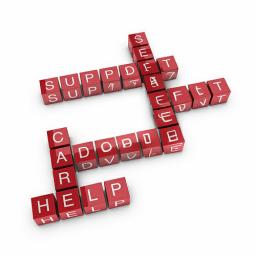}} &
{\includegraphics[width=0.157\textwidth]{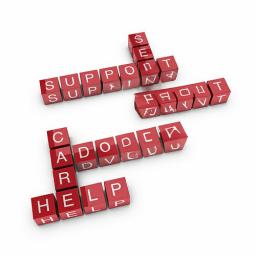}} &
{\includegraphics[width=0.157\textwidth]{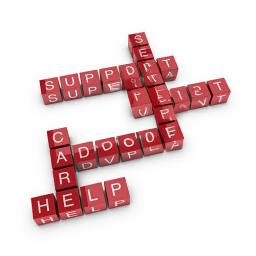}} &
{\includegraphics[width=0.157\textwidth]{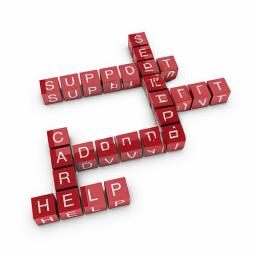}} &
{\includegraphics[width=0.157\textwidth]{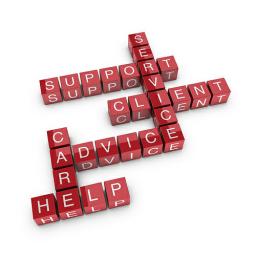}} \\


\end{tabular}
}
\end{center}
\vspace*{-0.6cm}
\caption{Diversity of \model outputs on image inpainting.}
\label{fig:inpainting_diversity_appendix}
\end{figure*}

\subsection{Uncropping}
\label{uncropping_appendix}
Many existing uncropping methods \citep{cheng2021out,teterwak2019boundless} have been trained on different subsets of Places2 \citep{zhou2017places} dataset. In order to maintain uniformity, we follow a similar setup as inpainting and train \model on a combined dataset of Places2 and ImageNet. While we train \model to extend the image in all directions or just one direction, to compare fairly against existing methods we evaluate \model on extending only the right half of the image. For Table \ref{tab:extrapolation_results}, we use ctest10k and places10k to report results on ImageNet and Places2 validation sets respectively.

We also perform category specific evaluation of \model with existing techniques - Boundless \citep{teterwak2019boundless} and InfinityGAN \citep{lin2021infinitygan}. Since Boundless is only trained on top-50 categories from Places2 dataset, we compare \model with Boundless specifically on these categories from Places2 validation set in Table \ref{tab:uncropping_boundless_comparison}. \model achieves significantly better performance compared to Boundless re-affirming the strength of our model. Furthermore, we compare \model with a more recent GAN based uncropping technique - InfinityGAN 
\citep{lin2021infinitygan}. In order to fairly compare \model with InfinityGAN, we specifically evaluate on the scenery categories from Places2 validation and test set. We use the samples generously provided by \cite{lin2021infinitygan}, and generate outputs for Boundless, and \model. Table \ref{tab:uncropping_infinitygan_comparison} shows that \model is significantly better than domain specific model InfinityGAN on scenery images in terms of automated metrics. 

\begin{table}[t]
    \centering
    {\small
    \begin{tabular}{lcc}
    \toprule
    \bfseries{Model} & \bfseries{FID} $\downarrow$ & \bfseries{PD} $\downarrow$ \\
    \midrule
    Boundless \citep{teterwak2019boundless} & 28.3  & 115.0  \\
    \model & \textbf{22.9} & \textbf{93.4}   \\
    \midrule
    Ground Truth & 23.6  & 0.0  \\
    \bottomrule
    \end{tabular}
    }
    \vspace*{0.1cm}
    \caption{Comparison with uncropping method Boundless \citep{teterwak2019boundless} on top-50 Places2 categories.}
    \label{tab:uncropping_boundless_comparison}
\vspace{-.6cm}    
\end{table}

\begin{table}[t]
    \centering
    {\small
    \begin{tabular}{lc}
    \toprule
    \bfseries{Model} & \bfseries{FID} $\downarrow$ \\
    \midrule
    Boundless \citep{teterwak2019boundless} & 12.7   \\
    InfinityGAN \citep{lin2021infinitygan} & 15.7 \\
    \model & \textbf{5.6}  \\
    \bottomrule
    \end{tabular}
    }
    \vspace*{0.1cm}
    \caption{Comparison with uncropping method InfinityGAN \citep{lin2021infinitygan} and Boundless \citep{teterwak2019boundless} on scenery categories.}
    \label{tab:uncropping_infinitygan_comparison}
    \vspace{-0.6cm}
\end{table}

\begin{figure*}[h]
\setlength{\tabcolsep}{2pt}
\begin{center}
{\small 
\begin{tabular}{ccccc}
{\small Input} & {\small Unconditional}  & {\small Multi-Task} & {\small Task Specific} &  {\small Original}
\\
\frame{\includegraphics[width=0.157\textwidth]{figs/uncond_inpaint/src_0.png}} &
\frame{\includegraphics[width=0.157\textwidth]{figs/uncond_inpaint/uncond_0.png}} &
\frame{\includegraphics[width=0.157\textwidth]{figs/uncond_inpaint/multitask_0.png}} &
\frame{\includegraphics[width=0.157\textwidth]{figs/uncond_inpaint/inpaint_0.png}} &
\frame{\includegraphics[width=0.157\textwidth]{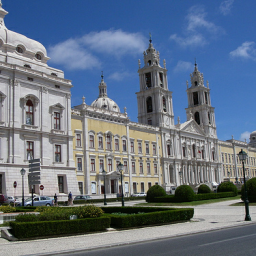}} \\

\frame{\includegraphics[width=0.157\textwidth]{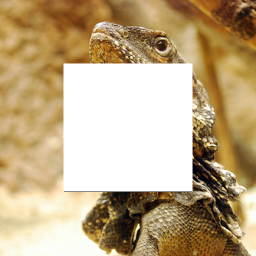}} &
\frame{\includegraphics[width=0.157\textwidth]{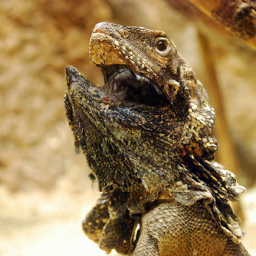}} &
\frame{\includegraphics[width=0.157\textwidth]{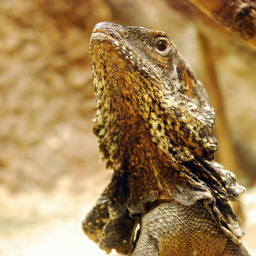}} &
\frame{\includegraphics[width=0.157\textwidth]{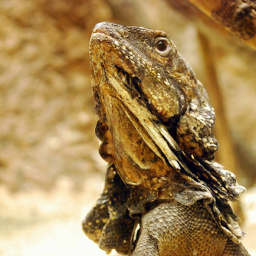}} &
\frame{\includegraphics[width=0.157\textwidth]{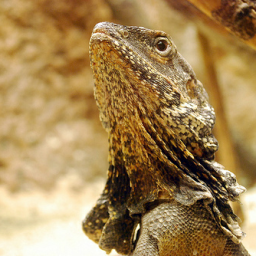}} \\

\frame{\includegraphics[width=0.157\textwidth]{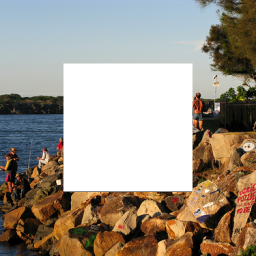}} &
\frame{\includegraphics[width=0.157\textwidth]{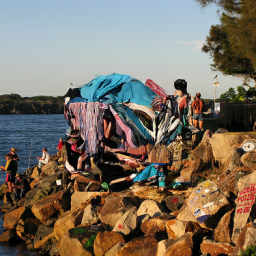}} &
\frame{\includegraphics[width=0.157\textwidth]{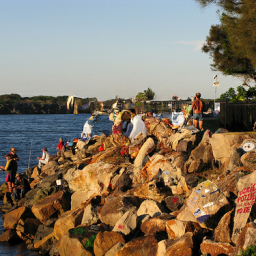}} &
\frame{\includegraphics[width=0.157\textwidth]{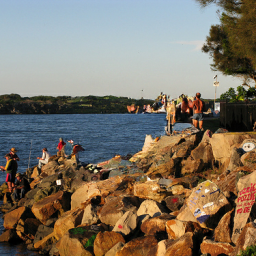}} &
\frame{\includegraphics[width=0.157\textwidth]{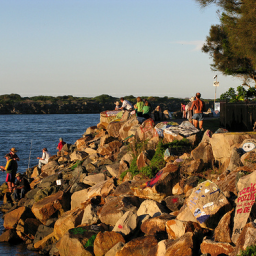}} \\

\frame{\includegraphics[width=0.157\textwidth]{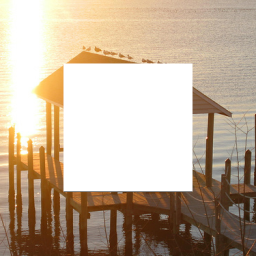}} &
\frame{\includegraphics[width=0.157\textwidth]{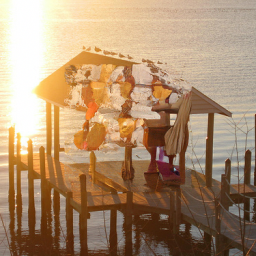}} &
\frame{\includegraphics[width=0.157\textwidth]{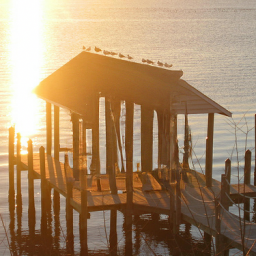}} &
\frame{\includegraphics[width=0.157\textwidth]{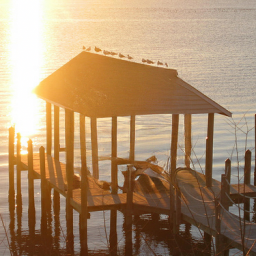}} &
\frame{\includegraphics[width=0.157\textwidth]{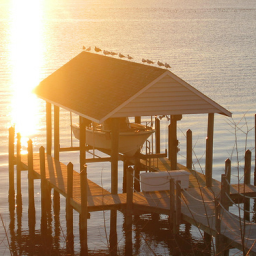}} \\


\frame{\includegraphics[width=0.157\textwidth]{figs/uncond_inpaint/src_5.png}} &
\frame{\includegraphics[width=0.157\textwidth]{figs/uncond_inpaint/uncond_5.png}} &
\frame{\includegraphics[width=0.157\textwidth]{figs/uncond_inpaint/multitask_5.png}} &
\frame{\includegraphics[width=0.157\textwidth]{figs/uncond_inpaint/inpaint_5.png}} &
\frame{\includegraphics[width=0.157\textwidth]{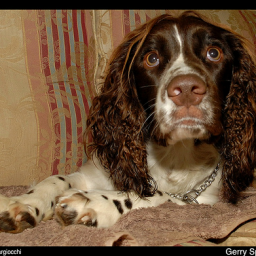}} \\

\frame{\includegraphics[width=0.157\textwidth]{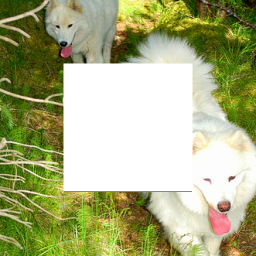}} &
\frame{\includegraphics[width=0.157\textwidth]{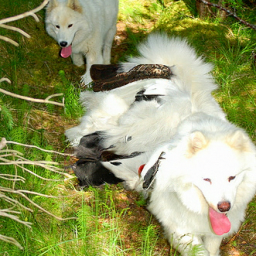}} &
\frame{\includegraphics[width=0.157\textwidth]{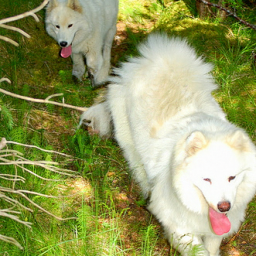}} &
\frame{\includegraphics[width=0.157\textwidth]{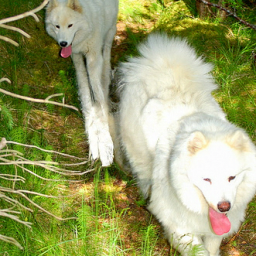}} &
\frame{\includegraphics[width=0.157\textwidth]{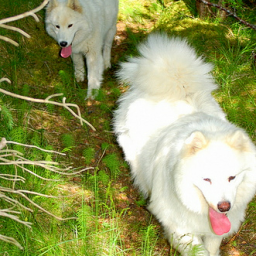}} \\

\frame{\includegraphics[width=0.157\textwidth]{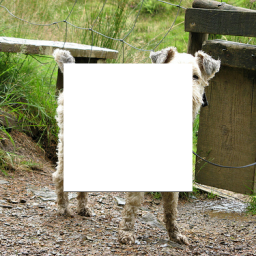}} &
\frame{\includegraphics[width=0.157\textwidth]{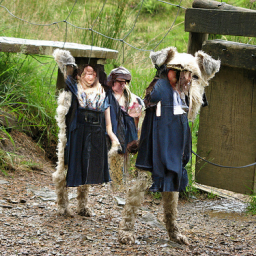}} &
\frame{\includegraphics[width=0.157\textwidth]{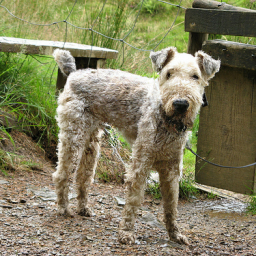}} &
\frame{\includegraphics[width=0.157\textwidth]{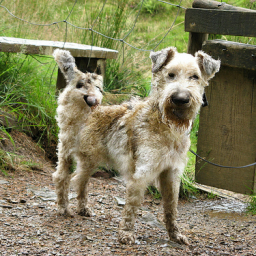}} &
\frame{\includegraphics[width=0.157\textwidth]{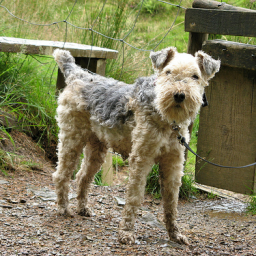}} \\

\end{tabular}
}
\end{center}
\vspace*{-0.5cm}
\caption{Comparison between an unconditional model repurposed for the task of inpainting \cite{song-iclr-2021}, a multi-task model trained on all four tasks, and an inpainting task specific model.
\label{fig:uncond_compare_inpaint}
}
\end{figure*}

\textbf{Human Evaluation:} Like colorization, we also report results from
human evaluation experiments.
Obtaining high fool rates for uncropping is a significantly more challenging task than colorization, because one half of the image area is fully generated by the model.  As a consequence there are more opportunities for synthetic artifacts. 
Because the baselines available for uncropping are trained and tested on Places2, we run human evaluation experiments only on Places2.
Beyond the choice of dataset, all other aspects of experimental design are identical to that used above for colorization, with two disjoint sets of test images, namely, Set-I and Set-II.  

\begin{figure*}[t]
\setlength{\tabcolsep}{2pt}
\begin{center}
\begin{tabular}{ccccccc}
{\small Masked Input} &
{\small Boundless\textsuperscript{\textdagger}} &{\small InfinityGAN \textsuperscript{\textdagger\textdagger}} & {\small \model (Ours)} & {\small Original} 
\\


\frame{\includegraphics[width=0.15\textwidth]{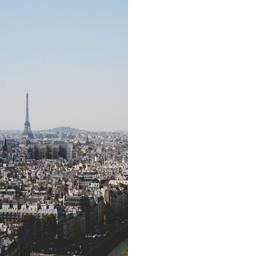}} &
\frame{\includegraphics[width=0.15\textwidth]{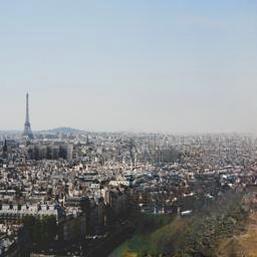}} &
\frame{\includegraphics[width=0.15\textwidth]{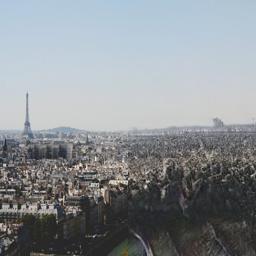}} &
\frame{\includegraphics[width=0.15\textwidth]{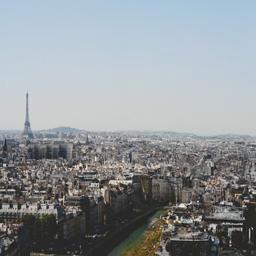}} &
\frame{\includegraphics[width=0.15\textwidth]{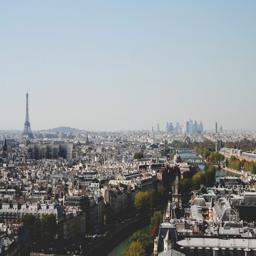}} \\

\frame{\includegraphics[width=0.15\textwidth]{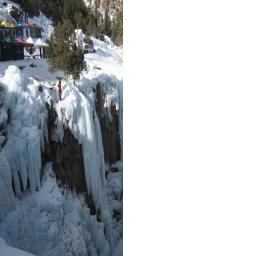}} &
\frame{\includegraphics[width=0.15\textwidth]{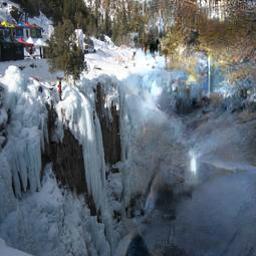}} &
\frame{\includegraphics[width=0.15\textwidth]{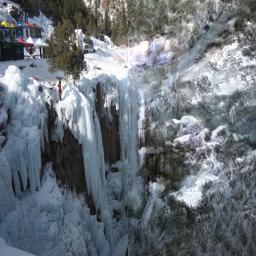}} &
\frame{\includegraphics[width=0.15\textwidth]{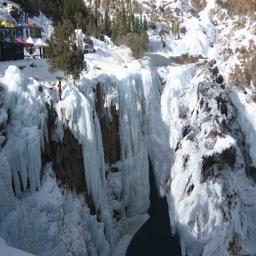}} &
\frame{\includegraphics[width=0.15\textwidth]{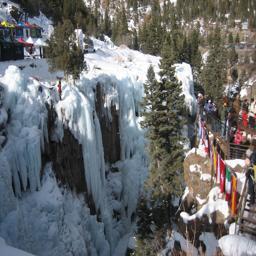}} \\

\frame{\includegraphics[width=0.15\textwidth]{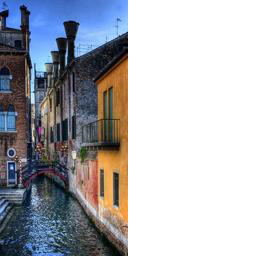}} &
\frame{\includegraphics[width=0.15\textwidth]{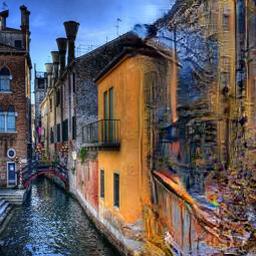}} &
\frame{\includegraphics[width=0.15\textwidth]{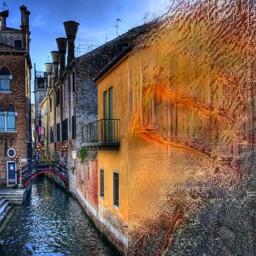}} &
\frame{\includegraphics[width=0.15\textwidth]{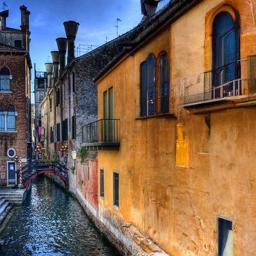}} &
\frame{\includegraphics[width=0.15\textwidth]{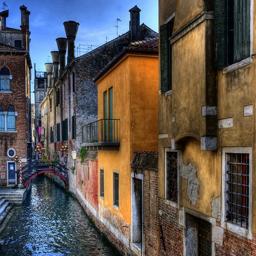}} \\

\frame{\includegraphics[width=0.15\textwidth]{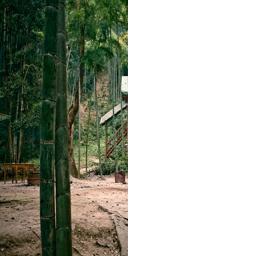}} &
\frame{\includegraphics[width=0.15\textwidth]{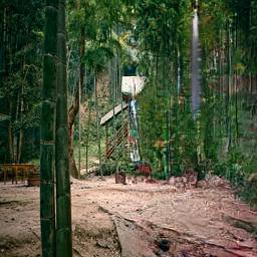}} &
\frame{\includegraphics[width=0.15\textwidth]{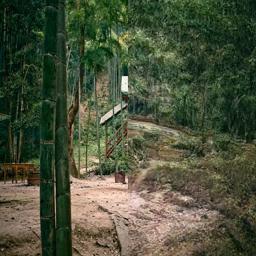}} &
\frame{\includegraphics[width=0.15\textwidth]{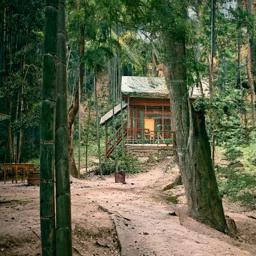}} &
\frame{\includegraphics[width=0.15\textwidth]{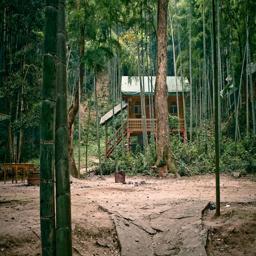}} \\


\frame{\includegraphics[width=0.15\textwidth]{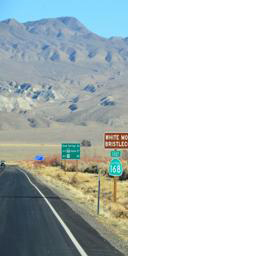}} &
\frame{\includegraphics[width=0.15\textwidth]{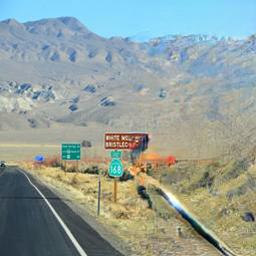}} &
\frame{\includegraphics[width=0.15\textwidth]{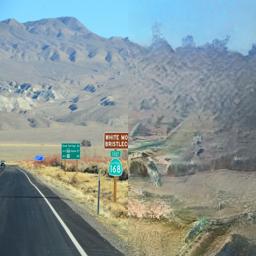}} &
\frame{\includegraphics[width=0.15\textwidth]{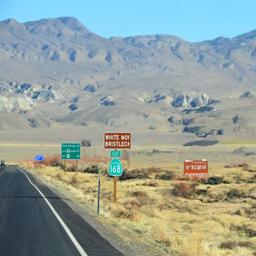}} &
\frame{\includegraphics[width=0.15\textwidth]{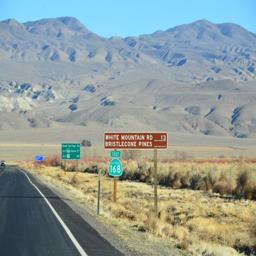}} \\

\frame{\includegraphics[width=0.15\textwidth]{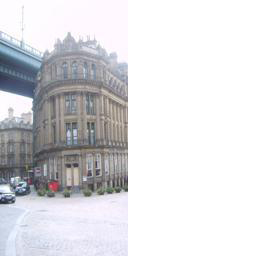}} &
\frame{\includegraphics[width=0.15\textwidth]{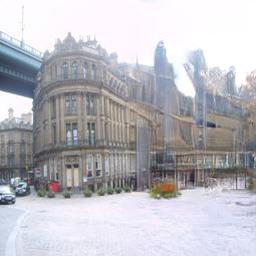}} &
\frame{\includegraphics[width=0.15\textwidth]{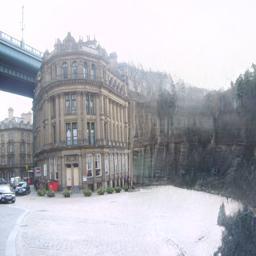}} &
\frame{\includegraphics[width=0.15\textwidth]{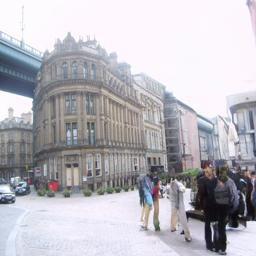}} &
\frame{\includegraphics[width=0.15\textwidth]{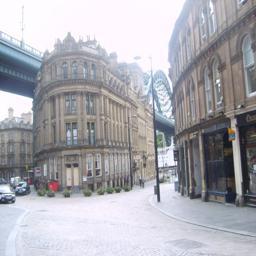}} \\

\end{tabular}
\end{center}
\vspace*{-0.4cm}
\caption{Image uncropping results on Places2 validation images. Baselines:  Boundless\textsuperscript{\textdagger} \citep{teterwak2019boundless} and  InfinityGAN\textsuperscript{\textdagger\textdagger} \citep{lin2021infinitygan} trained on a scenery subset of Places2. Samples for both baselines are generously provided by their respective authors. 
\label{fig:extrapolation_comparison_appendix}
}
\end{figure*}

The results are characterized in terms of the fool rate, and are shown in Figure \ref{fig:fool_rates_uncropping}.
\model obtains significantly higher fool rates on all human evaluation runs compared to existing methods, i.e., Boundless \citep{teterwak2019boundless} and InfinityGAN \citep{lin2021infinitygan}. Interestingly, when raters are given more time to inspect each pair of images, the fool rates for InfinityGAN and Boundless worsen considerably. \model, on the other hand, observes approximately similar fool rates.

\begin{figure*}[h]
\setlength{\tabcolsep}{2pt}
\begin{center}
{\small 
\begin{tabular}{cccccc}
{\small Input} & {\small Sample 1}  & {\small Sample 2} & {\small Sample 3} &  {\small Sample 4} & {Original} 
\\
\frame{\includegraphics[width=0.15\textwidth]{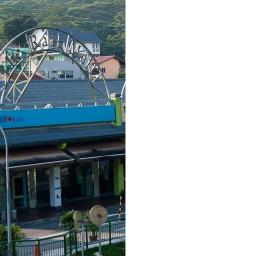}} &
\frame{\includegraphics[width=0.15\textwidth]{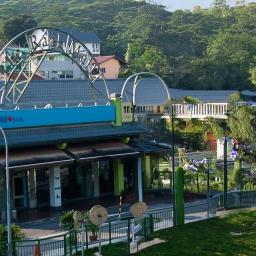}} &
\frame{\includegraphics[width=0.15\textwidth]{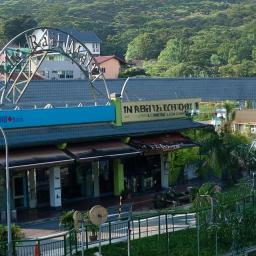}} &
\frame{\includegraphics[width=0.15\textwidth]{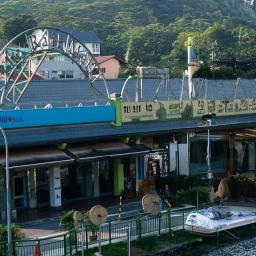}} &
\frame{\includegraphics[width=0.15\textwidth]{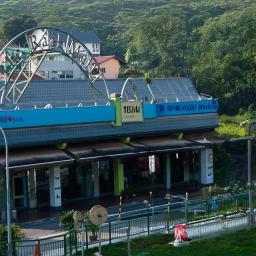}} &
\frame{\includegraphics[width=0.15\textwidth]{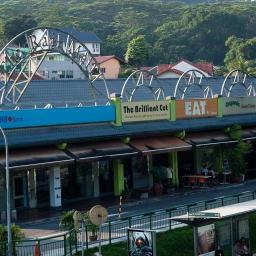}}\\

\frame{\includegraphics[width=0.15\textwidth]{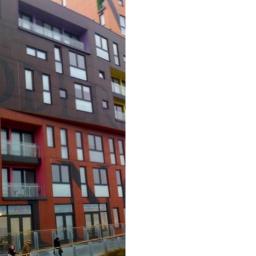}} &
\frame{\includegraphics[width=0.15\textwidth]{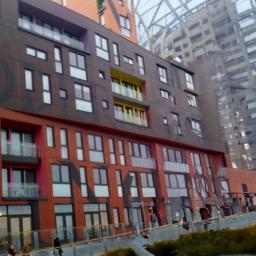}} &
\frame{\includegraphics[width=0.15\textwidth]{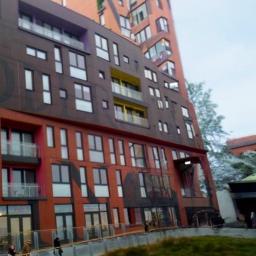}} &
\frame{\includegraphics[width=0.15\textwidth]{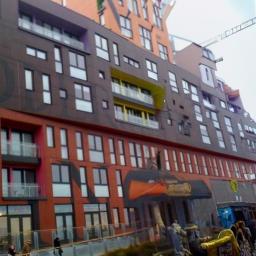}} &
\frame{\includegraphics[width=0.15\textwidth]{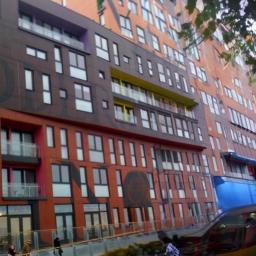}} &
\frame{\includegraphics[width=0.15\textwidth]{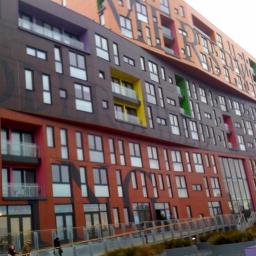}}\\

\frame{\includegraphics[width=0.15\textwidth]{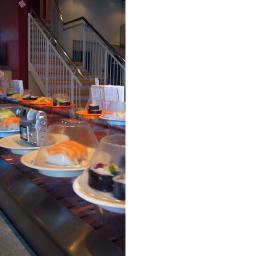}} &
\frame{\includegraphics[width=0.15\textwidth]{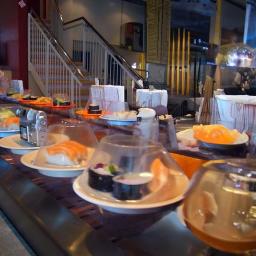}} &
\frame{\includegraphics[width=0.15\textwidth]{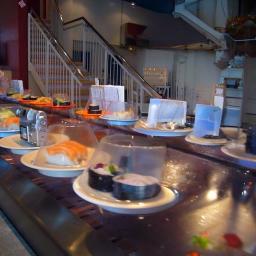}} &
\frame{\includegraphics[width=0.15\textwidth]{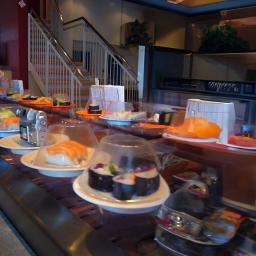}} &
\frame{\includegraphics[width=0.15\textwidth]{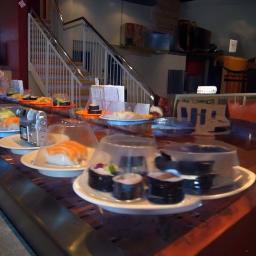}} &
\frame{\includegraphics[width=0.15\textwidth]{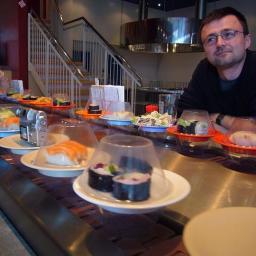}}\\

\frame{\includegraphics[width=0.15\textwidth]{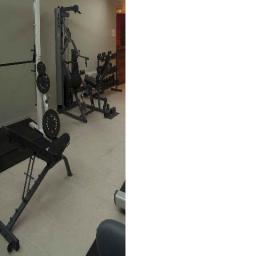}} &
\frame{\includegraphics[width=0.15\textwidth]{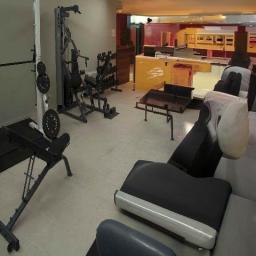}} &
\frame{\includegraphics[width=0.15\textwidth]{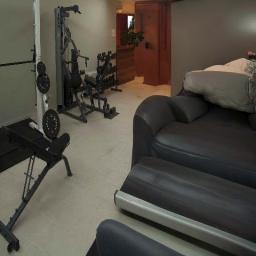}} &
\frame{\includegraphics[width=0.15\textwidth]{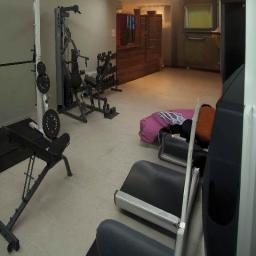}} &
\frame{\includegraphics[width=0.15\textwidth]{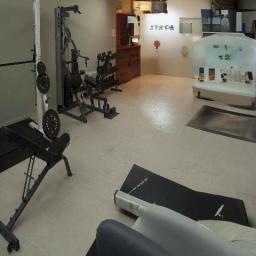}} &
\frame{\includegraphics[width=0.15\textwidth]{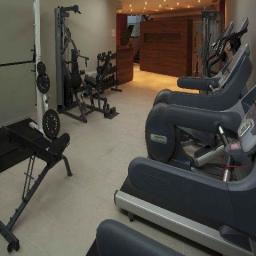}}\\

\frame{\includegraphics[width=0.15\textwidth]{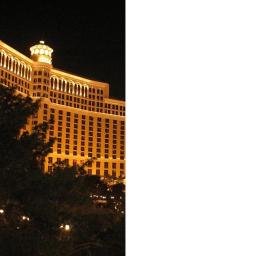}} &
\frame{\includegraphics[width=0.15\textwidth]{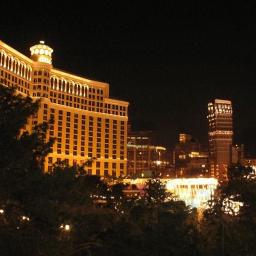}} &
\frame{\includegraphics[width=0.15\textwidth]{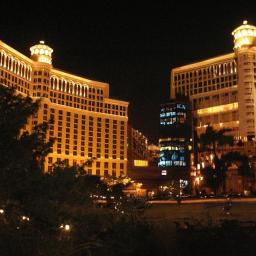}} &
\frame{\includegraphics[width=0.15\textwidth]{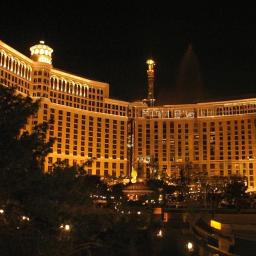}} &
\frame{\includegraphics[width=0.15\textwidth]{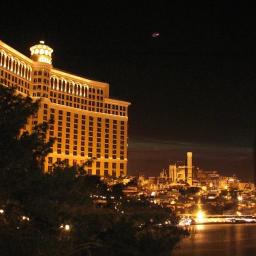}} &
\frame{\includegraphics[width=0.15\textwidth]{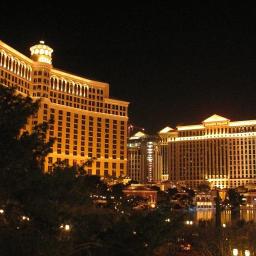}}\\

\frame{\includegraphics[width=0.15\textwidth]{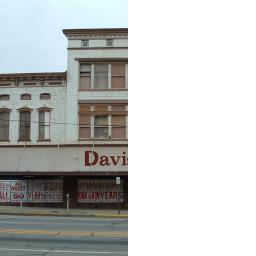}} &
\frame{\includegraphics[width=0.15\textwidth]{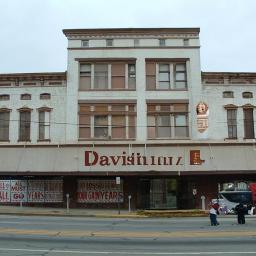}} &
\frame{\includegraphics[width=0.15\textwidth]{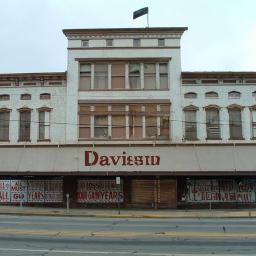}} &
\frame{\includegraphics[width=0.15\textwidth]{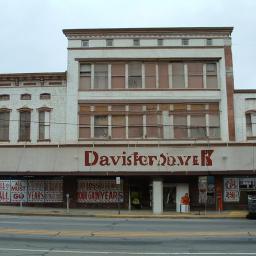}} &
\frame{\includegraphics[width=0.15\textwidth]{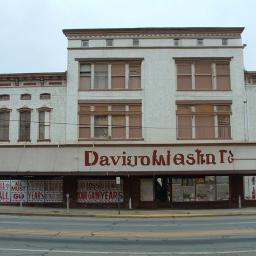}} &
\frame{\includegraphics[width=0.15\textwidth]{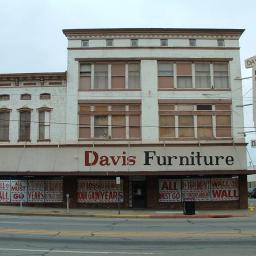}}\\

\frame{\includegraphics[width=0.15\textwidth]{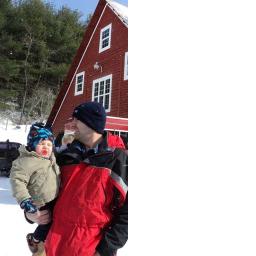}} &
\frame{\includegraphics[width=0.15\textwidth]{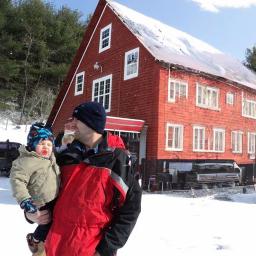}} &
\frame{\includegraphics[width=0.15\textwidth]{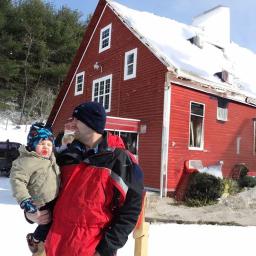}} &
\frame{\includegraphics[width=0.15\textwidth]{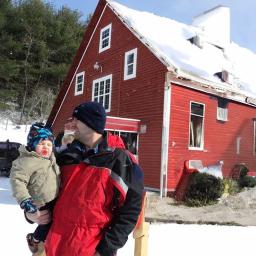}} &
\frame{\includegraphics[width=0.15\textwidth]{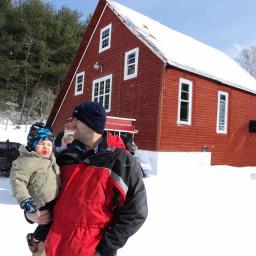}} &
\frame{\includegraphics[width=0.15\textwidth]{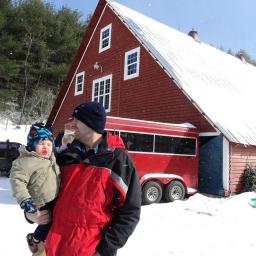}}\\


\end{tabular}
}
\end{center}
\vspace*{-0.5cm}
\caption{Diversity of \model outputs on Right Uncropping on Places2 dataset.
\label{fig:uncropping_diversity_appendix_1}
}
\end{figure*}

\begin{figure*}[h]
\setlength{\tabcolsep}{2pt}
\begin{center}
{\small 
\begin{tabular}{cccccc}
{\small Input} & {\small Sample 1}  & {\small Sample 2} & {\small Sample 3} &  {\small Sample 4} & {Original} 
\\
\frame{\includegraphics[width=0.15\textwidth]{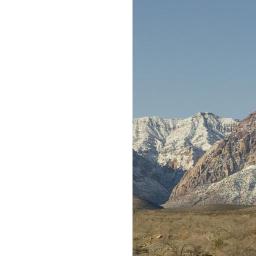}} &
\frame{\includegraphics[width=0.15\textwidth]{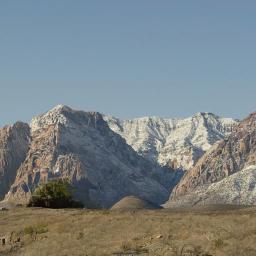}} &
\frame{\includegraphics[width=0.15\textwidth]{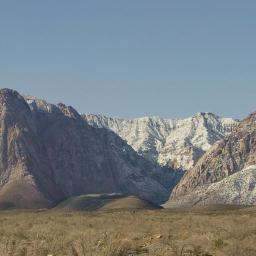}} &
\frame{\includegraphics[width=0.15\textwidth]{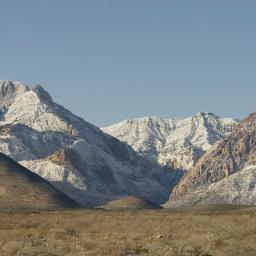}} &
\frame{\includegraphics[width=0.15\textwidth]{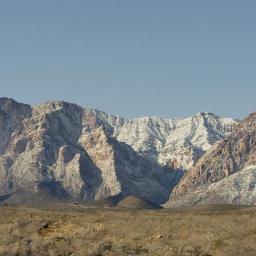}} &
\frame{\includegraphics[width=0.15\textwidth]{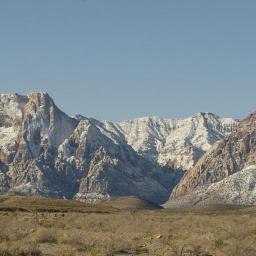}}\\

\frame{\includegraphics[width=0.15\textwidth]{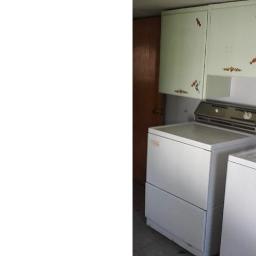}} &
\frame{\includegraphics[width=0.15\textwidth]{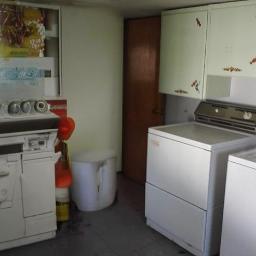}} &
\frame{\includegraphics[width=0.15\textwidth]{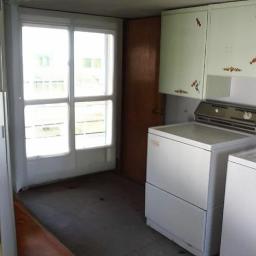}} &
\frame{\includegraphics[width=0.15\textwidth]{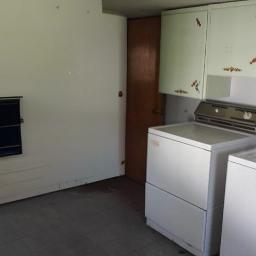}} &
\frame{\includegraphics[width=0.15\textwidth]{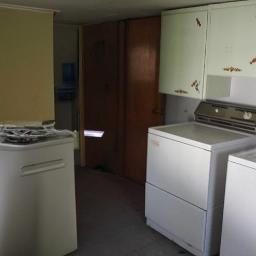}} &
\frame{\includegraphics[width=0.15\textwidth]{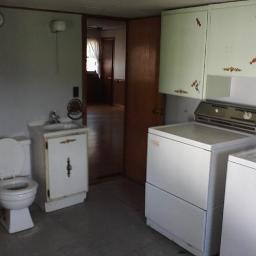}}\\

\frame{\includegraphics[width=0.15\textwidth]{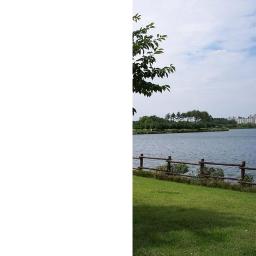}} &
\frame{\includegraphics[width=0.15\textwidth]{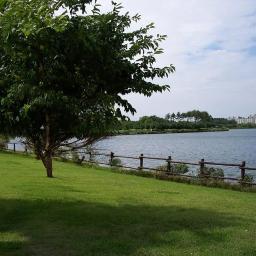}} &
\frame{\includegraphics[width=0.15\textwidth]{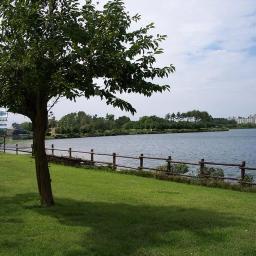}} &
\frame{\includegraphics[width=0.15\textwidth]{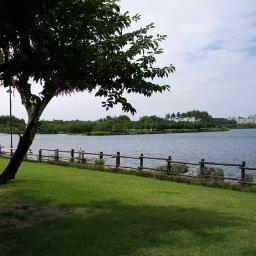}} &
\frame{\includegraphics[width=0.15\textwidth]{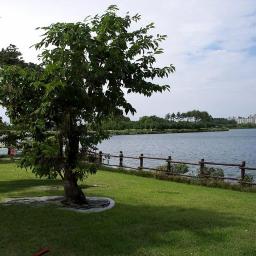}} &
\frame{\includegraphics[width=0.15\textwidth]{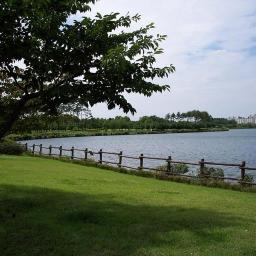}}\\

\frame{\includegraphics[width=0.15\textwidth]{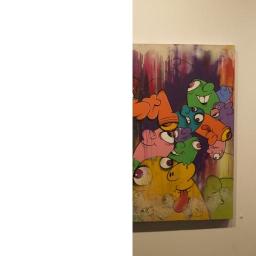}} &
\frame{\includegraphics[width=0.15\textwidth]{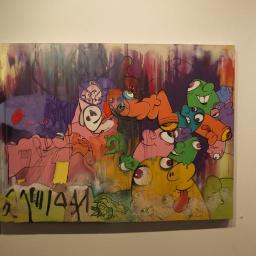}} &
\frame{\includegraphics[width=0.15\textwidth]{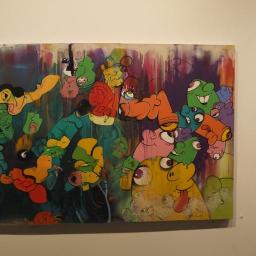}} &
\frame{\includegraphics[width=0.15\textwidth]{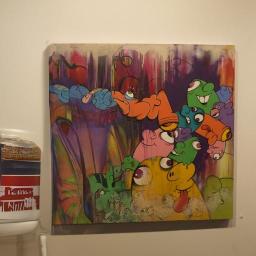}} &
\frame{\includegraphics[width=0.15\textwidth]{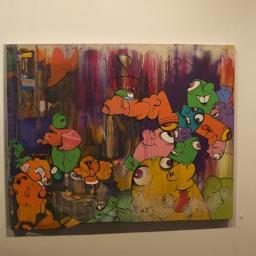}} &
\frame{\includegraphics[width=0.15\textwidth]{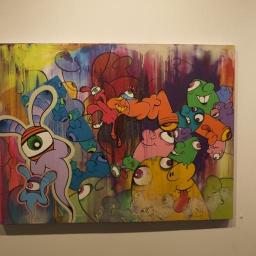}}\\

\frame{\includegraphics[width=0.15\textwidth]{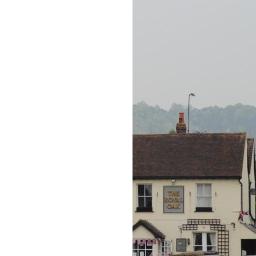}} &
\frame{\includegraphics[width=0.15\textwidth]{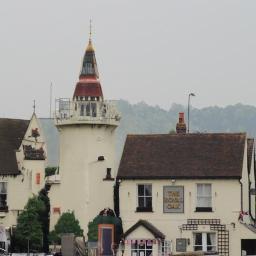}} &
\frame{\includegraphics[width=0.15\textwidth]{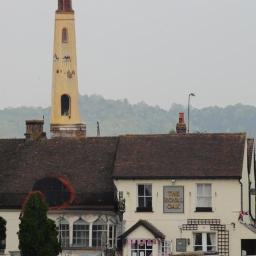}} &
\frame{\includegraphics[width=0.15\textwidth]{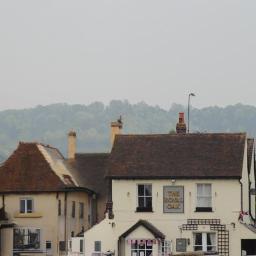}} &
\frame{\includegraphics[width=0.15\textwidth]{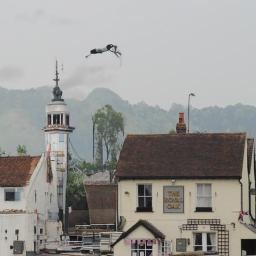}} &
\frame{\includegraphics[width=0.15\textwidth]{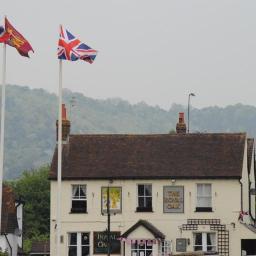}}\\

\frame{\includegraphics[width=0.15\textwidth]{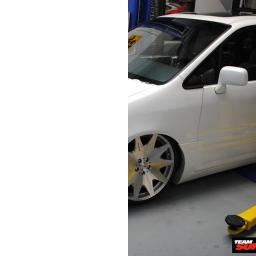}} &
\frame{\includegraphics[width=0.15\textwidth]{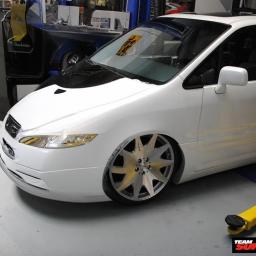}} &
\frame{\includegraphics[width=0.15\textwidth]{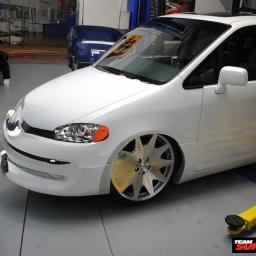}} &
\frame{\includegraphics[width=0.15\textwidth]{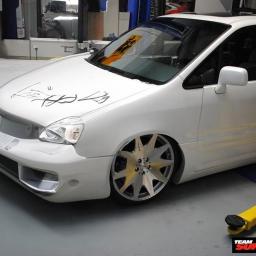}} &
\frame{\includegraphics[width=0.15\textwidth]{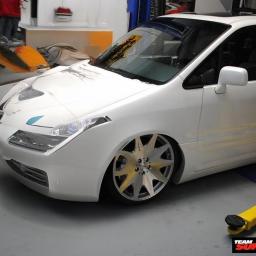}} &
\frame{\includegraphics[width=0.15\textwidth]{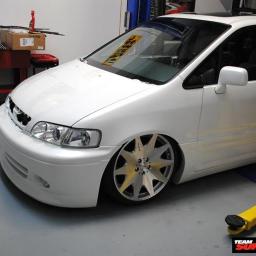}}\\

\frame{\includegraphics[width=0.15\textwidth]{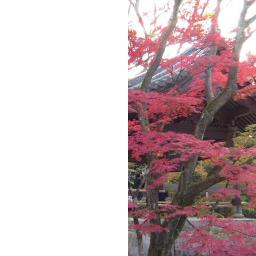}} &
\frame{\includegraphics[width=0.15\textwidth]{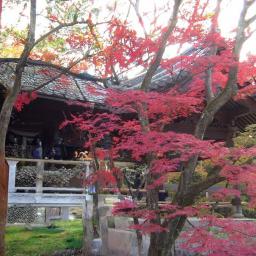}} &
\frame{\includegraphics[width=0.15\textwidth]{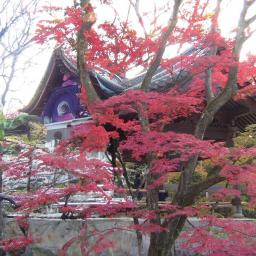}} &
\frame{\includegraphics[width=0.15\textwidth]{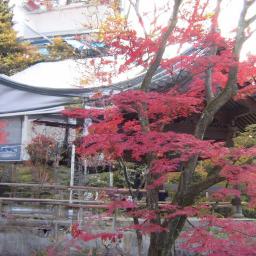}} &
\frame{\includegraphics[width=0.15\textwidth]{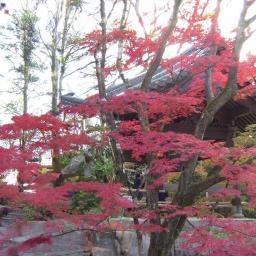}} &
\frame{\includegraphics[width=0.15\textwidth]{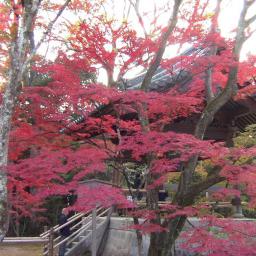}}\\


\end{tabular}
}
\end{center}
\vspace*{-0.5cm}
\caption{Diversity of \model outputs on Left uncropping on Places2 dataset.
\label{fig:uncropping_diversity_appendix_2}
}
\end{figure*}

\begin{figure*}[t]
\setlength{\tabcolsep}{2pt}
\begin{center}
{\small 
\begin{tabular}{cccccc}
{\small Input} & {\small Sample 1}  & {\small Sample 2} & {\small Sample 3} &  {\small Sample 4} & {Original} 
\\
\frame{\includegraphics[width=0.15\textwidth]{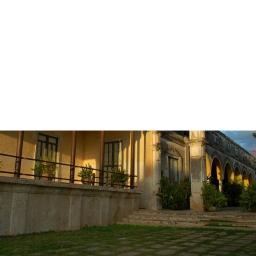}} &
\frame{\includegraphics[width=0.15\textwidth]{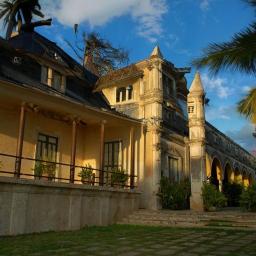}} &
\frame{\includegraphics[width=0.15\textwidth]{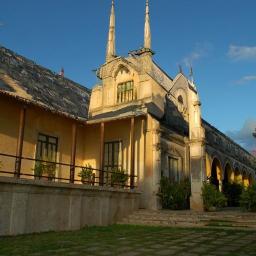}} &
\frame{\includegraphics[width=0.15\textwidth]{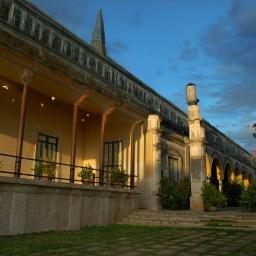}} &
\frame{\includegraphics[width=0.15\textwidth]{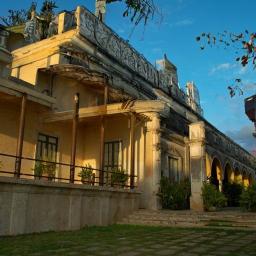}} &
\frame{\includegraphics[width=0.15\textwidth]{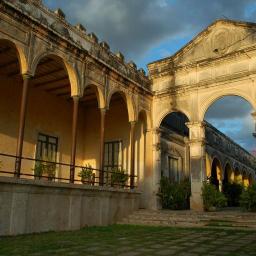}}\\

\frame{\includegraphics[width=0.15\textwidth]{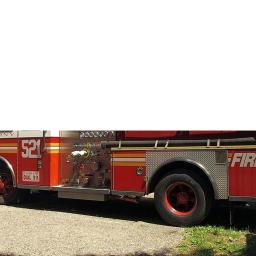}} &
\frame{\includegraphics[width=0.15\textwidth]{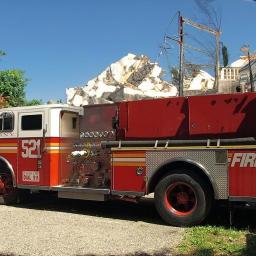}} &
\frame{\includegraphics[width=0.15\textwidth]{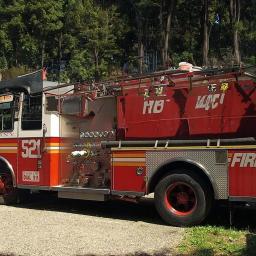}} &
\frame{\includegraphics[width=0.15\textwidth]{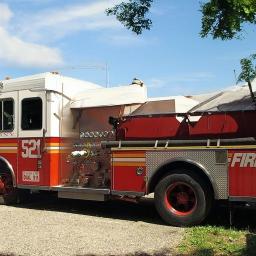}} &
\frame{\includegraphics[width=0.15\textwidth]{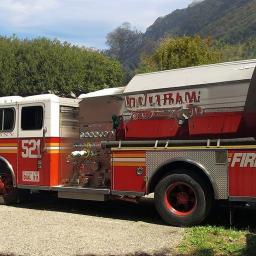}} &
\frame{\includegraphics[width=0.15\textwidth]{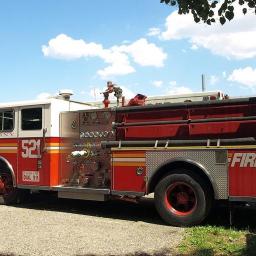}}\\

\frame{\includegraphics[width=0.15\textwidth]{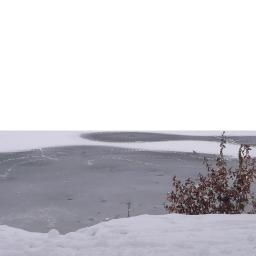}} &
\frame{\includegraphics[width=0.15\textwidth]{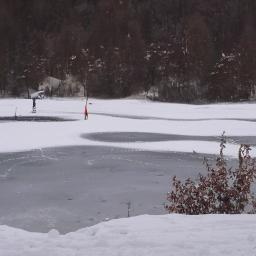}} &
\frame{\includegraphics[width=0.15\textwidth]{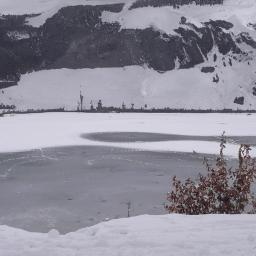}} &
\frame{\includegraphics[width=0.15\textwidth]{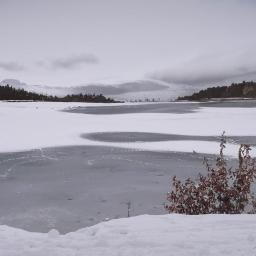}} &
\frame{\includegraphics[width=0.15\textwidth]{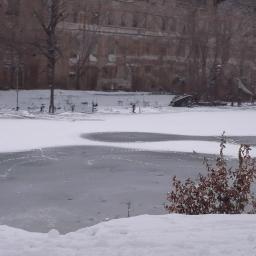}} &
\frame{\includegraphics[width=0.15\textwidth]{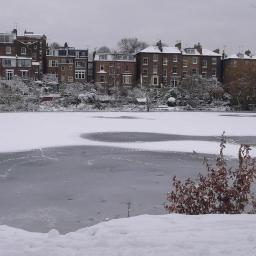}}\\

\frame{\includegraphics[width=0.15\textwidth]{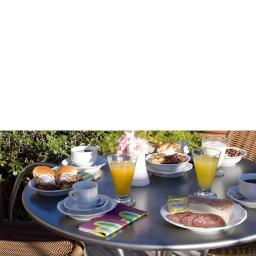}} &
\frame{\includegraphics[width=0.15\textwidth]{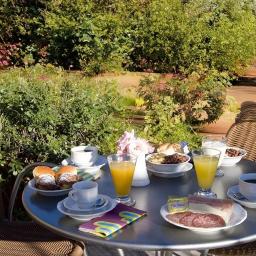}} &
\frame{\includegraphics[width=0.15\textwidth]{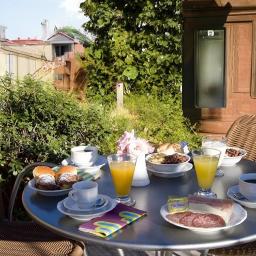}} &
\frame{\includegraphics[width=0.15\textwidth]{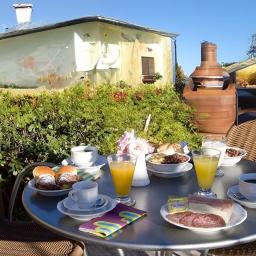}} &
\frame{\includegraphics[width=0.15\textwidth]{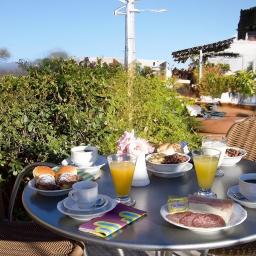}} &
\frame{\includegraphics[width=0.15\textwidth]{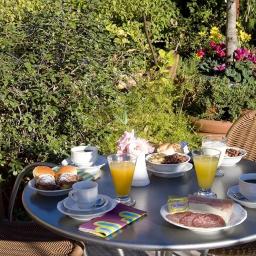}}\\

\frame{\includegraphics[width=0.15\textwidth]{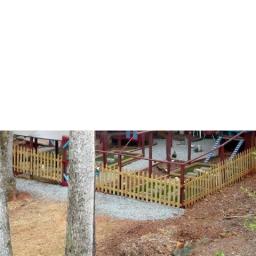}} &
\frame{\includegraphics[width=0.15\textwidth]{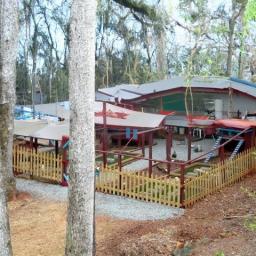}} &
\frame{\includegraphics[width=0.15\textwidth]{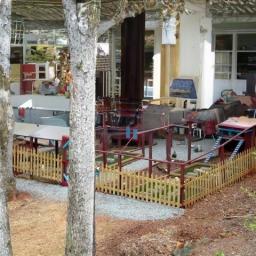}} &
\frame{\includegraphics[width=0.15\textwidth]{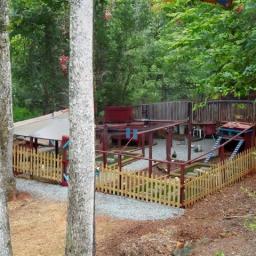}} &
\frame{\includegraphics[width=0.15\textwidth]{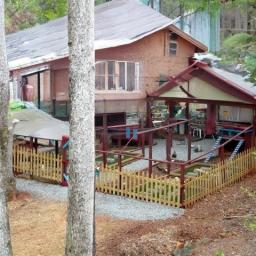}} &
\frame{\includegraphics[width=0.15\textwidth]{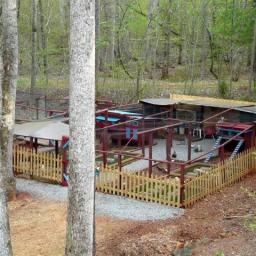}}\\

\frame{\includegraphics[width=0.15\textwidth]{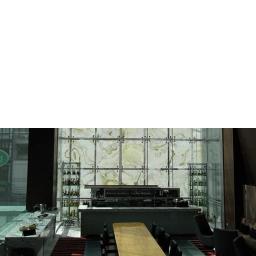}} &
\frame{\includegraphics[width=0.15\textwidth]{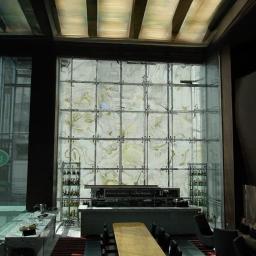}} &
\frame{\includegraphics[width=0.15\textwidth]{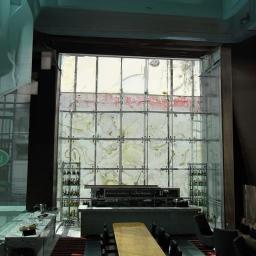}} &
\frame{\includegraphics[width=0.15\textwidth]{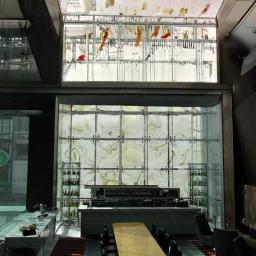}} &
\frame{\includegraphics[width=0.15\textwidth]{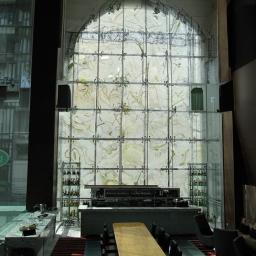}} &
\frame{\includegraphics[width=0.15\textwidth]{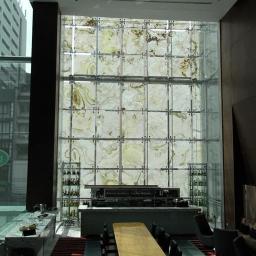}}\\

\frame{\includegraphics[width=0.15\textwidth]{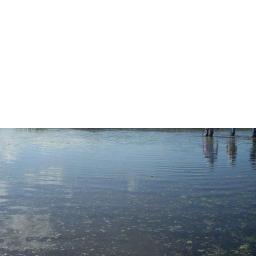}} &
\frame{\includegraphics[width=0.15\textwidth]{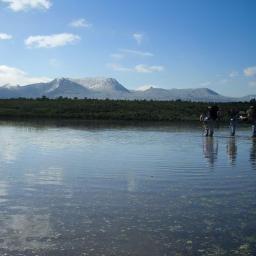}} &
\frame{\includegraphics[width=0.15\textwidth]{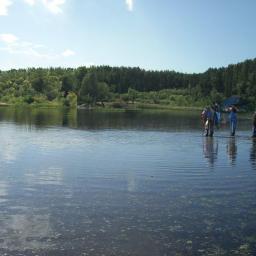}} &
\frame{\includegraphics[width=0.15\textwidth]{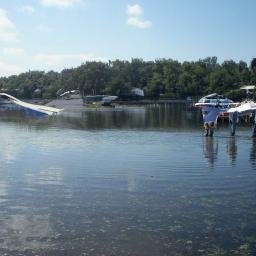}} &
\frame{\includegraphics[width=0.15\textwidth]{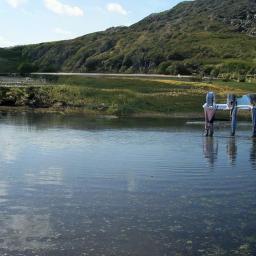}} &
\frame{\includegraphics[width=0.15\textwidth]{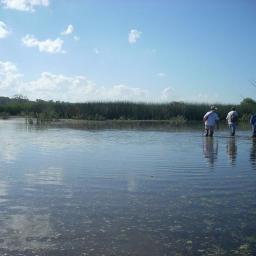}}\\


\end{tabular}
}
\end{center}
\vspace*{-0.5cm}
\caption{Diversity of \model outputs on Top uncropping on Places2 dataset.
\label{fig:uncropping_diversity_appendix_3}
}
\end{figure*}

\begin{figure*}[t]
\setlength{\tabcolsep}{2pt}
\begin{center}
{\small 
\begin{tabular}{cccccc}
{\small Input} & {\small Sample 1}  & {\small Sample 2} & {\small Sample 3} &  {\small Sample 4} & {Original} 
\\

\frame{\includegraphics[width=0.15\textwidth]{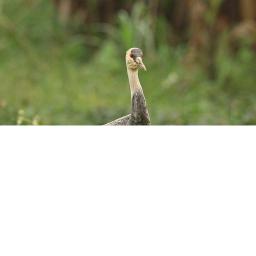}} &
\frame{\includegraphics[width=0.15\textwidth]{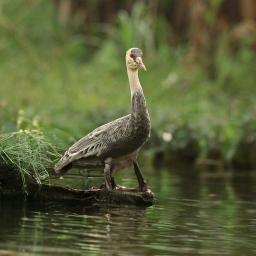}} &
\frame{\includegraphics[width=0.15\textwidth]{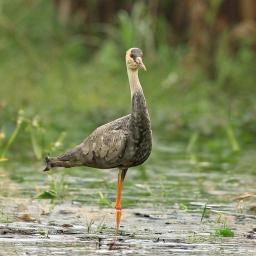}} &
\frame{\includegraphics[width=0.15\textwidth]{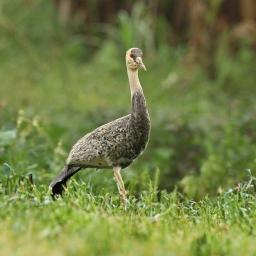}} &
\frame{\includegraphics[width=0.15\textwidth]{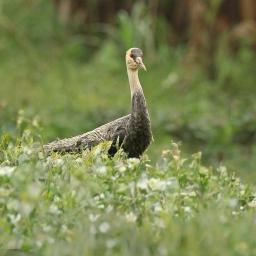}} &
\frame{\includegraphics[width=0.15\textwidth]{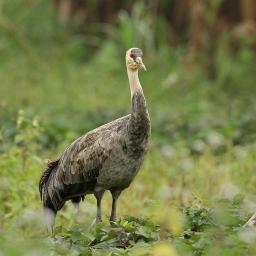}}\\

\frame{\includegraphics[width=0.15\textwidth]{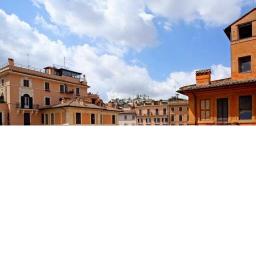}} &
\frame{\includegraphics[width=0.15\textwidth]{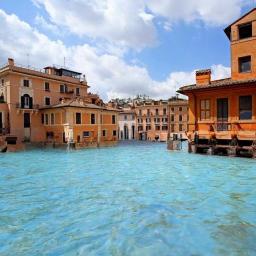}} &
\frame{\includegraphics[width=0.15\textwidth]{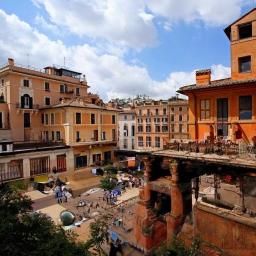}} &
\frame{\includegraphics[width=0.15\textwidth]{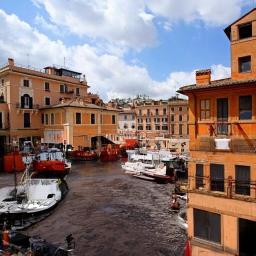}} &
\frame{\includegraphics[width=0.15\textwidth]{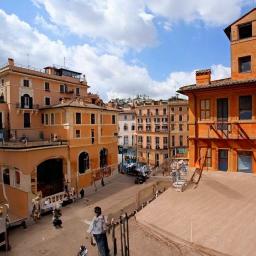}} &
\frame{\includegraphics[width=0.15\textwidth]{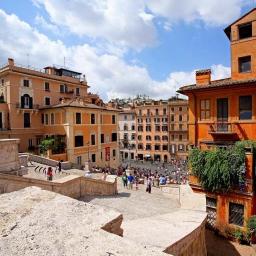}}\\

\frame{\includegraphics[width=0.15\textwidth]{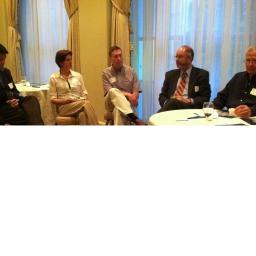}} &
\frame{\includegraphics[width=0.15\textwidth]{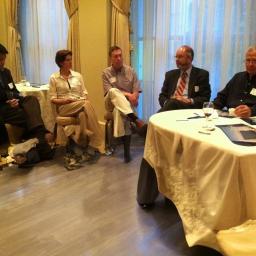}} &
\frame{\includegraphics[width=0.15\textwidth]{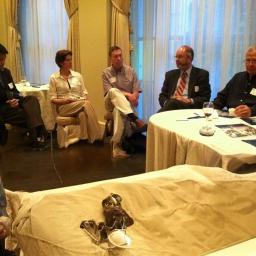}} &
\frame{\includegraphics[width=0.15\textwidth]{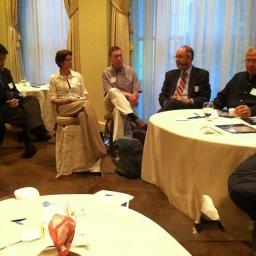}} &
\frame{\includegraphics[width=0.15\textwidth]{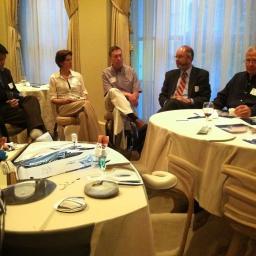}} &
\frame{\includegraphics[width=0.15\textwidth]{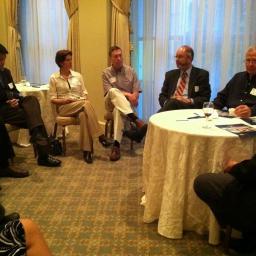}}\\

\frame{\includegraphics[width=0.15\textwidth]{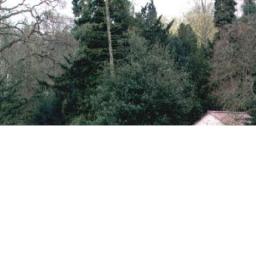}} &
\frame{\includegraphics[width=0.15\textwidth]{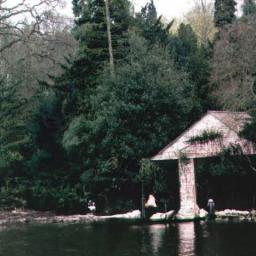}} &
\frame{\includegraphics[width=0.15\textwidth]{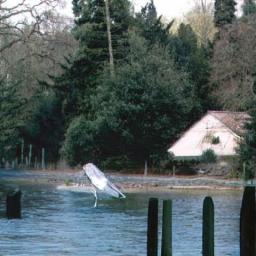}} &
\frame{\includegraphics[width=0.15\textwidth]{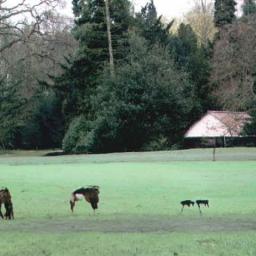}} &
\frame{\includegraphics[width=0.15\textwidth]{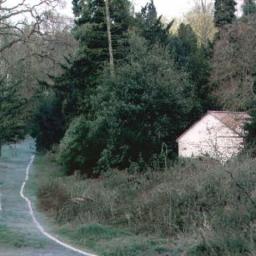}} &
\frame{\includegraphics[width=0.15\textwidth]{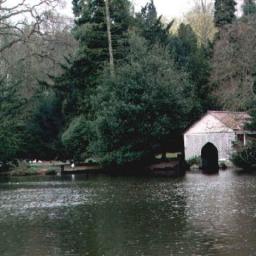}}\\

\frame{\includegraphics[width=0.15\textwidth]{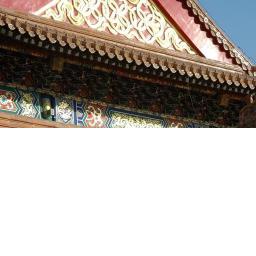}} &
\frame{\includegraphics[width=0.15\textwidth]{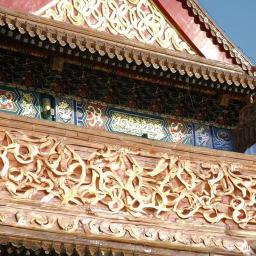}} &
\frame{\includegraphics[width=0.15\textwidth]{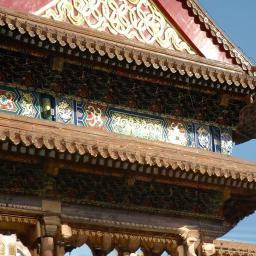}} &
\frame{\includegraphics[width=0.15\textwidth]{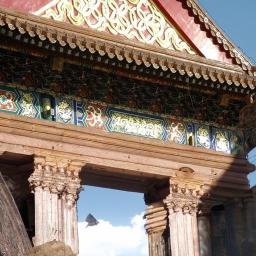}} &
\frame{\includegraphics[width=0.15\textwidth]{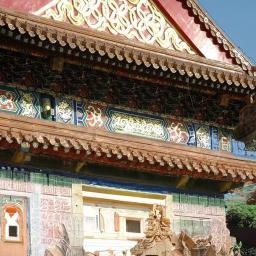}} &
\frame{\includegraphics[width=0.15\textwidth]{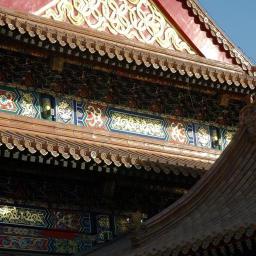}}\\

\frame{\includegraphics[width=0.15\textwidth]{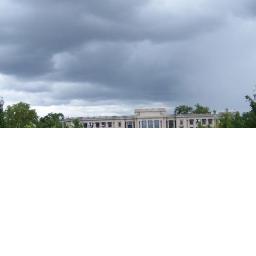}} &
\frame{\includegraphics[width=0.15\textwidth]{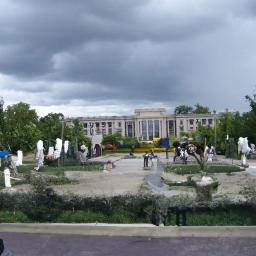}} &
\frame{\includegraphics[width=0.15\textwidth]{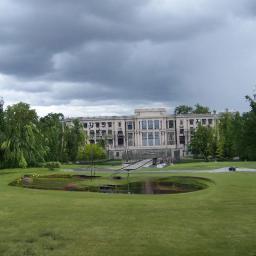}} &
\frame{\includegraphics[width=0.15\textwidth]{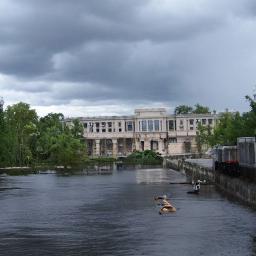}} &
\frame{\includegraphics[width=0.15\textwidth]{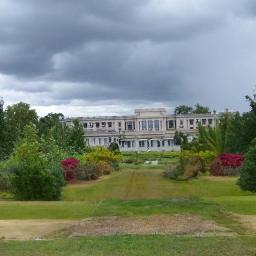}} &
\frame{\includegraphics[width=0.15\textwidth]{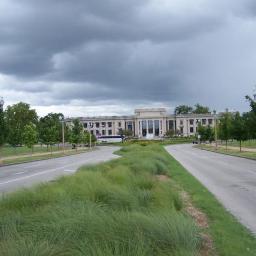}}\\

\frame{\includegraphics[width=0.15\textwidth]{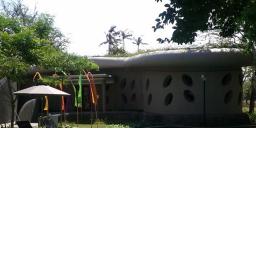}} &
\frame{\includegraphics[width=0.15\textwidth]{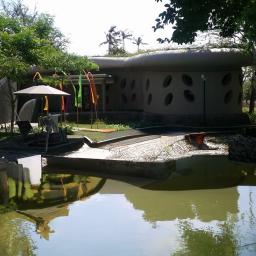}} &
\frame{\includegraphics[width=0.15\textwidth]{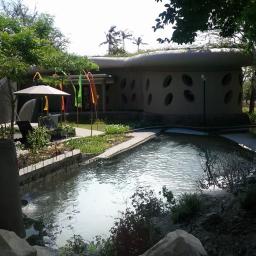}} &
\frame{\includegraphics[width=0.15\textwidth]{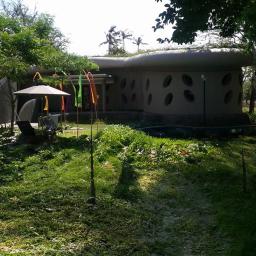}} &
\frame{\includegraphics[width=0.15\textwidth]{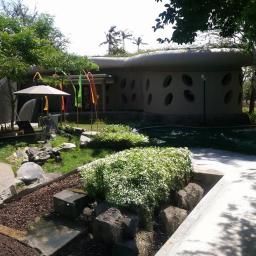}} &
\frame{\includegraphics[width=0.15\textwidth]{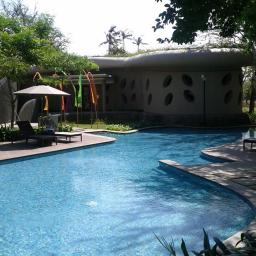}}\\


\end{tabular}
}
\end{center}
\vspace*{-0.5cm}
\caption{Diversity of \model outputs on Bottom uncropping on Places2 dataset.
\label{fig:uncropping_diversity_appendix_4}
}
\end{figure*}

\begin{figure*}[t]
\setlength{\tabcolsep}{2pt}
\begin{center}
{\small 
\begin{tabular}{cccccc}
{\small Input} & {\small Sample 1}  & {\small Sample 2} & {\small Sample 3} &  {\small Sample 4} & {Original} 
\\
\frame{\includegraphics[width=0.15\textwidth]{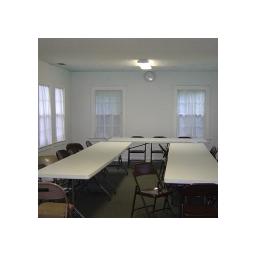}} &
\frame{\includegraphics[width=0.15\textwidth]{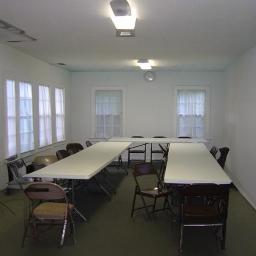}} &
\frame{\includegraphics[width=0.15\textwidth]{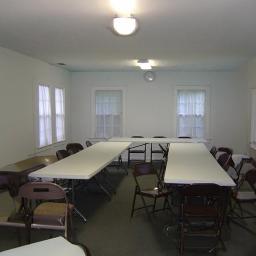}} &
\frame{\includegraphics[width=0.15\textwidth]{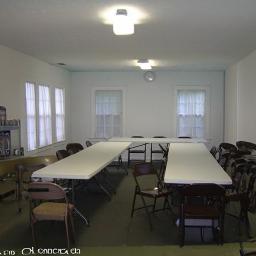}} &
\frame{\includegraphics[width=0.15\textwidth]{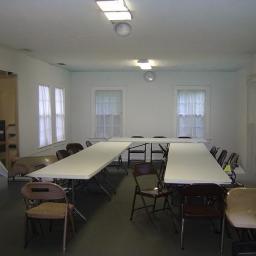}} &
\frame{\includegraphics[width=0.15\textwidth]{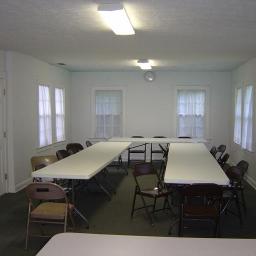}}\\

\frame{\includegraphics[width=0.15\textwidth]{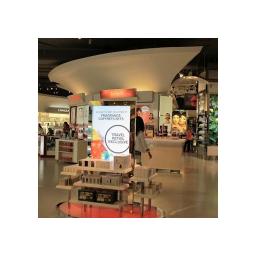}} &
\frame{\includegraphics[width=0.15\textwidth]{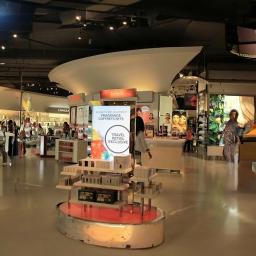}} &
\frame{\includegraphics[width=0.15\textwidth]{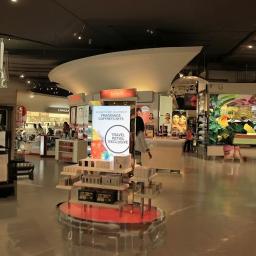}} &
\frame{\includegraphics[width=0.15\textwidth]{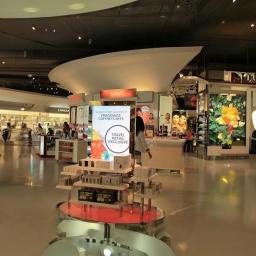}} &
\frame{\includegraphics[width=0.15\textwidth]{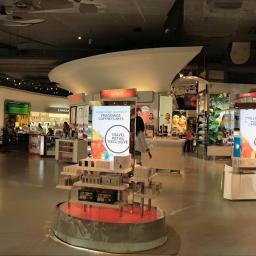}} &
\frame{\includegraphics[width=0.15\textwidth]{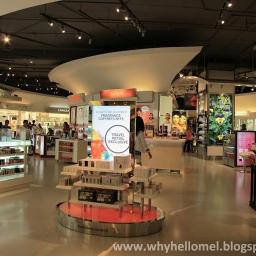}}\\

\frame{\includegraphics[width=0.15\textwidth]{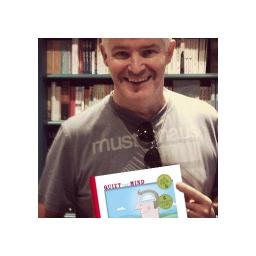}} &
\frame{\includegraphics[width=0.15\textwidth]{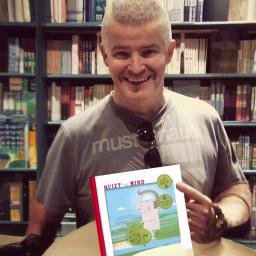}} &
\frame{\includegraphics[width=0.15\textwidth]{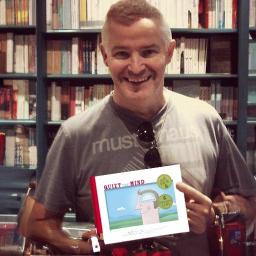}} &
\frame{\includegraphics[width=0.15\textwidth]{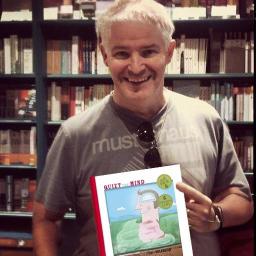}} &
\frame{\includegraphics[width=0.15\textwidth]{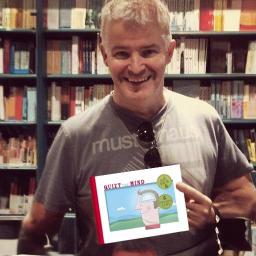}} &
\frame{\includegraphics[width=0.15\textwidth]{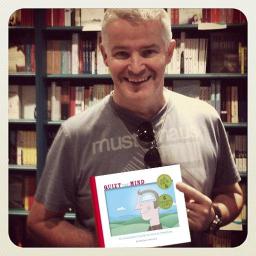}}\\

\frame{\includegraphics[width=0.15\textwidth]{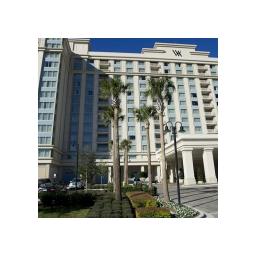}} &
\frame{\includegraphics[width=0.15\textwidth]{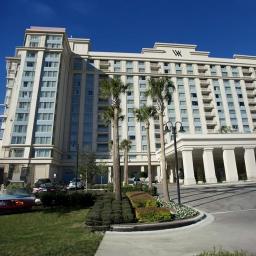}} &
\frame{\includegraphics[width=0.15\textwidth]{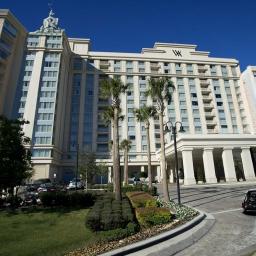}} &
\frame{\includegraphics[width=0.15\textwidth]{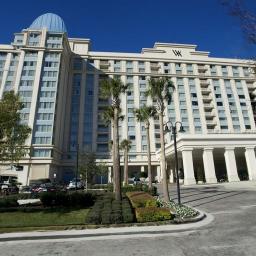}} &
\frame{\includegraphics[width=0.15\textwidth]{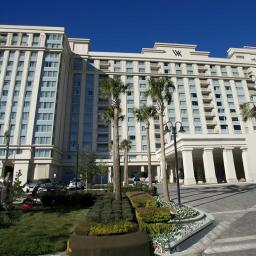}} &
\frame{\includegraphics[width=0.15\textwidth]{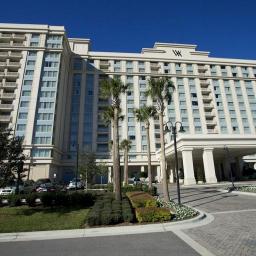}}\\

\frame{\includegraphics[width=0.15\textwidth]{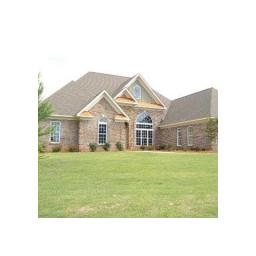}} &
\frame{\includegraphics[width=0.15\textwidth]{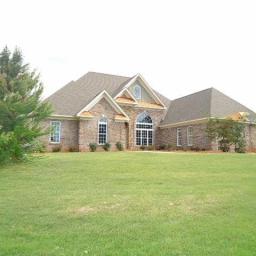}} &
\frame{\includegraphics[width=0.15\textwidth]{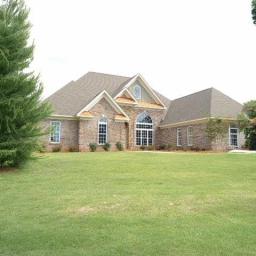}} &
\frame{\includegraphics[width=0.15\textwidth]{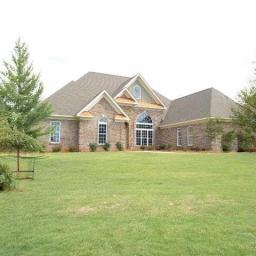}} &
\frame{\includegraphics[width=0.15\textwidth]{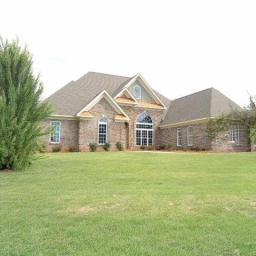}} &
\frame{\includegraphics[width=0.15\textwidth]{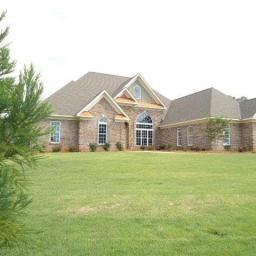}}\\

\frame{\includegraphics[width=0.15\textwidth]{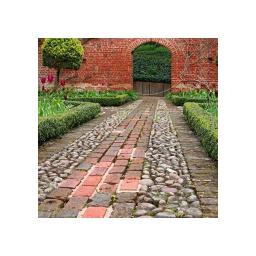}} &
\frame{\includegraphics[width=0.15\textwidth]{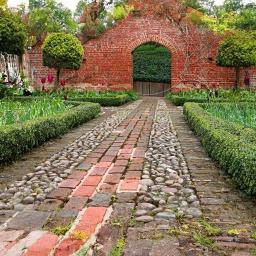}} &
\frame{\includegraphics[width=0.15\textwidth]{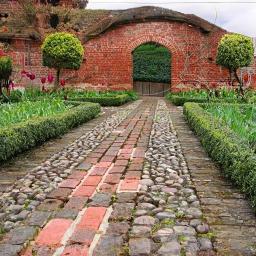}} &
\frame{\includegraphics[width=0.15\textwidth]{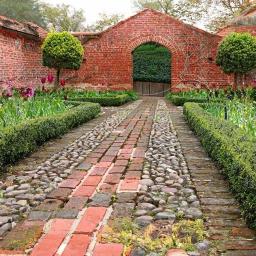}} &
\frame{\includegraphics[width=0.15\textwidth]{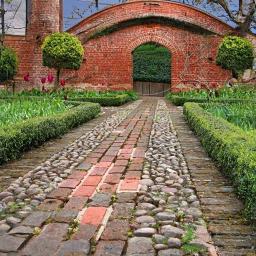}} &
\frame{\includegraphics[width=0.15\textwidth]{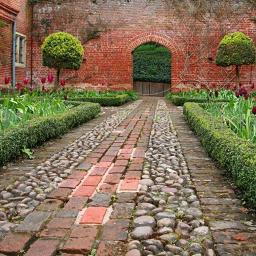}}\\

\frame{\includegraphics[width=0.15\textwidth]{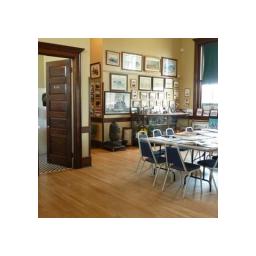}} &
\frame{\includegraphics[width=0.15\textwidth]{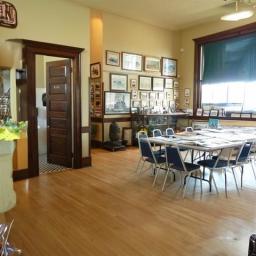}} &
\frame{\includegraphics[width=0.15\textwidth]{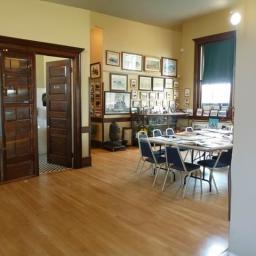}} &
\frame{\includegraphics[width=0.15\textwidth]{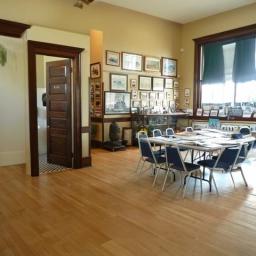}} &
\frame{\includegraphics[width=0.15\textwidth]{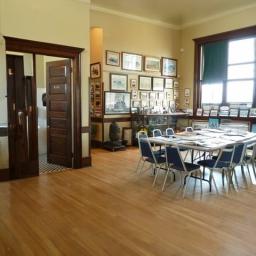}} &
\frame{\includegraphics[width=0.15\textwidth]{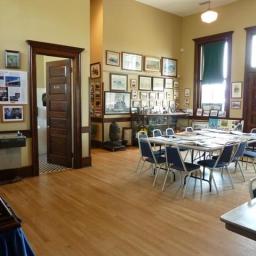}}\\


\end{tabular}
}
\end{center}
\vspace*{-0.5cm}
\caption{Diversity of \model outputs on Four Sided uncropping on Places2 dataset.
\label{fig:uncropping_diversity_appendix_5}
}
\end{figure*}

\subsection{JPEG Restoration}
In order to be consistent with other tasks, we perform training and evaluation on ImageNet dataset. Note that this is unlike most prior work \citep{dong2015compression, liu2018multi}, which mainly use small datasets such as DIV2K \citep{agustsson-cvprw-2017} and BSD500 \citep{MartinFTM01} for training and evaluation. Recent works such as \citep{galteri2019deep} use a relatively larger MS-COCO dataset for training, however, to the best of our knowledge, we are the first to train and evaluate JPEG restoration on ImageNet. We compare \model with a strong Regression baseline which uses an identical architecture. We report results on JPEG quality factor settings of 5, 10 and 20 in Table \ref{tab:compression_results}.

\subsection{Evaluation and Benchmarking Details}
\label{evaluation_detalis_appendix}

Several existing works report automated metrics such as FID, Inception Score, etc. \citep{coltran, lin2021infinitygan, yi2020contextual} but often lack key details such as the subset of images used for computing these metrics, or the reference distribution used for calculating FID scores. This makes direct comparison with such reported metrics difficult. Together with advocating for our proposed benchmark validation sets, we also provide all the necessary details to exactly replicate our reported results. We encourage future works to adopt a similar practice of reporting all the necessary evaluation details in order to facilitate direct comparison with their methods. 

\textbf{Benchmark datasets}: For ImageNet evaluation, we use the 10,000 image subset from ImageNet validation set - \textbf{ctest10k} introduced by \citep{larsson2016learning}. While this subset has been primarliy used for evaluation in the colorization literature \citep{guadarrama2017pixcolor, su2020instance, kim2021deep}, we extend its use for other image-to-image translation tasks. Many image-to-image translation tasks such as inpainting, uncropping are evaluated on Places2 dataset \citep{zhou2017places}. However, to the best of our knowledge, there is no such standardized subset for Places2 validation set used for benchmarking. To this end, we introduce \textbf{places10k}, a 10,950 image subset of Places2 validation set. Similar to ctest10k, we make places10k class balanced with 30 images per class (Places2 dataset has 365 classes/categories in total.). 

\textbf{Metrics}: We report several automated metrics for benchmarking and comparison with existing methods. Specifically, we report \textbf{Fréchet Inception Distance (FID)}, \textbf{Inception Score}, \textbf{Perceptual Distance} and \textbf{Classification Accuracy} for qualitative comparison. When computing FID scores, the choice of the reference distribution is important, but is often not clarified in existing works. In our work, we use the full validation set as the reference distribution, i.e. 50k images from ImageNet validation set for computing scores on ImageNet subset ctest10k, and 36.5k images from Places2 validation set for computing scores on Places2 subset places10k. For Perceptual Distance, we use the Euclidean distance in the $pool\_3$ feature space of the pre-trained InceptionV1 network (same as the features used for calculating FID scores). We use 
EfficientNet-B0 \footnote{\url{https://tfhub.dev/google/efficientnet/b0/classification/1}} top-1 accuracy for reporting Classification Accuracy scores.



\vspace{-.1cm}
\section{Limitations}
\vspace{-.1cm}
While \model achieves strong results on several image-to-image translation tasks demonstrating the generality and versatility of the emerging diffusion models, there are many important limitations to address. Diffusion models generally require large number of refinement steps during sample generation (e.g. we use 1k refinement steps for \model throughout the paper) resulting in significantly slower inference compared to GAN based models. This is an active area of research, and several new techniques \citep{nichol2021improved, watson2021learning, jolicoeur2021gotta} have been proposed to reduce the number of refinement steps significantly. We leave application of these techniques on \model to future work. \model's use of group-normalization and self-attention layers prevents its generalizability to arbitrary input image resolutions, limiting its practical usability. Techniques to adapt such models to arbitrary resolutions such as fine-tuning, or patch based inference can be an interesting direction of research. Like other generative models, \model also suffers from implicit biases, which should be studied and mitigated before deployment in practice.

\begin{figure*}[h]
\setlength{\tabcolsep}{2pt}
\begin{center}
{\small 
\begin{tabular}{cccc}
{\small Input (QF = 5) } & {\small Regression}  & {\small \model} & {\small Original} \\

{\includegraphics[width=0.163\textwidth]{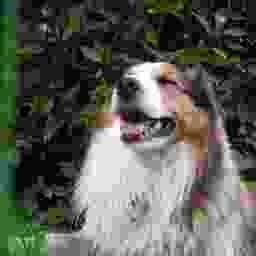}} &
{\includegraphics[width=0.163\textwidth]{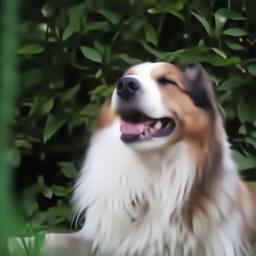}} &
{\includegraphics[width=0.163\textwidth]{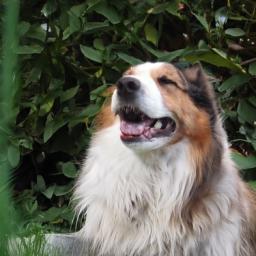}} &
{\includegraphics[width=0.163\textwidth]{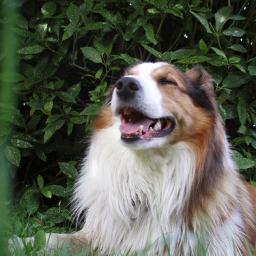}} \\

{\includegraphics[width=0.163\textwidth]{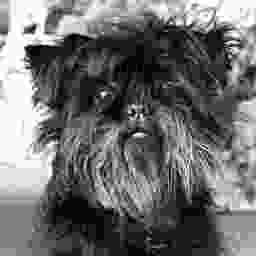}} &
{\includegraphics[width=0.163\textwidth]{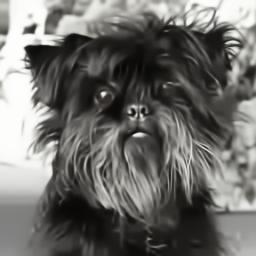}} &
{\includegraphics[width=0.163\textwidth]{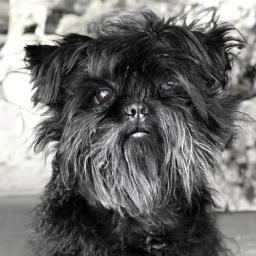}} &
{\includegraphics[width=0.163\textwidth]{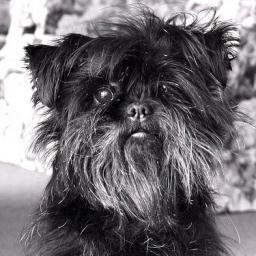}} \\

{\includegraphics[width=0.163\textwidth]{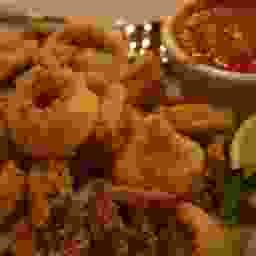}} &
{\includegraphics[width=0.163\textwidth]{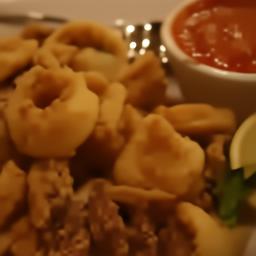}} &
{\includegraphics[width=0.163\textwidth]{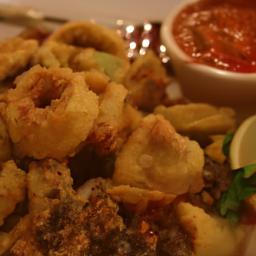}} &
{\includegraphics[width=0.163\textwidth]{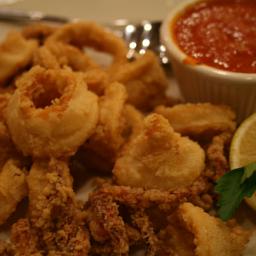}} \\

{\includegraphics[width=0.163\textwidth]{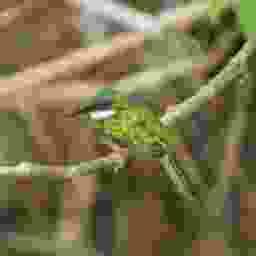}} &
{\includegraphics[width=0.163\textwidth]{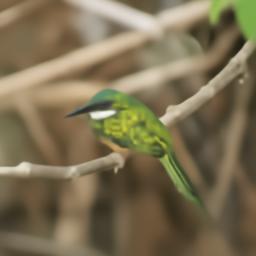}} &
{\includegraphics[width=0.163\textwidth]{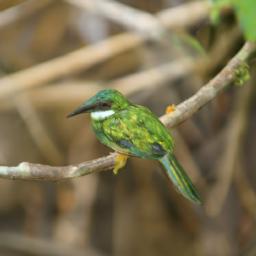}} &
{\includegraphics[width=0.163\textwidth]{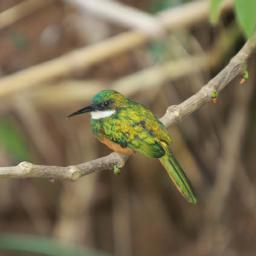}} \\

{\includegraphics[width=0.163\textwidth]{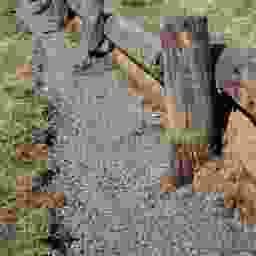}} &
{\includegraphics[width=0.163\textwidth]{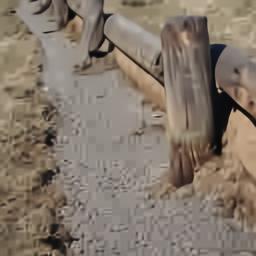}} &
{\includegraphics[width=0.163\textwidth]{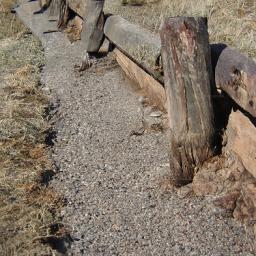}} &
{\includegraphics[width=0.163\textwidth]{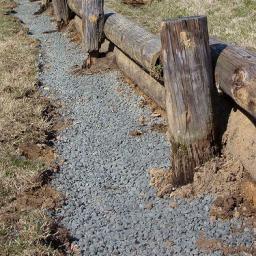}} \\

{\includegraphics[width=0.163\textwidth]{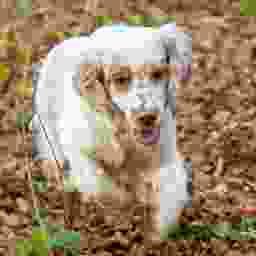}} &
{\includegraphics[width=0.163\textwidth]{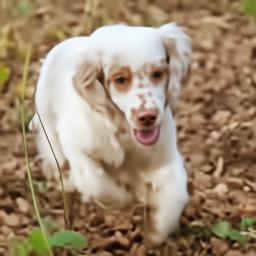}} &
{\includegraphics[width=0.163\textwidth]{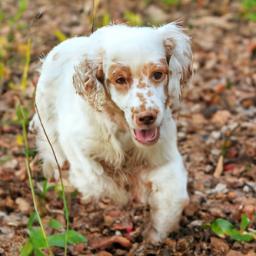}} &
{\includegraphics[width=0.163\textwidth]{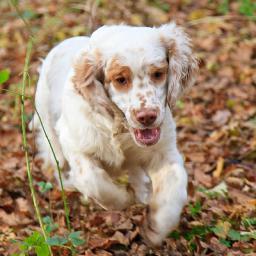}} \\

{\includegraphics[width=0.163\textwidth]{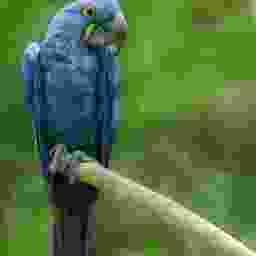}} &
{\includegraphics[width=0.163\textwidth]{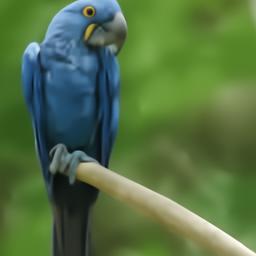}} &
{\includegraphics[width=0.163\textwidth]{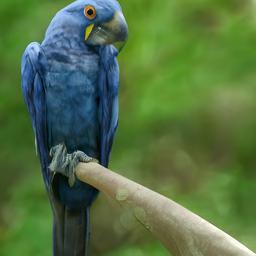}} &
{\includegraphics[width=0.163\textwidth]{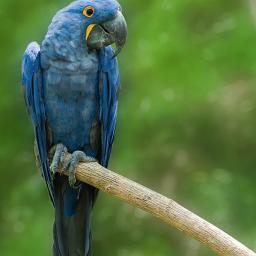}} \\


\end{tabular}
}
\end{center}
\vspace*{-0.6cm}
\caption{JPEG Restoration results on ImageNet images. \label{fig:jpeg_comparison_appendix}}
\end{figure*}

\begin{figure*}[h!]
\setlength{\tabcolsep}{1pt}
\begin{center}
{\small 
\begin{tabular}{c}
{\includegraphics[width=\textwidth]{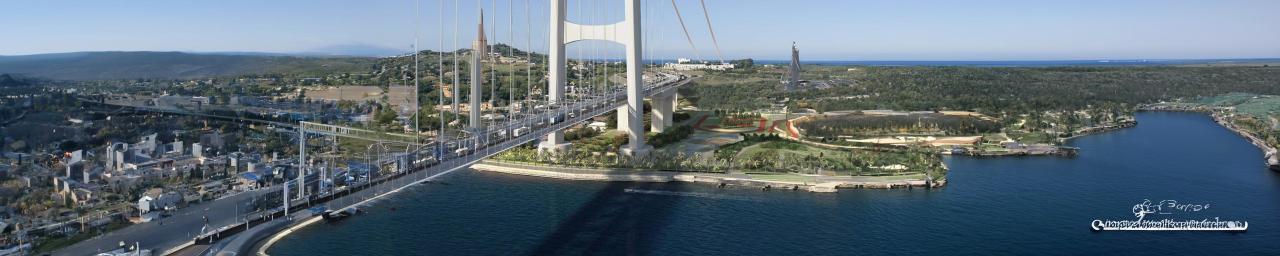}} \\
{\includegraphics[width=\textwidth]{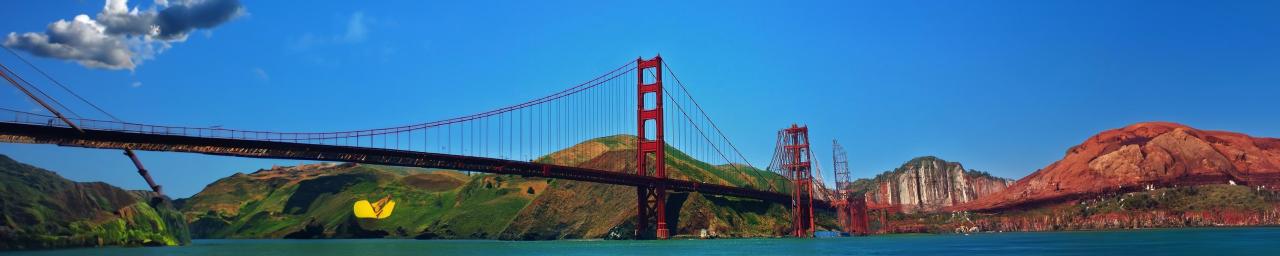}} \\
{\includegraphics[width=\textwidth]{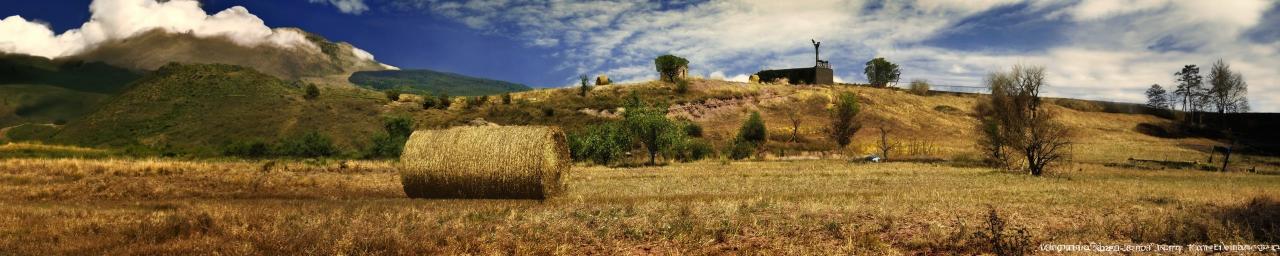}} \\
{\includegraphics[width=\textwidth]{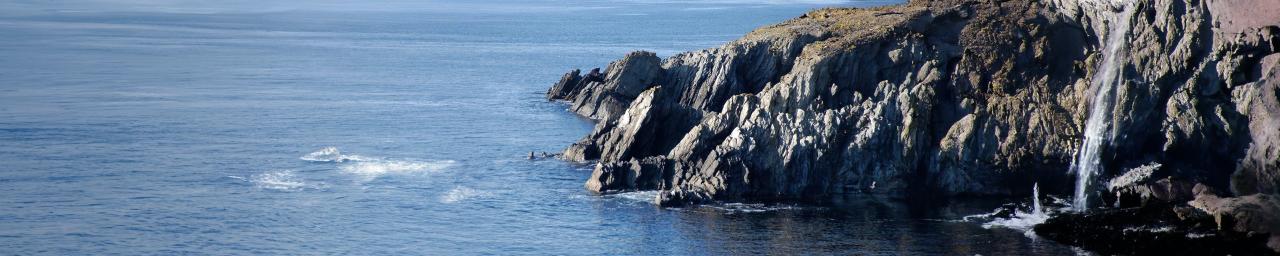}} \\
{\includegraphics[width=\textwidth]{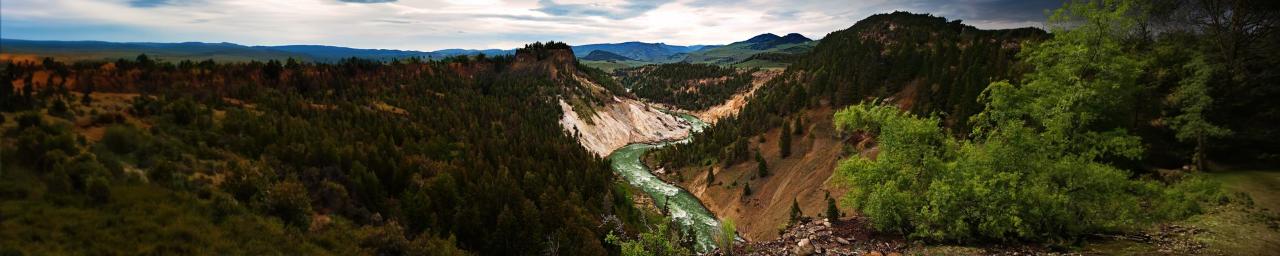}} \\
{\includegraphics[width=\textwidth]{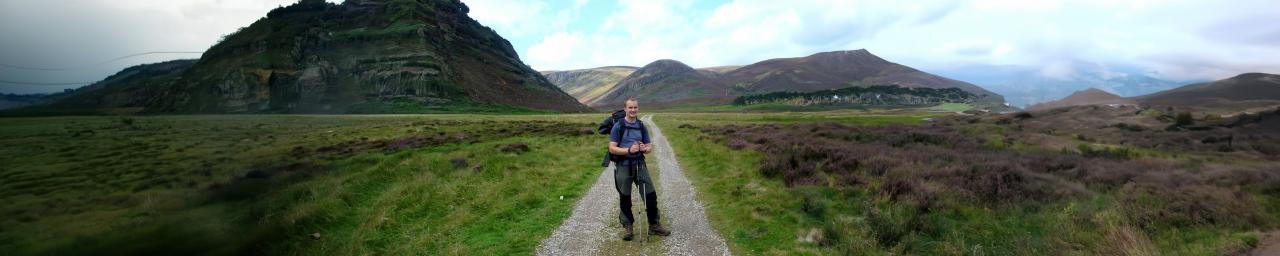}} \\
\end{tabular}
}
\end{center}
\vspace*{-0.35cm}
\caption{
\model panorama uncropping. Given the center 256$\times$256 pixels, 
we extrapolate 512 pixels to the right and to the left, in steps of 128 (via 50\% uncropping tasks), yielding a 256$\times$1280 panorama.
\label{fig:uncropping_panoramas_appendix}}
\end{figure*}


\begin{figure*}[h!]
\setlength{\tabcolsep}{1pt}
\begin{center}
{\small 
\begin{tabular}{c}
{\includegraphics[width=\textwidth]{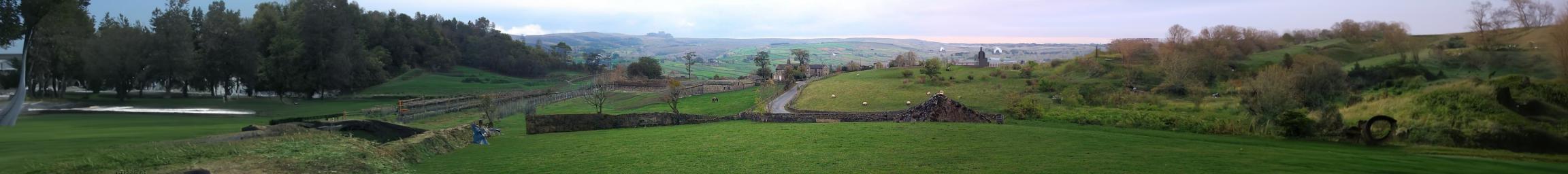}} \\
{\includegraphics[width=\textwidth]{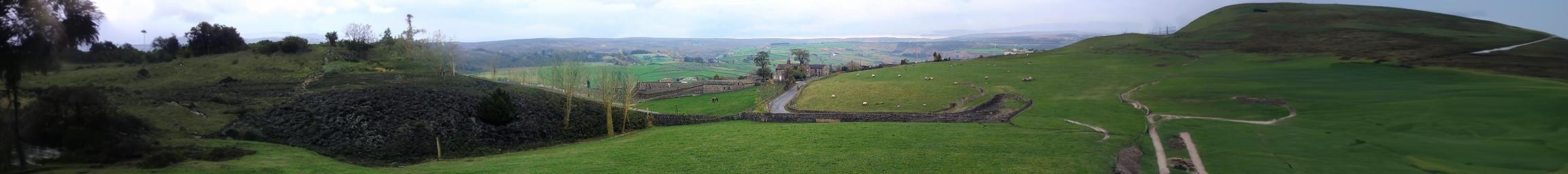}} \\
{\includegraphics[width=\textwidth]{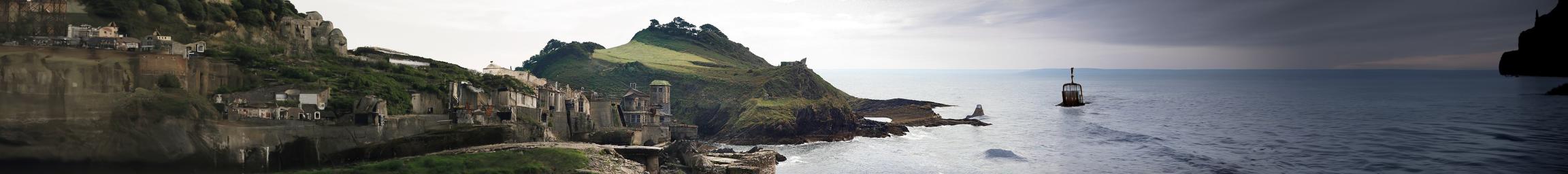}} \\
{\includegraphics[width=\textwidth]{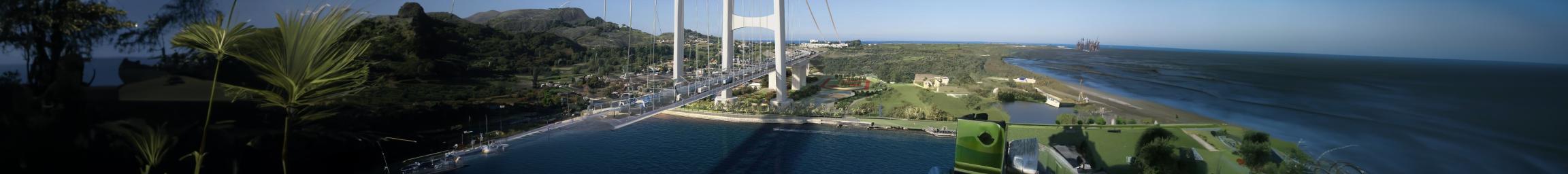}} \\
{\includegraphics[width=\textwidth]{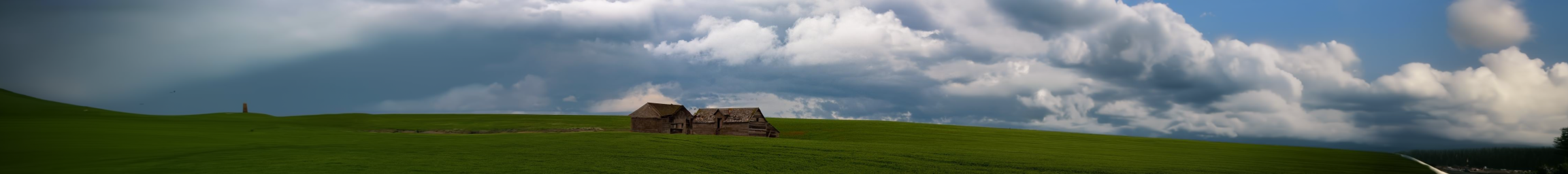}} \\
{\includegraphics[width=\textwidth]{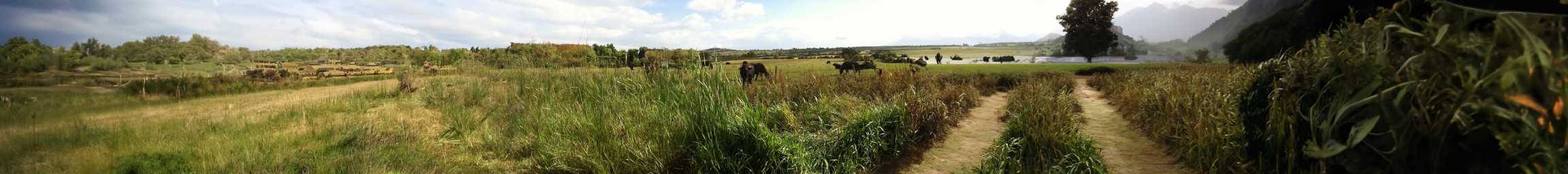}} \\
{\includegraphics[width=\textwidth]{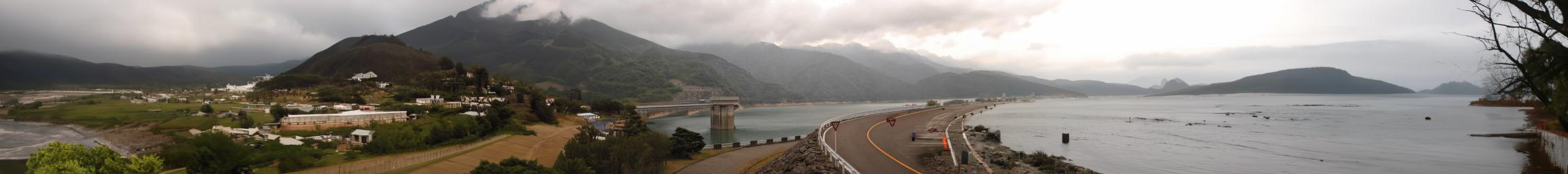}} \\
{\includegraphics[width=\textwidth]{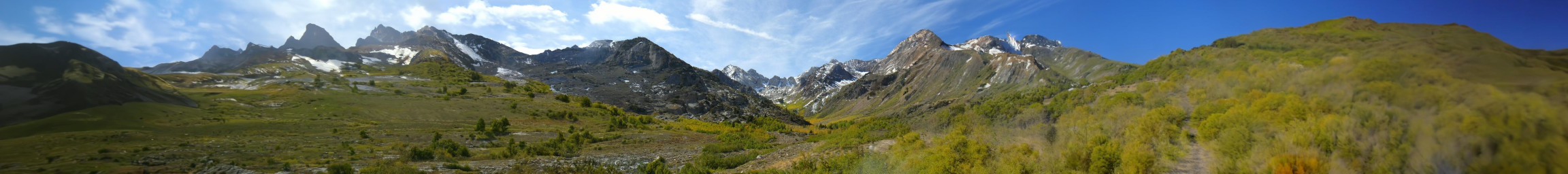}} \\ [.3em]
\end{tabular}
}
\end{center}
\vspace*{-0.35cm}
\caption{
\model panorama uncropping. Given the center 256$\times$256 pixels, 
we extrapolate 1024 pixels to the right and to the left, in steps of 128 (via 50\% uncropping tasks), yielding a 256$\times$2304 panorama.
\label{fig:uncropping_panoramas_appendix}}
\end{figure*}

\end{document}